\ifdefined\pdfminorversion\pdfminorversion=7\fi
\documentclass[pdflatex,sn-mathphys-ay]{topic3-arxiv}
\ifdefined\pdfsuppresswarningpagegroup\pdfsuppresswarningpagegroup=1\fi
\usepackage{amsmath,amssymb,amsthm,bm}
\usepackage{booktabs,enumitem,graphicx,placeins,makecell}
\usepackage{xcolor}
\setlist[itemize,1]{label=\ensuremath{\bullet}}
\hypersetup{
  hidelinks,
  pdftitle={Sparse Regression Distilled from a Single Robust Fit},
  pdfauthor={Wooyoung Shin and Seunghwan Park},
  pdfsubject={Robust sparse global explanation},
  pdfkeywords={robust regression, sparse global surrogate, explainable
    modelling, SCAD, explanation stability, variable selection}
}

\newtheorem{theorem}{Theorem}
\newtheorem{lemma}{Lemma}
\newtheorem{proposition}{Proposition}
\newtheorem{corollary}{Corollary}
\newtheorem{remark}{Remark}
\newtheorem{assumption}{Assumption}
\newcounter{algorithm}

\newcommand{\X}{\mathbf{X}}
\newcommand{\y}{\mathbf{y}}
\newcommand{\bbeta}{\bm{\beta}}
\newcommand{\btil}{\tilde{\bm{\beta}}}
\newcommand{\bcomp}{\hat{\bm{\beta}}^{\mathrm{comp}}}
\newcommand{\bora}{\hat{\bm{\beta}}^{\mathrm{ora}}}
\newcommand{\bLSA}{\hat{\bm{\beta}}^{\mathrm{LSA}}}
\newcommand{\cA}{\mathcal{A}}
\newcommand{\cZ}{\mathcal{Z}}
\newcommand{\RSS}{\mathrm{RSS}}
\DeclareMathOperator*{\argmin}{arg\,min}

\newif\iftopicthreemainrefs
\newif\iftopicthreesupprefs
\topicthreemainrefstrue
\topicthreesupprefstrue
\newcommand{\cvheading}[1]{}
\newcommand{\dimtable}[1]{\csname topicthreeDim#1\endcsname}

\newcommand{\dimref}[1]{\ref{#1}}
\newcommand{\proofref}[1]{Appendix~\ref{#1}}
\newcommand{\splineref}{Appendix~\ref{app:spline}}
\newcommand{\ucixheading}[1]{\subsection{#1}}
\newcommand{\ORsec}[1]{Appendix~\ref{#1}}
\newcommand{\ORsub}[1]{Section~\ref{#1}}

\newcommand{\ORtab}[1]{Table~\ref{#1}}
\newcommand{\provref}{Appendix~\ref{app:prov}}

\newcommand{\screenthy}{}

\newcommand{\modernchecksintro}{Two further checks are included in this integrated preprint.}
\title[Sparse regression distilled from a robust fit]{Sparse Regression
Distilled from a Single Robust Fit}

\author[1]{\fnm{Wooyoung} \sur{Shin}}
\author*[1]{\fnm{Seunghwan} \sur{Park}}\email{stat.shpark@kangwon.ac.kr}
\affil[1]{\orgdiv{Department of Information Statistics},
  \orgname{Kangwon National University},
  \orgaddress{\street{1, Kangwondaehak-gil}, \city{Chuncheon-si},
    \state{Gangwon-do}, \postcode{24341}, \country{Republic of Korea}}}

\begin{document}
\abstract{
Robust linear fits can resist response contamination yet remain too dense or
unstable for useful global explanations. We propose \emph{penalized
distillation}, which fits a smoothly clipped absolute deviation (SCAD)
estimator to a robust initial estimator's empirical fitted surface along a
safeguarded coordinate-descent path and evaluates candidate states separately
for fidelity, parsimony, perturbation stability, and held-out prediction.
The new results attach to the states the algorithm actually computes.
Conditional on a fixed uncontaminated design, deterministic bounds transfer
response-replacement boundedness from the initial fit to every retained path
state. Turning to fixed dimension, we characterize the oracle-support branch
by its empirical-Gram projection and influence function, give conditions for
covariance-weighted least-squares approximation equivalence, and establish a
path-conditional generalized information criterion. By contrast, at large
dimension-to-sample ratios the full-coordinate robust fit collapses without
warning, and screening restores the construction. Under a sure-screening framework, the robustness bound and
the support and selection guarantees transfer to the screened fit.
Simulations separate robustness transfer from support recovery, efficiency,
and computation across the dimension-to-sample ratio, with $p$ up to $240$,
and the signal density, which isolates what the sparse stage adds once the
screen over-selects. In a duplicate-grouped superconductivity study, the
distilled estimator remains predictively stable under prespecified
training-response shifts but retains 66.8--68.8 of 81 slopes. Stronger
sparsification reduces the model to 12.6--14.0 slopes only at visible
fidelity and prediction cost. Distillation therefore preserves predictive
stability on these data without substantiating a compact coordinate-level
explanation.
}



\maketitle

\section{Introduction}\label{sec:intro}

A robust regression estimator protects the fit from contaminated
observations, but protecting the fit is not the same as explaining it. A
robust linear fit of a response on dozens of correlated measurements assigns
every measurement a coefficient, and each coefficient is individually
readable. The fitted surface as a whole is, however, still too dense to
communicate, and which coordinates carry it can change under modest
perturbations of the data. A sparse linear fit can compress that surface, but
it is a credible global explanation only to the extent that distillation
fidelity, parsimony, perturbation stability, and held-out prediction are
audited separately.

This explanation task combines response contamination with a compression
question, namely whether a small subset of the recorded covariates can
reproduce the fitted surface well enough to support the intended global
explanation. Robust estimation and sparse modelling each have mature
literatures. On the
robustness side,
M-estimators \citep{huber81}, high-breakdown
least-median-of-squares estimators \citep{rousseeuw84}, and MM-estimators
\citep{yohai87} deliver estimates that remain bounded and reasonably
efficient when a substantial fraction of the data is arbitrary. On the
sparsity side, penalized least squares with the lasso
\citep{tibshirani96}, SCAD \citep{fanli01} or MCP \citep{zhang10}
penalties selects variables and estimates coefficients in one operation,
with oracle guarantees available for the folded-concave penalties.

For the explanation task we instead treat sparsification as a separate
compression problem. Suppose a robust estimate $\btil$ has been computed once
by an estimator whose assumptions and diagnostics are appropriate for the
application, with fitted values $\X\btil$. We then compute a penalized
least-squares path in which those fitted values play the role of the response.
After profiling the common unpenalized intercept, the path uses the criterion
\begin{equation}\label{eq:distill}
Q_\lambda(\bbeta;\btil) \;=\;
\frac{1}{2n}\,\|\X\btil - \X\bbeta\|_2^2
+ \sum_{j=1}^p p_{\lambda}(|\beta_j|).
\end{equation}
We call \eqref{eq:distill} \emph{penalized distillation}: the fitted surface of a robust initial estimator is compressed into a sparse distilled estimator. The estimator we report is the GIC-selected member
of the safeguarded local coordinate-descent path in
Algorithm~\ref{alg:distill}.
The second stage is an ordinary penalized least-squares problem, so the
whole procedure costs one robust fit plus one Gram-matrix path.
The raw responses enter the path only through $\btil$ and
enter the GIC through one operational residual scale. Because of this separation,
boundedness under response-only replacement transfers in finite samples from
the initial fit to the states the path retains, conditional on a fixed,
uncontaminated design. The audit then treats prediction, support recovery,
and stability as separate questions, and Section~\ref{sec:gic} tabulates the
scope of every guarantee.

\subsection{Related work}\label{sec:related}

The least-absolute-deviations (LAD) lasso of \citet{wang07} adds an $\ell_1$
penalty to the absolute loss and enjoys selection consistency, but inherits
the vulnerability of $L_1$ regression to leverage. A single bad high-leverage
point suffices to break it, so its breakdown point against contamination in
the covariates is zero. Sparse least-trimmed-squares (LTS) regression
\citep{alfons13} instead minimizes the sum of the $h$ smallest squared
residuals plus an $\ell_1$ penalty. It attains a breakdown point of
$(n-h+1)/n$ that does not depend on $p$ and is computed by concentration
steps from many elemental starts, with a reweighting step to recover
efficiency. Its authors note that the estimator lacks an asymptotic theory
and that efficiency remains a concern. MM-lasso and adaptive MM-lasso
\citep{smucler17} couple a bounded, redescending M-loss with (adaptive)
$\ell_1$ penalties. The adaptive version attains the oracle property, but the
theory presumes a consistent estimate of the residual scale, and the
criterion is nonconvex. Penalized elastic-net S-estimation (PENSE)
\citep{cohenfreue19} and adaptive PENSE \citep{kepplinger23} extend this
programme to elastic-net penalization of S- and M-estimation. Adaptive PENSE
is closest in spirit to our proposal in that it, too, is a two-stage
construction whose properties are stated relative to a preliminary
estimator. That paper also emphasizes, as we do below, that the tuning
procedure itself must be robust for the breakdown guarantee to be
meaningful. Other integrated routes include $\gamma$-divergence with sparse
regularization in linear regression \citep{kawashimafujisawa17} and with an
elastic-net penalty in logistic regression \citep{cornillyetal24}. In contrast to these integrated routes, we
penalize the fitted values of a robust estimator that has already been
computed, so that the robust fit and the sparsification remain separate
stages.

In the explainable-modelling taxonomy, our distilled estimator is a
model-based sparse global surrogate, in that it approximates one initial fit
over a specified design distribution rather than explaining one prediction
locally \citep{burkarthuber21}. This positioning also fixes what must be
evaluated. In particular, the predictive--descriptive--relevance framework
of \citet{murdochetal19} separates predictive accuracy from how faithfully
an interpretation describes its source model and from its usefulness to a
stated audience. In addition, related work treats explicitness, fidelity,
and stability as distinct desiderata \citep{alvarezmelisjaakkola18}. Here the intended
audience is an analyst seeking a compact linear summary of a robust linear
fit, so we report held-out prediction, distillation fidelity, sparsity, and
perturbation stability separately. Stability can also be built into the
selector, as in the loss-guided stability selection of \citet{werner25}. We
instead keep it as a reported diagnostic, so that instability stays visible.

Related fitted-value constructions include preconditioning
\citep{paul08}, which applies the lasso to de-noised
supervised-principal-component predictions, and Bayesian decoupled shrinkage
and selection and projective prediction \citep{hahn15,piironen20}, which
approximate a posterior predictive fit by a sparse model. Transparent global
model distillation also predates the present construction. For example,
\citet{tanetal18} use an interpretable surrogate to mimic black-box scores
and compare it with a transparent outcome model for auditing. More recently, \citet{zhouetal24}
address reproducibility when candidate surrogates vary across regenerated
pseudo-samples. Their central-limit and multiple-testing procedure selects a
pseudo-sample size intended to stabilize the selected surrogate for a fixed
reference model. We instead retain the observed empirical design and
generate no pseudo-sample. Our stability diagnostics concern grouped folds,
prespecified response perturbations, correlated-coordinate substitution, and
the retained local path.

We distinguish the ordinary casewise replacement breakdown point
\citep{donoho83}, which permits replacement of whole $(\bm x_i,y_i)$ rows,
from the response-only breakdown point conditional on a fixed design that is
used in our transfer result. The integrated-method literature, including
\citet{alfons13} and \citet{kepplinger23}, generally reports the former or
an estimator-specific variant. For influence analysis, \citet{avella17}
shows why a global classical influence function is problematic for penalized
estimators with non-differentiable penalties. For this reason we condition
explicitly on a fixed-support, flat-active-status branch and analyze the
ordinary G\^ateaux derivative of that smooth branch functional. For tuning,
information criteria with robust scale estimates are standard in this
literature \citep{alfons13}, and the same device appears outside regression.
For example, \citet{cappozzo20} select among trimmed and constrained
classification models with a criterion evaluated on the retained
observations only. Relatedly, the pairing of penalty and selector
matters for oracle efficiency \citep{wangleng07}. One feature of the distillation-fidelity GIC result
established in Proposition~\ref{prop:gic} is that the scale estimate
entering the criterion need only be bounded, not consistent.

In this paper, we propose penalized distillation as a modular route to a
sparse robust regression and study it as an explanation of the
initial fit. Compared with the lineage above, our contribution is to
combine four elements that have not been treated jointly in the cited
work. First, we transfer boundedness under response-only replacement
from the initial fit to every state the path retains. Second, we
derive, for a robust linear initial fit, the empirical-Gram projection
and the influence function of the oracle-support branch. Third, we give
a selection-consistency result for the GIC conditional on the computed
path. Fourth, we audit both the nonconvex computation and the
explanation trade-offs explicitly.

The rest of the paper is organized as follows. Section~\ref{sec:method}
defines penalized distillation, its algorithm and its defaults.
Section~\ref{sec:theory} states the finite-sample and asymptotic
guarantees and tabulates their scope. Section~\ref{sec:sim} reports the
Monte Carlo study in fixed dimension, including the comparison with modern
robust-sparse procedures, and Section~\ref{sec:screen} extends the
construction to large $p/n$ through screening and reports its own study.
Section~\ref{sec:real} presents the superconductivity study, and
Section~\ref{sec:disc} concludes. Proofs, secondary simulations, full
tables and the computational provenance record are collected in Online
Resource~1. Throughout, sections, tables, figures and remarks prefixed by S
refer to it.

\section{Penalized distillation}\label{sec:method}

There are two ways to make an estimator simultaneously robust and sparse.
The \emph{integrated} route couples a robust loss with a sparsity penalty
in one criterion, as in sparse LTS \citep{alfons13}, MM-lasso
\citep{smucler17} or PENSE \citep{cohenfreue19}. It is direct, but the
resulting criterion is nonconvex with many local optima, must be
re-optimized at every point of the tuning grid, and for M- and S-type
losses it requires a residual scale that must itself be estimated
robustly. The \emph{modular} route, which we take, separates the two
concerns: one robust fit establishes what the clean signal looks like, and
a standard folded-concave path compresses that fit into a sparse model.

\begin{figure}[!htbp]
\centering
\includegraphics[width=\linewidth]{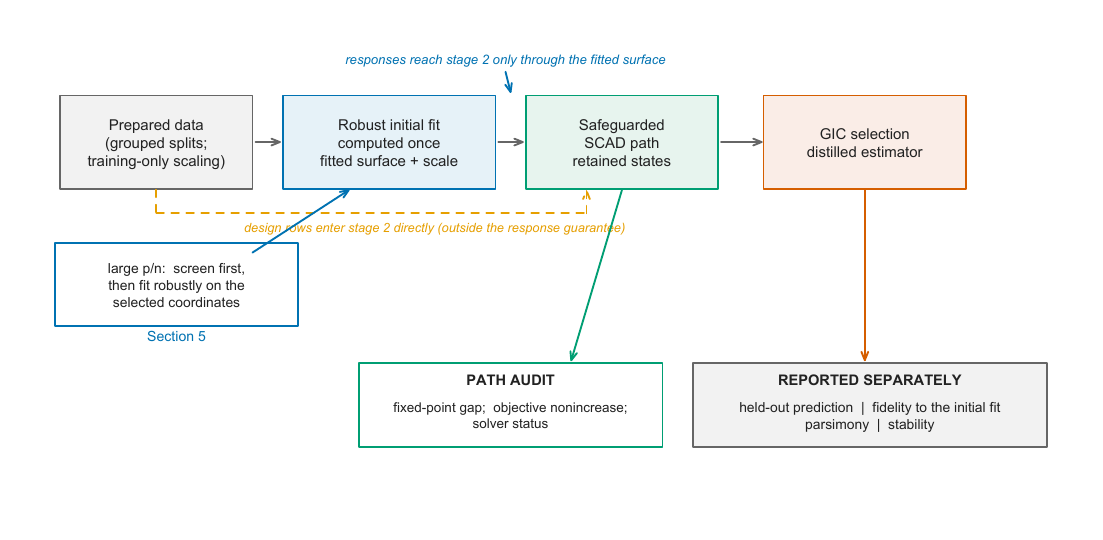}
\caption{Penalized-distillation workflow. The robust initial estimator is
fitted once; a safeguarded SCAD path is selected by GIC and then assessed on
four separate empirical axes. When $p/n$ is large the
initial fit is preceded by a screening step, developed in
Section~\ref{sec:screen}.}
\label{fig:workflow-audit}
\end{figure}

Figure~\ref{fig:workflow-audit} gives the workflow and separates the two
audits that accompany the distilled fit. Numerical checks establish which
local path state was retained, whereas the empirical assessment keeps the
four assessment axes of Section~\ref{sec:intro} distinct.

\subsection{The estimator}\label{sec:est}

Throughout we observe pairs $(\bm x_i,y_i)$, $i=1,\dots,n$, with
$\bm x_i\in\mathbb R^p$ and $p<n$ fixed. Here $\X$ is the $n\times p$ matrix
with rows $\bm x_i^\top$, and $\y=(y_1,\dots,y_n)^\top$. We
include a common unpenalized intercept in every
initial fit, distilled estimator, and comparator. Let $\tilde\alpha$ and $\btil$ denote the
initial intercept and slope vector, and write
$\tilde\y=\tilde\alpha\bm1+\X\btil$. Let $\bbeta_0$ denote the slope
estimand of the initial estimator, with active set
$\cA=\{j:\beta_{0j}\neq0\}$ and $\cZ=\cA^c$. Write
$\mathcal D_n=\{(\bm x_i,y_i):1\leq i\leq n\}$, choose a robust regression
rule $\mathcal T$, and first set
\begin{equation}\label{eq:step1}
(\tilde\alpha,\btil)=\mathcal T(\mathcal D_n).
\end{equation}
This abstraction includes MM, S, LTS, Huber-M, and quantile regression. We also record the operational residual scale used by the
implementation. If $\bm r=\y-\tilde\alpha\bm1-\X\btil$, let
$s_{\rm MAD}=\mathrm{MAD}(\bm r)$ be the median absolute deviation using the
normal-consistency factor $1.4826$, and let
$s_{\rm RMS}=\{n^{-1}\sum_i r_i^2\}^{1/2}$ be the root mean square (RMS).
Then set
\begin{equation}\label{eq:sigma}
\hat\sigma=
\begin{cases}
s_{\rm MAD}, & s_{\rm MAD}\text{ is finite and }s_{\rm MAD}>10^{-10},\\
s_{\rm RMS}, & \text{otherwise, if }s_{\rm RMS}\text{ is finite and }s_{\rm RMS}>10^{-10},\\
1, & \text{otherwise}.
\end{cases}
\end{equation}
This scale plays no part in estimation but calibrates the tuning criterion of
Section~\ref{sec:algo}; the RMS and unit branches keep the implemented
selector defined when the residual MAD is zero or nonfinite. Second, define the
explicit-intercept criterion
\begin{equation}\label{eq:step2}
Q_\lambda^+(\alpha,\bbeta;\tilde\y) \;=\;
\frac{1}{2n}\,\big\|\tilde\y-\alpha\bm1-\X\bbeta\big\|_2^2
+\sum_{j=1}^p p_{\lambda}(|\beta_j|),
\end{equation}
with $p_\lambda$ the SCAD penalty \citep{fanli01} with $a=3.7$. The
intercept is left unpenalized throughout so that it absorbs location shifts
under asymmetric errors. For each $\lambda$, Algorithm~\ref{alg:distill} defines the retained
computed state $(\hat\alpha^{\mathrm{comp}}(\lambda),\bcomp(\lambda))$ by a
safeguarded recursion over local coordinate-descent candidates, not by global
minimization of \eqref{eq:step2}.

For the theory we profile out the intercept and reuse $\X$ and $\btil$ for
the resulting centered slope problem. We reserve $\bcomp(\lambda)$ for a
retained computed path state and $\bora$ for the oracle-support local branch.
We use $\bar{\bm x}$ and $\bar{\tilde y}$ for the sample means of the
original, pre-centering covariates and fitted response.
Thus later displays that use $\X\btil$ concern the profiled problem, whereas
the implementation keeps the intercept explicit throughout.

At $\lambda=0$ the fitted surface is
reproduced exactly, and when the augmented design has full column rank the
coefficient vector equals $(\tilde\alpha,\btil)$, so the penalty path is
anchored at the initial estimator. At the
other extreme, when the design is orthogonal, $\X^\top\X/n=\mathbf I$, the
quadratic term separates across coordinates and \eqref{eq:step2} has the
closed form
\begin{equation}\label{eq:ortho}
\beta_j^\star(\lambda)\;=\;\Theta_{\mathrm{SCAD}}\big(\tilde\beta_j;\lambda\big),
\qquad j=1,\dots,p,
\end{equation}
where, under this normalization and with
$S(z,\lambda)=\operatorname{sign}(z)(|z|-\lambda)_+$,
\[
\Theta_{\mathrm{SCAD}}(z;\lambda)=
\begin{cases}
S(z,\lambda), & |z|\le2\lambda,\\
\{(a-1)z-a\lambda\operatorname{sign}(z)\}/(a-2),
  & 2\lambda<|z|\le a\lambda,\\
z, & |z|>a\lambda.
\end{cases}
\]
Thus, under an orthogonal design, distillation reduces to SCAD
thresholding of the coefficients of the robust fit. For a general design,
\eqref{eq:step2} contains one further ingredient, namely a projection
that recycles the information in the coordinates that are set to zero,
and it is this projection that yields the efficiency gain established in
Theorem~\ref{thm:oracle}.

Turning to the choice of $\mathcal T$\label{sec:init}, the asymptotic theory of
Section~\ref{sec:theory} requires $\btil$ to be
$\sqrt n$-consistent for its estimand, while the finite-sample transfer
requires boundedness under response-only replacement with $\X$ fixed. Any
initial estimator with these two properties qualifies. In practice, our default is a high-breakdown
  MM-estimator with nominal $50\%$ ordinary casewise breakdown and $95\%$
  asymptotic Gaussian efficiency. Its exact finite-sample ordinary casewise and
conditional response-only breakdown points are distinct quantities and may
depend on $n$, $p$, and the design, and Corollary~\ref{cor:bp} uses only the
latter. The base study also reports
Huber-M and OLS initial estimators. An LTS initial estimator is admissible in the construction,
but our numerical LTS entry is the integrated sparse-LTS comparator rather
than a distilled LTS fit. We also make the implementation failure-aware, so
that a failed initial fit falls back along a prespecified cascade of
alternative initial fits. The cascade and its scope are described in
\ORsec{supp:impl-cascade}.

Because contaminated rows survive into the second-stage design, we also
define \emph{weighted distillation}, an exploratory variant that reweights
the quadratic term and the GIC residual sum of squares by MCD-based
leverage weights. Its definition is in
\ORsec{supp:impl-weighted}.

\subsection{Relation to existing estimators}\label{sec:relation}

With the empirical Gram matrix as its metric, \eqref{eq:step2} lies in the
quadratic-surrogate family represented by LSA, Q-SCAD, and APE
\citep{wangleng07,kwonchoikim11,leekimkwon12}. Here, however, the quadratic loss is an exact fidelity
criterion for the initial estimator's fitted values, not a Taylor claim about the
initial estimator's original robust objective.

In addition, distillation is
conceptually different from robust adaptive lasso schemes. Those retain the
observed response inside a robust residual loss and use a preliminary fit to
construct adaptive penalty \emph{weights}. Distillation instead replaces the
second-stage response by the initial estimator's fitted values and leaves the penalty
unweighted. As a consequence, contamination enters the two constructions through
different maps: in the integrated schemes it acts through a bounded residual
loss and data-dependent weights, whereas here it acts through $\btil$ and
the tuning scale.

Relative to sparse LTS,
MM-lasso and (adaptive) PENSE, distillation trades some generality for
modularity. It cannot recover predictive information absent
from the initial estimator's fitted surface, and the sparse compression step can
itself add finite-sample loss.
In return, the method can convert any fitted linear initial estimator with finite coefficients into
a sparse distilled estimator without
repeating the robust fit at every tuning value. The remaining computation
is one SCAD coordinate-descent path with the safeguards of
Section~\ref{sec:algo}.\footnote{The name follows the
machine-learning usage of \emph{distillation} for training a compact student on teacher outputs \citep{hinton15}. Here both models are linear, and the focus is robust sparse estimation
and the assessment of the resulting explanation.}

We use \emph{explanation} in a deliberately model-specific sense. For a
robust linear initial fit, the distilled estimator is a single sparse global linear approximation
over the empirical design distribution. After documented training-only
scaling, its selected variables, coefficient signs and magnitudes, and fitted
surface summarize which recorded coordinates reproduce the initial estimator's
predictions and in which linear directions.
These quantities are descriptive summaries of that approximation, not
causal effects or model-free variable importances. At the same time, in
correlated designs different sparse supports may approximate nearly the same
fitted surface. We therefore report
high predictive fidelity without support and coefficient stability
separately, rather than interpreting it as coordinate-level explanatory
evidence \citep{tolosilengauer11}.

\subsection{Algorithm, computation and defaults}\label{sec:algo}

Let $\mathbf Z=[\bm 1,\X]$, let $f_0=0$ and $f_j>0$ denote the
penalty factors for the slopes, and define
\[
  G=\mathbf Z^\top\mathbf Z/n,\qquad
  \bm c=\mathbf Z^\top\tilde\y/n.
\]
For a current coefficient vector $\bm b$, the conditional objective for
coordinate $j$ is, up to an additive constant,
\begin{equation}\label{eq:scalar-scad}
  q_j(b)=\frac{v_j}{2}b^2-\zeta_j b+
  p_{\ell_j}(|b|),\qquad
  v_j=G_{jj},\quad
  \zeta_j=c_j-(G\bm b)_j+v_jb_j,\quad
  \ell_j=\lambda f_j .
\end{equation}
Unlike a unit-column-norm threshold, this expression retains the actual
$v_j>0$. For $\ell_j>0$, we evaluate \eqref{eq:scalar-scad} at the
current coefficient, at the three SCAD boundaries, and at every feasible
stationary point. Every coordinate update therefore globally minimizes its
actual one-dimensional conditional SCAD problem. The exact candidate set is listed in
\ORsec{supp:impl-candidates}.

The path starts from the intercept-only least-squares state. We construct
$\lambda_{\max}$ so that this state is a global minimum of every scalar
coordinate subproblem at the first grid point, and the path then follows a
decreasing geometric grid, which we call the null grid. At each grid point
the algorithm runs Gram-matrix Gauss--Seidel sweeps with warm starts and
applies a two-part convergence and full-objective acceptance test, whose
exact form is given in \ORsec{supp:impl-defaults}.
For a retained state $(\hat\alpha_\lambda,\hat\bbeta_\lambda)$, write
$\RSS_d(\lambda)=\|\tilde\y-\hat\alpha_\lambda\bm1-
\X\hat\bbeta_\lambda\|_2^2$ for its distillation-fidelity residual sum of squares.

\refstepcounter{algorithm}\label{alg:distill}
\begin{center}\small
\fbox{\begin{minipage}{0.93\textwidth}
\textbf{Algorithm \thealgorithm} (Penalized distillation).\\
\textbf{Input}: prepared training data $(\X,\y)$; initial rule $\mathcal T$;
SCAD constant $a$; path length and endpoint ratio; GIC multiplier $\kappa$.\\
1. Compute $(\tilde\alpha,\btil)$ and $\hat\sigma$ from
\eqref{eq:step1}--\eqref{eq:sigma}, and set
$\tilde\y=\tilde\alpha\bm1+\X\btil$.\\
2. Construct the null grid above. At each grid point run the Gram
Gauss--Seidel solver, which retains the actual $v_j$, from the preceding
retained state, and apply the convergence and full-objective safeguard.\\
3. Select
$\hat\lambda=\argmin_{\lambda\in\Lambda}
 \{\RSS_d(\lambda)/\max(\hat\sigma^2,10^{-12})
 +\kappa\log(n)\mathrm{df}(\lambda)\}$,
where $\mathrm{df}$ includes the unpenalized intercept.\\
\textbf{Output}: the selected retained state and, for every grid point,
the coefficients, iterations, fixed-point gap, start/candidate/retained
objectives, acceptance indicator, and failure reason.
\end{minipage}}
\end{center}

Step 1 of Algorithm~\ref{alg:distill} is paid once, and Step 2 reuses the Gram
matrix and warm starts. Step 3 selects by a GIC whose residual term is
$\RSS_d(\lambda)$, referred to below as the fidelity GIC. All reported
unweighted distilled fits set $f_0=0$ and $f_j=1$ for every slope, and
\ORtab{tab:impl-defaults} collects the implementation defaults. Unless stated otherwise, every reported fit uses $\kappa=2$, a
100-point grid, and a lower endpoint of $10^{-3}\lambda_{\max}$
($0.05\lambda_{\max}$ when $n\le p+1$). The
implemented GIC and Bayesian information criterion (BIC) degrees of freedom
count the intercept and the slopes exceeding the numerical threshold of
\ORtab{tab:impl-defaults}, and displayed support metrics use the
prespecified numerical-zero threshold there.
Proposition~\ref{prop:gic} is an exact-arithmetic statement with degrees of
freedom equal to the mathematical support size, so the two numerical
thresholds play no part in that proof and identify the theoretical support
only under coefficient separation.

Cross-validation on $\tilde\y$ evaluates fidelity to the initial estimator's fitted
surface rather than prediction of the original response. Because this target
already lies in the span of the full second-stage design, its empirical error curve
can be nearly flat and can favor the dense, weakly penalized end of the path.
We use the fidelity GIC because it places an explicit price on model size while
retaining the contamination separation of the second stage.

\section{Theoretical properties}\label{sec:theory}

Every guarantee in this section attaches to a state that
Algorithm~\ref{alg:distill} actually returns, or to the explicitly
constructed oracle branch it approximates.
Throughout, $\y$ enters \eqref{eq:step2} only through
$\X\btil$, which converts every robustness question about the retained
$\bcomp$ into a question about $\btil$ plus a deterministic argument.
The distillation response is \emph{noiseless only in a
computational sense}. Conditional on the estimated initial fit there is
no new independent response error, although $\X\btil$ retains the initial
estimator's sampling error. The oracle model can therefore fit it with an
$O_p(1)$ residual sum of squares instead of the $O_p(n)$ floor of raw
penalized regression. All proofs are in \proofref{app:proofs3}.

\subsection{Assumptions}\label{sec:assum}

\begin{assumption}\label{ass:1}
(A1) $\X^\top\X/n\to M\succ0$ with bounded entries.
\end{assumption}
This is the standard fixed-$p$ design condition, as in
\citet{fanli01,wangleng07}. We work with fixed $p$
throughout, as does the oracle theory of \citet{smucler17}; the
diverging-$p$ regime is discussed in Section~\ref{sec:disc}. Because the
intercept has been profiled out, $\X$ here is the column-centered design.

\begin{assumption}\label{ass:2}
(A2) The target $\bbeta_0$ and its support $\cA$ do not vary with $n$,
$0<|\cA|<p$, $\min_{j\in\cA}|\beta_{0j}|>0$, and
$\sqrt n(\btil-\bbeta_0)=O_p(1)$.
\end{assumption}
This is the principal rate requirement on the initial estimator.
Specifically, the particular examples carry their usual model, design, and
moment conditions,
under which the root-$n$ rate component is satisfied by M-, S- and
MM-estimators \citep{yohai87}, by LTS, and by quantile regression. Adaptive
PENSE requires $\sqrt n$-consistency of a preliminary estimator in the
same way \citep{kepplinger23}. Unlike MM-lasso \citep{smucler17}, we do
not require a \emph{consistent} residual-scale estimate, because no scale
enters the estimation criterion \eqref{eq:step2} at all. A scale does
enter the default tuning rule, through $\hat\sigma$ in
\eqref{eq:sigma}, but only as a normalizing constant that fixes the
relative units of the two terms in the criterion of
Section~\ref{sec:algo}. Proposition~\ref{prop:gic} shows that selection
consistency survives any $\hat\sigma$ bounded away from zero and infinity
in probability. Non-smooth initials such as LTS are therefore
admissible.

\begin{assumption}\label{ass:3}
(A3) $\lambda_n\to0$ and $\sqrt n\lambda_n\to\infty$.
\end{assumption}
The usual SCAD rate condition \citep{fanli01}.

\subsection{Finite-sample robustness}\label{sec:finite}

For an estimator rule $T$ and a sample $\mathcal D_n$ of size $n$, let
$\mathfrak C_m(\mathcal D_n)$ contain the samples that differ from
$\mathcal D_n$ in at most $m$ complete $(\bm x_i,y_i)$ pairs. The ordinary
casewise replacement breakdown point \citep{donoho83} is
\[
 \varepsilon^*_{n,\mathrm{case}}(T;\mathcal D_n)
 =\min\left\{\frac mn:
   \sup_{\mathcal D'_n\in\mathfrak C_m(\mathcal D_n)}
   \|T(\mathcal D'_n)\|_2=\infty\right\}.
\]
For a fixed design
$\X$ and observed response vector $\y$, we define instead
\[
\begin{aligned}
 \mathcal Y_m(\y)
   &=\{\y':\#\{i:y'_i\ne y_i\}\le m\},\\
 \varepsilon^*_{n,y}(T;\y\mid\X)
   &=\min\left\{\frac mn:
     \sup_{\y'\in\mathcal Y_m(\y)}\|T(\X,\y')\|_2=\infty\right\}.
\end{aligned}
\]
The conditioning bar records that $\X$ is held fixed and uncontaminated,
so $\varepsilon^*_{n,y}$ is a response-only conditional breakdown
point and not shorthand for $\varepsilon^*_{n,\mathrm{case}}$. If the defining
set is empty (the rule remains bounded under arbitrary replacement of
all $n$ responses), we set the breakdown point to $(n+1)/n$ by
convention, and the same convention applies to
$\varepsilon^*_{n,\mathrm{case}}$. The corollary
below compares initial fit and distilled estimator under this conditional
response-replacement metric.

For this subsection only, we retain the explicit intercept. Write
$\mathbf Z=[\bm1,\X]$, $\tilde\theta=(\tilde\alpha,\btil^\top)^\top$,
$\theta=(\alpha,\bbeta^\top)^\top$, and
\[
Q_\lambda^+(\theta)=\frac1{2n}\|\mathbf Z\tilde\theta-
\mathbf Z\theta\|^2+\sum_{j=1}^p p_\lambda(|\beta_j|).
\]
Thus the intercept is present in the objective but is not
penalized. Let $\theta_0=(\bar{\tilde y},\bm0^\top)^\top$ denote the
intercept-only least-squares fit to the initial fitted surface.

\begin{proposition}[Deterministic fidelity and norm bounds]\label{prop:norm}
Let $s_{\min}$ and $s_{\max}$ denote the smallest and largest eigenvalues
of $\mathbf Z^\top\mathbf Z/n$, with $s_{\min}>0$.
\begin{enumerate}[label=(\alph*),leftmargin=2em,itemsep=0pt]
\item Any $\hat\theta$ with
$Q_\lambda^+(\hat\theta)\le Q_\lambda^+(\tilde\theta)$ satisfies
\[
n^{-1/2}\|\mathbf Z(\hat\theta-\tilde\theta)\|_2
\le\lambda\sqrt{p(a+1)},\qquad
\|\hat\theta-\tilde\theta\|_2
\le\lambda\sqrt{p(a+1)/s_{\min}}.
\]
\item Any $\hat\theta$ with
$Q_\lambda^+(\hat\theta)\le Q_\lambda^+(\bm 0)$ satisfies
$\|\hat\theta\|_2\le2\sqrt{s_{\max}/s_{\min}}\,
\|\tilde\theta\|_2$.
\end{enumerate}
\end{proposition}

The first inequality in part (a) is an empirical-design fidelity certificate:
it directly bounds the root-mean-square discrepancy between a qualifying
candidate state and its initial fit. Even so, the part-(a) sublevel
hypothesis, unlike the null-model condition in part (b), is not automatic
along the safeguarded path and must be checked.

\begin{lemma}[Safeguarded computed paths qualify]\label{lem:warm}
$p_\lambda(t)$ is nondecreasing in $\lambda$ for each fixed $t$. Consider
a decreasing grid beginning at $\theta_0$ and a path that
accepts a candidate only when its current-$\lambda$ objective does not
exceed the warm start's, otherwise retaining that start. Every retained
path state satisfies
$Q_\lambda^+(\hat\theta)\le Q_\lambda^+(\theta_0)\le
Q_\lambda^+(\bm0)$ and hence the hypothesis of
Proposition~\ref{prop:norm}(b), by induction along the grid.
\end{lemma}

\begin{corollary}[Conditional response-breakdown inheritance]\label{cor:bp}
Fix $\X$ with $\mathbf Z=[\bm1,\X]$ of full column rank. Let
$\hat\theta_k$ denote the estimator rule returning the retained state at
grid position $k$, and let $\hat\theta_{\mathrm{sel}}$ be any possibly
data-dependent rule selecting among the finite set of retained states. Then
\[
 \varepsilon^*_{n,y}(\hat\theta_k;\y\mid\X)
 \ge \varepsilon^*_{n,y}(\tilde\theta;\y\mid\X)
 \quad\text{for every }k,\qquad
 \varepsilon^*_{n,y}(\hat\theta_{\mathrm{sel}};\y\mid\X)
 \ge \varepsilon^*_{n,y}(\tilde\theta;\y\mid\X).
\]
\end{corollary}

The proofs are elementary and are given in \proofref{app:norm}.
Corollary~\ref{cor:bp} says that penalization cannot destroy what the
initial estimator preserved under the specified contamination
neighborhood: the distilled estimator inherits the initial estimator's
conditional response-replacement lower bound without additional loss.
Robust tuning is
therefore needed for efficiency and selection quality, not for the
conditional response-only breakdown itself. In methods tuned by a
contaminated cross-validation criterion, by contrast, the robustness can
be lost at the tuning stage \citep[a caveat stressed by][]{kepplinger23}.

\subsection{Oracle-support local branch and conditional efficiency}\label{sec:asymp}

\begin{theorem}[Oracle-support local branch]\label{thm:support}
Under Assumptions \ref{ass:1}--\ref{ass:3}, with probability tending to
one there is a strict local minimizer
$(\alpha^{\mathrm{ora}},\bora)$ of $Q^+_{\lambda_n}$ in
\eqref{eq:step2}, where
$\bora_\cZ=\bm0$,
$\bora_\cA=(\X_\cA^\top\X_\cA)^{-1}\X_\cA^\top\X\btil$, and
$\alpha^{\mathrm{ora}}=\bar{\tilde y}-\bar{\bm x}^{\top}\bora$.
Here $\bar{\bm x}$ is the mean of the original, pre-centering covariates.
\end{theorem}

\begin{theorem}[Oracle local-branch limit and conditional efficiency]\label{thm:oracle}
For the oracle-support local minimizer in
Theorem~\ref{thm:support}, if
$\sqrt n(\btil-\bbeta_0)\xrightarrow{d}N(0,\tilde\Sigma)$, then
$\sqrt n(\bora_\cA-\bbeta_{0\cA})\xrightarrow{d}N(0,T\tilde\Sigma T^\top)$
with $T=M_{\cA\cA}^{-1}M_{\cA\cdot}$. If in addition
$\tilde\Sigma=c_\rho M^{-1}$ for a scalar $c_\rho>0$, then
$T\tilde\Sigma T^\top=c_\rho M_{\cA\cA}^{-1}$. For initial-estimator classes whose
true-submodel refit has the same proportional-covariance form with the same
scalar $c_\rho$, this equals that oracle refit's asymptotic variance. Moreover,
$(M^{-1})_{\cA\cA}\succeq M_{\cA\cA}^{-1}$ with equality if and only if
$M_{\cA\cZ}=0$.
\end{theorem}

The limit itself is immediate once Theorem~\ref{thm:support} is
available; the proof (\proofref{app:oracle}) adds the truncation comparison
and an exact finite-sample identity for the OLS initial. In the final
display, $c_\rho(M^{-1})_{\cA\cA}$ is the limiting variance of the
truncated subvector $\btil_\cA$, so the projection improves on truncation
unless $M_{\cA\cZ}=0$.
The oracle-branch diagnostic in \ORsub{sec:sim-oracle} checks these
branch-level implications numerically. The condition
$\tilde\Sigma=c_\rho M^{-1}$ is the familiar one under which
M-estimators have a scalar-times-$M^{-1}$ sandwich, and it holds for M-,
S- and MM-estimators when the errors are independent of the covariates.
It fails for GM-estimators that downweight leverage and under
heteroscedasticity, where $T\tilde\Sigma T^\top$ remains valid but is no
longer the oracle variance. The identity of Theorem~\ref{thm:if} is
likewise a property of the branch functional as defined, and its
stability hypothesis only makes that functional a faithful local
description of the penalized estimator.

\begin{theorem}[Asymptotic equivalence with LSA]\label{thm:equiv}
Assume \ref{ass:1}--\ref{ass:3}, and let
  $\hat\Sigma$ estimate the sampling covariance of $\btil$, be symmetric
  positive definite with probability tending to one, and satisfy
$n\hat\Sigma\xrightarrow{p}c_\rho M^{-1}$ for a scalar $c_\rho>0$.
Suppose the SCAD--LSA
criterion of \citet{wangleng07}, with weight $\hat\Sigma^{-1}$, has an
oracle-support local branch whose active coefficients lie in the flat
region of SCAD. Then that branch and the oracle local branch of
\eqref{eq:step2} have the same active set with probability tending to one
and satisfy
$\bLSA_\cA-\bora_\cA=o_p(n^{-1/2})$.
\end{theorem}

The theorem explicitly assumes consistency and positive definiteness of
$\hat\Sigma$, whereas distillation does not estimate that matrix. Conditional
on the two flat-active oracle branches, the argument in
\proofref{app:equiv} needs convergence of the weight but no additional
rate: the leading term of the projection difference cancels.

\subsection{Influence function}\label{sec:if}

We have stated the preceding results for a fixed-design sequence. For
influence analysis only, we switch to a random-design functional $F$ with
the intercept profiled, so the population design matrix is
$M(F)=\operatorname{Cov}_F(\bm x)$. A global influence function for the
nonsmooth penalized selector is not the right object here. For this reason
we fix a neighbourhood in which the active set and the flat-active SCAD
status do not change. In that neighbourhood the active branch is the smooth
functional
$b_{\cA}(F)=M(F)_{\cA\cA}^{-1}M(F)_{\cA\mathbin{\cdot}}\btil(F)$,
with $b_{\cZ}(F)=0$. We define its influence function as the ordinary
G\^ateaux derivative along
$F_\varepsilon=(1-\varepsilon)F_0+\varepsilon\delta_z$. Support changes of
the global selector lie outside
it, and, as \citet{avella17} emphasizes, the influence function of a
penalized estimator depends on the penalty and differs between zero and
nonzero coordinates.

\begin{theorem}[Active-branch influence identity]\label{thm:if}
Let $F_0$ have finite second moments, satisfy
$E_{F_0}(y\mid\bm x)=\alpha_0+\bm x^\top\bbeta_0$, and obey
$M(F_0)=\operatorname{Cov}_{F_0}(\bm x)\succ0$. Define
$\cA=\operatorname{supp}(\bbeta_0)$, and let $\btil$ be Fisher consistent at
$F_0$. Consider the active-branch functional defined above.
If the active set and flat-active-penalty status defining that branch are
stable in a neighbourhood of $F_0$, then for every contamination point
$z=(\bm x_0,y_0)$ at which the influence function of the initial estimator exists,
$\mathrm{IF}(z;b_\cA,F_0)=T_0\mathrm{IF}(z;\btil,F_0)$, where
$T_0=M(F_0)_{\cA\cA}^{-1}M(F_0)_{\cA\mathbin{\cdot}}$ and the left-hand
  side is the G\^ateaux influence function of that support- and
  flat-active-status-stable active-branch functional.
\end{theorem}

The proof shows that the term arising from perturbation of the design
second moments vanishes identically at the model, because it is
proportional to the distillation residual
$\bm u_{0\cZ}^\top\bbeta_{0\cZ}=0$, where
$\bm u_0=\bm x_0-E_{F_0}(\bm x)$. Consequently, on any contamination
class over which the initial estimator's influence function is bounded, the
influence function of the active branch is bounded as well, with gross-error
sensitivity inflated by at most the induced operator norm $\|T_0\|$.

\subsection{Selection conditional on computed-path inclusion}\label{sec:gic}

Theory for penalized estimators is usually stated for a deterministic
$\lambda_n$ satisfying (A3). In practice, however, the choice is
data-driven. We give a conditional result for the GIC used in
Algorithm~\ref{alg:distill}.

\begin{proposition}[Selection consistency conditional on path inclusion]\label{prop:gic}
Under Assumptions~\ref{ass:1} and \ref{ass:2}, suppose there exists a
retained, possibly data-dependent path index $\lambda^\circ$ such that
$P\{\mathrm{supp}\,\bcomp(\lambda^\circ)=\cA\}\to1$ and
$\RSS_d(\lambda^\circ)=O_p(1)$. Let
$0<c\le\hat\sigma\le C<\infty$ with probability tending to one, and let
the multiplier $a_n=\kappa\log n$ satisfy $a_n\to\infty$ and $a_n=o(n)$.
Then the GIC minimizer over that candidate path satisfies
$P\{\mathrm{supp}\,\bcomp(\hat\lambda)=\cA\}\to1$.
\end{proposition}

The form of the criterion reflects this structure. With the observed
response, classical BIC-type selectors use $n\log\{\RSS(\lambda)/n\}$,
which puts residual improvements on a likelihood scale that grows like $n$.
The distillation response is the initial estimator's fitted surface, so the
best fit on any support containing $\cA$ leaves a distillation residual sum
of squares that is $O_p(1)$ rather than of order $n$. A logarithmic
transform would then work in the wrong direction: it would magnify the
negligible $O_p(1)$ fidelity differences among supports that already contain
$\cA$ (differences the selector must discount) into criterion gaps of
order $n$ that no $\log(n)\,\mathrm{df}$ penalty can offset. In addition,
it diverges to $-\infty$ as $\RSS_d(\lambda)\to0$ at the weakly penalized
end of the path. The criterion therefore keeps
the fidelity term linear, normalized only by the operational scale
$\hat\sigma^2$ that fixes its units against the penalty
$\kappa\log(n)\,\mathrm{df}$.
The weak-signal study in
\ORsub{sec:sim-tuning} therefore treats $\kappa$ as a finite-sample
trade-off, conditional on the qualifying-path requirement.

\begin{remark}[From support to coefficients]\label{rem:selected}
Proposition~\ref{prop:gic} deliberately stops at the support. If the selected
state has support $\cA$, is stationary at its selected $\hat\lambda$, and
satisfies $|\bcomp_j(\hat\lambda)|>a\hat\lambda$ for every $j\in\cA$,
its active stationarity equations give the projection in
Theorem~\ref{thm:support}. For numerical coefficients, define the normalized
active stationarity error
$\bm r_n=n^{-1}\X_\cA^\top\{\X\bcomp(\hat\lambda)-\X\btil\}$.
On the selected-support event, the coefficient gap from that projection is
$(\X_\cA^\top\X_\cA/n)^{-1}\bm r_n$.
Under (A1) and the initial-estimator asymptotic normality of
Theorem~\ref{thm:oracle}, the displayed limit transfers to the reported
active coefficients if the support and flat-region events have probability
tending to one and $\|\bm r_n\|=o_p(n^{-1/2})$.
These are additional conditions, not conclusions of Proposition~\ref{prop:gic}:
(A3) concerns a deterministic sequence and does not establish the flat-region
event at $\hat\lambda$. Nor does the fixed numerical tolerance establish the
required error rate. A warm state retained after rejection is not certified
as stationary at the current $\lambda$. The branch-gap study in
\ORsub{sec:sim-oracle} is therefore a numerical diagnostic, not a proof
of an oracle limit for the reported coefficients.
\end{remark}

In Table~\ref{tab:theory-scope} we summarize what each result guarantees,
the state or branch it applies to, and its key conditions. The last
column records what each result leaves uncovered.

\begin{table}[!htbp]
\caption{Scope of the guarantees of Section~\ref{sec:theory}. Each row
records the object a result applies to, the key conditions it uses, and
what it does not cover. The finite-sample statements hold for every fixed
$\lambda>0$; the asymptotic statements use the conditions listed in their
rows, and in particular Proposition~\ref{prop:gic} does not use (A3).
The screened-fit transfer is stated in Corollary~\ref{cor:scr-modular};
it is a fixed-$p$ statement for the unweighted estimator and is silent
on finite-sample behaviour at moderate $p/n$.}
\label{tab:theory-scope}
\centering
\scriptsize
\setlength{\tabcolsep}{3.5pt}
\renewcommand{\arraystretch}{1.08}
\begin{tabular}{@{}p{0.27\textwidth}p{0.20\textwidth}p{0.22\textwidth}p{0.23\textwidth}@{}}
\toprule
Guarantee & Applies to & Key conditions & Not covered \\
\midrule
Fidelity and norm bounds; conditional response-breakdown inheritance
(Prop.~\ref{prop:norm}, Lem.~\ref{lem:warm}, Cor.~\ref{cor:bp}) &
every retained path state and any rule selecting among them &
fixed $\lambda>0$; $\mathbf Z$ of full column rank; safeguarded path (the
part-(a) sublevel condition is checked separately) &
casewise or contaminated-design replacement; accuracy under
contamination; robustness of selection \\
Oracle-support local branch exists (Thm.~\ref{thm:support}) &
oracle local branch of \eqref{eq:step2} &
(A1)--(A3) &
computed-path inclusion; global optimality \\
Oracle local-branch limit and conditional efficiency
(Thm.~\ref{thm:oracle}) &
oracle local branch &
(A1)--(A3); asymptotic normality of $\btil$;
$\tilde\Sigma=c_\rho M^{-1}$ for the oracle-variance claim &
GM-type initials and heteroscedasticity, where the limit remains valid
but is no longer the oracle variance \\
Asymptotic equivalence with LSA (Thm.~\ref{thm:equiv}) &
the two flat-active oracle branches &
(A1)--(A3); positive-definite $\hat\Sigma$ with
$n\hat\Sigma\to_pc_\rho M^{-1}$ (proportional form); sufficient
condition only &
contaminated regimes; necessity of the condition; the weighted variant \\
Active-branch influence identity (Thm.~\ref{thm:if}) &
support- and flat-active-stable branch at $F_0$ &
model $F_0$; Fisher-consistent $\btil$; influence function of $\btil$
exists at $z$ &
support changes of the global selector; leverage growth away from the
model \\
GIC selection consistency (Prop.~\ref{prop:gic}) &
GIC minimizer over the computed candidate path &
(A1)--(A2); qualifying path state with $O_p(1)$ residual sum of squares;
$\hat\sigma$ bounded above and below; $a_n=\kappa\log n$ &
existence of the qualifying state; $\hat\sigma$ bounds under every
contamination scheme; selected-coefficient limit without the additional
conditions of Remark~\ref{rem:selected} \\
\bottomrule
\end{tabular}
\end{table}

\section{Monte Carlo evidence}\label{sec:sim}

The Monte Carlo study in this section measures accuracy and selection,
which the boundedness result does not imply. We report the base study
first and then the comparison with modern robust-sparse procedures on the
same design. Companion studies, reported in \ORsub{sec:sim-oracle} and the two
subsections that follow it, check the projection variance of
Theorem~\ref{thm:oracle}, the computed-path counterpart of
Theorem~\ref{thm:equiv}, and the GIC multiplier of
Proposition~\ref{prop:gic}.

\subsection{Design of the study}\label{sec:sim-design}

Throughout, $\bbeta_0=(3,1.5,0,0,2,0,\dots,0)$ with $p=12$ unless stated
otherwise, covariates are drawn from an AR(1) design with correlation
$0.5$, and errors are normal, $t_3$, or centred $\chi^2_5$. We include the
last law to test performance under asymmetric errors and the importance of
the common unpenalized intercept. Contamination, where present, shifts a
uniformly sampled fixed-size subset of responses by $+8$. Normal errors
have unit variance, the $t_3$ draws are on their raw scale (variance $3$),
and the centred $\chi^2_5$ errors are divided by $\sqrt{10}$ to have unit
variance.

Accuracy is measured by $\mathrm{MSE}=(\widehat{\bbeta}-\bbeta_0)^\top\Sigma
(\widehat{\bbeta}-\bbeta_0)$ and selection by the exact-recovery rate, the
proportion of replications in which the estimated support equals $\cA$
exactly. We report the distilled estimator with three initial fits, OLS,
Huber and MM, written D-OLS, D-Huber and D-MM, against raw-data SCAD and,
as a benchmark, least squares on the true support, which we call the LS
oracle. In addition, a separate paired experiment compares each distilled
path directly with its covariance-weighted SCAD--LSA counterpart. The
empirical initial-estimator comparison is limited to OLS, Huber, and MM;
quantile-regression and distilled LTS initial estimators remain future
work.

Simulation configurations use 100--500 replications, with exact counts in
the table captions, and all estimators within a configuration are computed
on the same samples. A targeted paired path audit crossed grid density,
endpoint depth, and warm-start policy. Every controlled comparison in that
audit selected the same support, and its design and numbers are in
\provref{}.

All computations were carried out in R 4.4.2. The MM initial fit is
computed with \texttt{rlm(method = "MM")} from the \texttt{MASS} package
\citep{venablesripley02}, the Huber initial fit uses the same routine with
a Huber $\psi$-function, RLARS-MM uses \texttt{robustHD} \citep{alfons21},
and adaptive PENSE uses \texttt{pense} 2.5.2. All timings in
Table~\ref{tab:modern} are wall-clock seconds per fit on an Intel Core
i9-12900KF workstation with 16 cores and 64~GB of memory. For the
modern-comparator production run, the numerical thread environment
variables and the \texttt{robustHD} backend budget were set to 14,
whereas the adaptive-PENSE run used a one-thread setting. These are
configured backend budgets, not measurements of actual thread utilization.
The seconds column therefore compares complete workflows under their
recorded settings, rather than an equal-thread benchmark.

\subsection{Robustness transfer and its cost}\label{sec:sim-main}

\begin{figure}[!htbp]
\centering
\includegraphics[width=\linewidth]{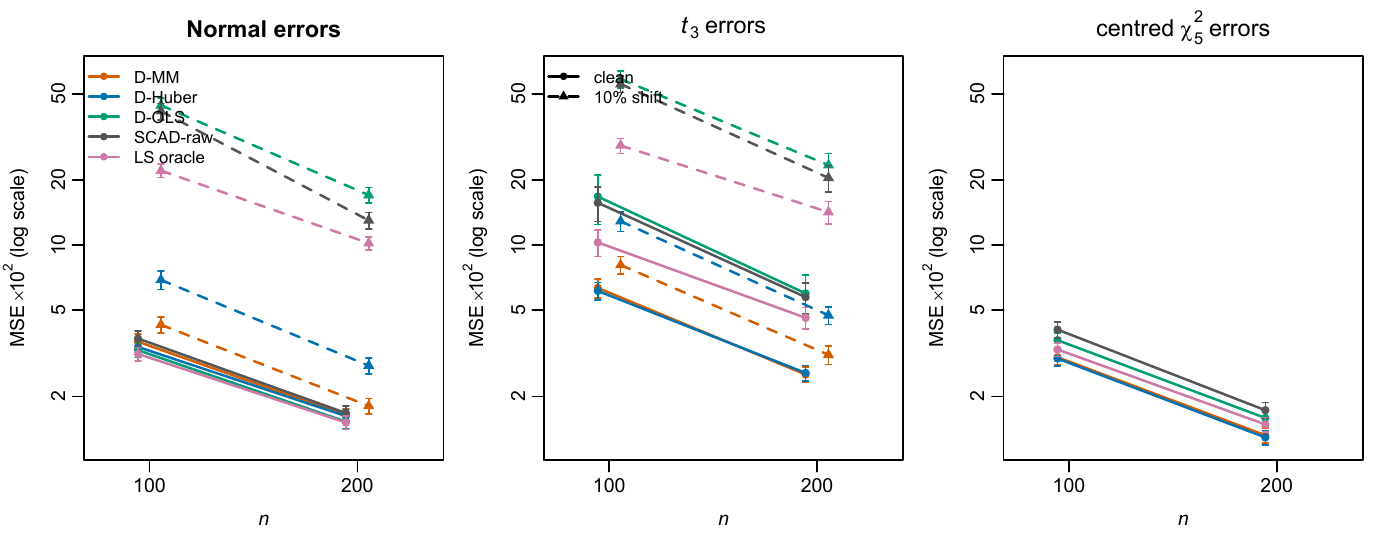}
\caption{Base study, 500 replications per cell. Each panel shows one error
law; solid lines are clean training, dashed lines the fixed-size $10\%$
response shift of $+8$, and bars are pointwise 95\% Monte Carlo intervals
formed from the archived Monte Carlo standard errors. MSE is multiplied by
$10^{2}$, matching \ORtab{tab:main}, and is on a logarithmic scale; the
centred $\chi^2_5$ design is clean-only.}
\label{fig:sim-contamination}
\end{figure}

Figure~\ref{fig:sim-contamination} displays the full base study, with five
estimators, both sample sizes, three error laws, and both contamination
levels, and \ORtab{tab:main} reports the numbers. All MSE values quoted in
this subsection are on the $10^{2}$ scale of that table.

Three findings stand out. First, as expected, under $10\%$ contamination a
robust initial fit changes both accuracy and selection. At $n=100$ under
normal errors, D-MM records MSE $4.27$ versus $39.77$ for SCAD-raw, with
exact recovery $0.964$ versus $0.654$, and Table~\ref{tab:main} shows the
same pattern under $t_3$ errors and at $n=200$. Second, D-OLS records MSE
$44.07$ where D-MM attains $4.27$ in the contaminated normal row.
Distillation compresses whatever its initial fit believes; when the initial fit has been misled, the distilled estimator inherits the error.
Third, in the clean normal rows D-MM and SCAD-raw are close, with $3.57$
versus $3.73$ at $n=100$ and $1.63$ versus $1.68$ at $n=200$, although
equal risks are not established; the paired Monte Carlo intervals are in
\provref{}.

Turning to the clean $t_3$ errors, D-MM ($6.31$) outperforms not only
SCAD-raw ($15.79$) but the LS oracle ($10.29$), an estimator that is told
which variables matter. The oracle is a least-squares fit, and under $t_3$
errors least squares is inefficient, so the efficiency advantage of the
robust initial exceeds the information advantage of knowing the support.
By contrast, under the asymmetric centred $\chi^2_5$ errors, D-Huber leads
with an MSE of $3.01$ versus $4.15$ for SCAD-raw at $n=100$,
and the two $\chi^2_5$ rows of Table~\ref{tab:main} show the same ordering
at $n=200$. Overall, robustness transfers from the initial fit to the distilled
estimator, at no discernible clean-normal cost at this Monte Carlo
resolution.

\begin{table}[!htbp]\centering\tiny
\caption{Base study, 500 replications per row. Designs have $p=12$,
$\bbeta_0=(3,1.5,0,0,2,0,\ldots,0)$, and AR(1) correlation $0.5$; $n$ and
the error law are shown by row. Normal and centred $\chi_5^2/\sqrt{10}$
errors have unit variance, whereas raw $t_3$ errors have variance 3.
Contamination adds 8 to a uniformly sampled fixed-size 10\% response subset.
Each cell gives MSE (Monte Carlo standard error) on the first line and
exact-support recovery (binomial Monte Carlo standard error) on the
second; MSE entries and their standard errors are multiplied by $10^{2}$.
All SCAD fits use $a=3.7$ and a 100-point path; the
distilled path uses GIC $\kappa=2$ and SCAD-raw uses BIC. Boldface marks
the lowest non-oracle MSE in each row.}
\label{tab:main}
\setlength{\tabcolsep}{1.8pt}
\begin{tabular}{@{}lccccc@{}}
\toprule
Scenario & D-MM & D-Huber & D-OLS & SCAD-raw & oracle LS\\
\midrule
$n{=}100$, normal, 0\% &
\makecell{3.57 (0.15)\\0.974 (0.007)} &
\makecell{3.39 (0.13)\\0.980 (0.006)} &
\makecell{\textbf{3.27} (0.13)\\0.984 (0.006)} &
\makecell{3.73 (0.16)\\0.910 (0.013)} &
\makecell{3.14 (0.12)\\1.000 (0.000)}\\
$n{=}200$, normal, 0\% &
\makecell{1.63 (0.06)\\0.996 (0.003)} &
\makecell{1.62 (0.06)\\0.996 (0.003)} &
\makecell{\textbf{1.53} (0.06)\\0.998 (0.002)} &
\makecell{1.68 (0.07)\\0.948 (0.010)} &
\makecell{1.51 (0.05)\\1.000 (0.000)}\\
$n{=}100$, $t_3$, 0\% &
\makecell{6.31 (0.31)\\0.944 (0.010)} &
\makecell{\textbf{6.21} (0.31)\\0.952 (0.010)} &
\makecell{16.34 (2.16)\\0.868 (0.015)} &
\makecell{15.79 (1.55)\\0.862 (0.015)} &
\makecell{10.29 (0.74)\\1.000 (0.000)}\\
$n{=}200$, $t_3$, 0\% &
\makecell{\textbf{2.51} (0.10)\\0.976 (0.007)} &
\makecell{2.58 (0.11)\\0.974 (0.007)} &
\makecell{6.00 (0.66)\\0.920 (0.012)} &
\makecell{5.82 (0.48)\\0.916 (0.012)} &
\makecell{4.61 (0.27)\\1.000 (0.000)}\\
$n{=}100$, $\chi^2_5$, 0\% &
\makecell{3.03 (0.13)\\0.968 (0.008)} &
\makecell{\textbf{3.01} (0.12)\\0.970 (0.008)} &
\makecell{3.68 (0.15)\\0.956 (0.009)} &
\makecell{4.15 (0.18)\\0.870 (0.015)} &
\makecell{3.29 (0.12)\\1.000 (0.000)}\\
$n{=}200$, $\chi^2_5$, 0\% &
\makecell{1.32 (0.05)\\0.984 (0.006)} &
\makecell{\textbf{1.29} (0.05)\\0.988 (0.005)} &
\makecell{1.59 (0.07)\\0.972 (0.007)} &
\makecell{1.73 (0.07)\\0.918 (0.012)} &
\makecell{1.48 (0.06)\\1.000 (0.000)}\\
$n{=}100$, normal, 10\% &
\makecell{\textbf{4.27} (0.18)\\0.964 (0.008)} &
\makecell{6.96 (0.34)\\0.914 (0.013)} &
\makecell{44.07 (1.84)\\0.518 (0.022)} &
\makecell{39.77 (1.77)\\0.654 (0.021)} &
\makecell{22.08 (0.80)\\1.000 (0.000)}\\
$n{=}200$, normal, 10\% &
\makecell{\textbf{1.81} (0.08)\\0.994 (0.003)} &
\makecell{2.77 (0.12)\\0.980 (0.006)} &
\makecell{17.22 (0.73)\\0.692 (0.021)} &
\makecell{12.57 (0.52)\\0.894 (0.014)} &
\makecell{10.21 (0.37)\\1.000 (0.000)}\\
$n{=}100$, $t_3$, 10\% &
\makecell{\textbf{8.00} (0.38)\\0.950 (0.010)} &
\makecell{12.41 (0.60)\\0.900 (0.013)} &
\makecell{58.06 (2.71)\\0.492 (0.022)} &
\makecell{54.07 (2.46)\\0.578 (0.022)} &
\makecell{28.85 (1.17)\\1.000 (0.000)}\\
$n{=}200$, $t_3$, 10\% &
\makecell{\textbf{3.11} (0.16)\\0.988 (0.005)} &
\makecell{4.62 (0.20)\\0.968 (0.008)} &
\makecell{23.29 (1.57)\\0.698 (0.021)} &
\makecell{20.19 (1.50)\\0.848 (0.016)} &
\makecell{14.20 (0.85)\\1.000 (0.000)}\\
\bottomrule
\end{tabular}
\end{table}
{}
\FloatBarrier

\subsection{Does the compression improve efficiency?}\label{sec:sim-oracle}

Theorem~\ref{thm:oracle} concerns the oracle local-branch projection, not
the GIC-selected computed estimator. We therefore estimated both, together
with the true-submodel MM refit and the naive active sub-vector of the full
MM initial fit, under clean $t_3$ errors at
$n\in\{200,400,800\}$ with 500 replications, scaling by $n$ so that the
theoretical limits are constants. We also audited whether the computed
100-point path contained the oracle support and recorded the smallest
distillation RSS among such path states.

Across the design the branch/oracle variance ratios remain within about
$2.5\%$ of one, with the per-design values retained in the numerical
records described in the data and code availability statement. The naive
sub-vector is noticeably more variable for $\beta_2$ and $\beta_5$, which
is consistent with the cross-block Gram correction in
Theorem~\ref{thm:oracle}. In this design every computed path contained an
oracle-support state, and the median and 90th percentile of its minimum RSS
remain roughly stable as $n$ increases, so the selected D-MM ratios track
the branch ratios here. That supports the qualifying-path condition in this
design but proves nothing about the algorithm in general.

\subsection{Direct LSA comparison at the heteroscedastic boundary}
\label{sec:sim-lsa}

Theorem~\ref{thm:equiv} compares oracle local branches under a
proportional-covariance condition and says nothing about the GIC-selected
computed paths, so we compared the two computed procedures head to head.
In each of 500 paired replications we held fixed $n=200$, $p=12$, the base
coefficient vector, and an AR(1) design with correlation $0.5$. The homoscedastic response used $\epsilon_i=z_i$, whereas the
heteroscedastic response used
$\epsilon_i=s_i z_i$ with
$s_i=\exp(0.5x_{i1}-0.25)$, so $E(s_i^2)=1$. The two responses shared the
same $\X$ and standard-normal innovation $z$. For OLS, Huber, and MM
initial estimators, direct LSA used the inverse empirical sandwich estimate of
$n\operatorname{Var}(\btil)$. Every estimate was positive definite without
regularization. Both methods used 100-point safeguarded SCAD paths with
$a=3.7$ and the same $2\log(n)$ model-size multiplier.

\begin{table}[ht]\centering\scriptsize
\caption{Direct covariance-weighted LSA comparison, 500 paired replications
per regime ($n=200$, $p=12$, $\bbeta_0=(3,1.5,0,0,2,0,\ldots,0)$,
AR(1) correlation $0.5$, normal innovations).
The D and LSA columns give MSE (Monte Carlo standard error) on the first line
and exact-support recovery (binomial Monte Carlo standard error) on the
second; MSE entries, their standard errors, and the differences are
multiplied by $10^{2}$. Difference is paired
LSA-minus-distillation MSE with a pointwise 95\% Monte Carlo interval.}
\label{tab:lsa-hetero}
\begin{tabular}{@{}lccc@{}}
\toprule
Initial estimator & D & LSA & paired difference [95\% interval]\\
\midrule
\multicolumn{4}{@{}l}{\emph{Panel A: homoscedastic errors}}\\
\midrule
OLS &
\makecell{1.556 (0.055)\\0.994 (0.003)} &
\makecell{1.716 (0.062)\\0.986 (0.005)} &
$0.160$ [$0.103,0.216$]\\
Huber &
\makecell{1.662 (0.058)\\0.992 (0.004)} &
\makecell{1.881 (0.066)\\0.990 (0.004)} &
$0.219$ [$0.151,0.287$]\\
MM &
\makecell{1.671 (0.061)\\0.994 (0.003)} &
\makecell{1.889 (0.068)\\0.986 (0.005)} &
$0.218$ [$0.149,0.287$]\\
\midrule
\multicolumn{4}{@{}l}{\emph{Panel B: heteroscedastic errors}}\\
\midrule
OLS &
\makecell{2.117 (0.083)\\0.964 (0.008)} &
\makecell{1.802 (0.070)\\0.988 (0.005)} &
$-0.315$ [$-0.417,-0.213$]\\
Huber &
\makecell{1.393 (0.052)\\0.984 (0.006)} &
\makecell{1.384 (0.052)\\0.990 (0.004)} &
$-0.009$ [$-0.074,0.055$]\\
MM &
\makecell{1.477 (0.056)\\0.986 (0.005)} &
\makecell{1.475 (0.054)\\0.986 (0.005)} &
$-0.002$ [$-0.073,0.069$]\\
\bottomrule
\end{tabular}
\end{table}

As Table~\ref{tab:lsa-hetero} shows, the computed paths are close but not
identical. Under homoscedasticity the selected LSA path has larger MSE for
all three initial estimators. Under heteroscedasticity, however, LSA improves on
D-OLS by $0.315$, while the paired Huber and MM differences are centred
near zero with intervals covering zero. Overall, covariance weighting can matter when the
proportional form fails, but it does not uniformly improve the selected
finite-sample path.

\FloatBarrier

\subsection{Comparison with modern robust-sparse procedures}
\label{sec:sim-modern}

We next put the methods on 1000 common samples in three regimes. The first
is the homoscedastic base design, the second a heteroscedastic variant that
scales each error by $s_i=\exp(0.5x_{i1}-0.25)$, normalized so that
$E(s_i^2)=1$, and the third the homoscedastic design with a random
fixed-size $10\%$ subset of responses shifted by $+8$. In addition to D-MM,
SCAD-raw, five-fraction sparse LTS, and direct LSA-MM, we include robust
least-angle regression followed by an MM refit
\citep{khanetal07} (RLARS-MM) and the official
adaptive-PENSE implementation \citep{kepplinger23}.
RLARS-MM evaluates its returned five-model-size sequence by BIC before the
final MM refit, and our implementation of it uses
the \texttt{robustHD} package \citep{alfons21}.

We fit adaptive PENSE through the \texttt{pense} 2.5.2 function
\texttt{adapense\_cv} with five-fold robust information-sharing
cross-validation \citep{kepplingerwei26}, and the full configuration is in
\provref{}.
Its primary rule is the package-default robustness-weighted root mean squared
prediction-error (WRMSPE) selector, denoted AdaPENSE-WRMSPE. The package's
post-hoc $\tau$-size selector uses the
identical preliminary fit, adaptive loadings, final path, and folds. The
seconds column of Table~\ref{tab:modern} reports
end-to-end time for each method's prespecified complete workflow on common
data, not an equalized candidate or tuning budget, and a separate
equal-resource sensitivity remains open.

\begin{table}[htbp]\centering\scriptsize
\caption{Modern-comparator experiment on 1000 paired replications per
scenario with $n=200$, $p=12$, three nonzero slopes, and AR(1) correlation
$0.5$. The rows cover homoscedastic errors, heteroscedastic errors, and a
fixed-size $10\%$ vertical-contamination setting. MSE is followed by its Monte
Carlo standard error (MCSE), both multiplied by $10^{2}$; TPR/FPR and exact recovery are followed by their
replication-level Monte Carlo standard errors. Mean selected size and mean
end-to-end seconds are also shown. All coefficient vectors are finite;
diagnostic flags and complete tuning workflows are described in the text.}
\label{tab:modern}
\setlength{\tabcolsep}{3.3pt}
\begin{tabular}{@{}lccccc@{}}
\toprule
Method & MSE (MCSE) & TPR/FPR & exact & size & seconds\\
\midrule
\multicolumn{6}{@{}l}{\emph{Homoscedastic}}\\
D-MM             & \textbf{1.667} (0.044) & \makecell{1.000/0.001\\(0.000/0.000)} & 0.99 (0.003) & 3.01 & 0.10\\
LSA-MM           & 1.891 (0.050) & \makecell{1.000/0.002\\(0.000/0.000)} & 0.98 (0.004) & 3.02 & 0.10\\
RLARS-MM         & 1.913 (0.058) & \makecell{1.000/0.020\\(0.000/0.001)} & 0.82 (0.012) & 3.18 & 1.59\\
AdaPENSE-WRMSPE  & 4.536 (0.116) & \makecell{1.000/0.024\\(0.000/0.003)} & 0.90 (0.009) & 3.22 & 2.45\\
SCAD-raw         & 1.793 (0.054) & \makecell{1.000/0.012\\(0.000/0.001)} & 0.93 (0.008) & 3.11 & 0.07\\
sparse LTS       & 6.306 (0.128) & \makecell{1.000/0.242\\(0.000/0.005)} & 0.14 (0.011) & 5.17 & 0.18\\
\addlinespace
\multicolumn{6}{@{}l}{\emph{Heteroscedastic}}\\
D-MM             & \textbf{1.468} (0.039) & \makecell{1.000/0.002\\(0.000/0.000)} & 0.98 (0.004) & 3.02 & 0.10\\
LSA-MM           & 1.493 (0.039) & \makecell{1.000/0.002\\(0.000/0.000)} & 0.98 (0.004) & 3.02 & 0.11\\
RLARS-MM         & 1.486 (0.040) & \makecell{1.000/0.013\\(0.000/0.001)} & 0.88 (0.010) & 3.12 & 1.59\\
AdaPENSE-WRMSPE  & 2.558 (0.062) & \makecell{1.000/0.004\\(0.000/0.001)} & 0.98 (0.004) & 3.03 & 2.19\\
SCAD-raw         & 2.284 (0.068) & \makecell{1.000/0.011\\(0.000/0.001)} & 0.93 (0.008) & 3.10 & 0.07\\
sparse LTS       & 4.503 (0.104) & \makecell{1.000/0.276\\(0.000/0.005)} & 0.10 (0.010) & 5.49 & 0.17\\
\addlinespace
\multicolumn{6}{@{}l}{\emph{Vertical $10\%$}}\\
D-MM             & \textbf{1.801} (0.047) & \makecell{1.000/0.001\\(0.000/0.000)} & 0.99 (0.003) & 3.01 & 0.09\\
LSA-MM           & 2.005 (0.054) & \makecell{1.000/0.002\\(0.000/0.000)} & 0.99 (0.003) & 3.01 & 0.09\\
RLARS-MM         & 3.825 (0.676) & \makecell{0.997/0.019\\(0.001/0.001)} & 0.83 (0.012) & 3.16 & 1.59\\
AdaPENSE-WRMSPE  & 12.876 (0.521) & \makecell{1.000/0.002\\(0.000/0.001)} & 0.98 (0.004) & 3.02 & 2.14\\
SCAD-raw         & 12.618 (0.380) & \makecell{1.000/0.017\\(0.000/0.001)} & 0.88 (0.010) & 3.16 & 0.08\\
sparse LTS       & 5.513 (0.109) & \makecell{1.000/0.247\\(0.000/0.005)} & 0.13 (0.011) & 5.23 & 0.20\\
\bottomrule
\end{tabular}
\end{table}

Table~\ref{tab:modern} shows that under homoscedastic errors the MM-based
procedures and SCAD-raw are close to each other. D-MM attains the smallest
MSE ($1.667$), followed within $0.25$ by SCAD-raw, LSA-MM and RLARS-MM,
whereas adaptive PENSE ($4.536$) and sparse LTS ($6.306$) are less
accurate. Sparse LTS also over-selects, with a
false-positive rate of $0.242$ and a mean selected size of $5.17$. Under
heteroscedastic errors D-MM again has the smallest MSE ($1.468$), with
RLARS-MM and LSA-MM within $0.03$ of it, while SCAD-raw ($2.284$),
adaptive PENSE ($2.558$) and sparse LTS ($4.503$) fall further back.
Turning to the $10\%$ vertical-contamination rows, SCAD-raw deteriorates to
$12.618$ and adaptive PENSE to $12.876$, RLARS-MM rises to $3.825$ with a
much larger Monte Carlo standard error, and D-MM ($1.801$) and LSA-MM
($2.005$) change little. The true positive rate is $1.000$ for every
method except RLARS-MM under contamination ($0.997$), so the methods
differ in false positives and hence in exact recovery. Exact recovery
ranges from $0.98$ to $0.99$ for D-MM and LSA-MM and from $0.10$ to
$0.14$ for sparse LTS. In terms of computing time, D-MM, LSA-MM and
SCAD-raw need about $0.1$~s per fit and sparse LTS about $0.2$~s, whereas
RLARS-MM needs $1.6$~s and adaptive PENSE between $2.1$ and $2.5$~s.

\begin{table}[htbp]\centering\tiny
\caption{Paired MSE differences for the primary adaptive-PENSE comparator on
the replications of Table~\ref{tab:modern}, reported as the estimate with
its pointwise 95\% Monte Carlo interval, multiplied by $10^{2}$ as in
Table~\ref{tab:modern}; the selector row reports only the
interval of the primary-minus-$\tau$ paired difference.}
\label{tab:modern-diff}
\setlength{\tabcolsep}{2.5pt}
\begin{tabular}{@{}lccc@{}}
\toprule
Difference & Homoscedastic & Heteroscedastic & Vertical $10\%$\\
\midrule
WRMSPE $-$ D-MM & $2.868$ [$2.661,3.076$] & $1.091$ [$0.981,1.201$] & $11.075$ [$10.056,12.094$]\\
WRMSPE $-$ SCAD-raw & $2.743$ [$2.522,2.963$] & $0.275$ [$0.116,0.433$] & $0.258$ [$-0.997,1.513$]\\
WRMSPE $-$ $\tau$ selector & [$-99.326,-75.575$] & [$-139.194,-108.547$] & [$-9.214,-4.324$]\\
\bottomrule
\end{tabular}
\end{table}
{}

The primary adaptive-PENSE comparator recovers the support accurately, with exact recovery between $0.90$ and $0.98$, but it is not
the most accurate method in terms of MSE. \ORtab{tab:modern-diff} reports
its paired MSE differences: those to D-MM are positive in all three
regimes, and those to SCAD-raw are positive under homoscedastic and
heteroscedastic errors, whereas under vertical contamination the interval
covers zero.

The choice of the tuning rule has a substantial effect. With
the $\tau$-size selector, adaptive PENSE attains MSE values of $91.99$,
$126.43$ and $19.65$ in the same three regimes, and the paired intervals
for the difference between the primary rule and the $\tau$ rule in
\ORtab{tab:modern-diff} are wholly negative. We therefore report the
package-default robustness-weighted prediction selector as primary and
preserve the $\tau$ rule only to document selector sensitivity. An
integrated robust criterion can be statistically competitive, but robust
path construction and cross-validation are substantially more expensive
than one MM fit followed by the SCAD path in this implementation.

We complete the contamination geometry with a bad-leverage cell. It uses
the same design with the fixed-size $10\%$ subset moved to high-leverage positions,
every predictor coordinate drawn from $N(5,1)$, and given signal-free
responses, and \ORtab{tab:leverage} reports all seven workflows on 1000
paired replications. The distilled estimators pass through this cell almost
unaffected. LSA-MM records MSE $2.01$ with exact recovery $0.98$ and D-MM
$2.26$ with $0.99$, because the redescending MM initial fit assigns the
off-plane cluster negligible weight and distillation inherits that
rejection. By contrast, SCAD-raw moves to $863$ with a false-positive rate
of $0.655$. The two adaptive-PENSE selection rules trade places relative
to the vertical cell, as the primary WRMSPE rule records $921$ while the
$\tau$-size sensitivity stays at $5.45$.

\begin{table}[htbp]\centering\scriptsize
\caption{Bad-leverage cell on 1000 paired replications: the modern-comparator
design ($n=200$, $p=12$, three nonzero slopes, AR(1) correlation $0.5$,
normal errors) with the fixed-size $10\%$ contaminated row subset moved to
high-leverage positions (every predictor coordinate drawn from $N(5,1)$)
and given signal-free $N(0,1)$ responses. MSE is followed by its Monte
Carlo standard error (MCSE), both multiplied by $10^{2}$; TPR/FPR and
exact recovery are followed by their replication-level Monte Carlo
standard errors. The AdaPENSE-$\tau$ row is the selection-rule
sensitivity of the same fit as AdaPENSE-WRMSPE and is excluded from the
boldface comparison. Three RLARS-MM finite nonconverged refits remain
included.}
\label{tab:leverage}
\setlength{\tabcolsep}{3.3pt}
\begin{tabular}{@{}lccccc@{}}
\toprule
Method & MSE (MCSE) & TPR/FPR & exact & size & seconds\\
\midrule
D-MM                     & 2.258 (0.065) & \makecell{1.000/0.002\\(0.000/0.000)} & 0.99 (0.004) & 3.01 & 0.19\\
LSA-MM                   & \textbf{2.011} (0.055) & \makecell{1.000/0.002\\(0.000/0.001)} & 0.98 (0.004) & 3.02 & 0.10\\
RLARS-MM                 & 206.164 (6.510) & \makecell{0.775/0.069\\(0.006/0.003)} & 0.21 (0.013) & 2.94 & 1.65\\
AdaPENSE-WRMSPE          & 921.120 (17.965) & \makecell{0.755/0.350\\(0.009/0.007)} & 0.03 (0.005) & 5.41 & 3.38\\
AdaPENSE-$\tau$ (sens.)  & 5.447 (0.508) & \makecell{1.000/0.272\\(0.000/0.009)} & 0.28 (0.014) & 5.45 & 3.38\\
SCAD-raw                 & 863.038 (2.025) & \makecell{0.976/0.655\\(0.003/0.005)} & 0.00 (0.000) & 8.83 & 0.24\\
sparse LTS               & 5.927 (0.119) & \makecell{1.000/0.236\\(0.000/0.006)} & 0.20 (0.013) & 5.12 & 0.23\\
\bottomrule
\end{tabular}
\end{table}
{}

\modernchecksintro{} Every displayed
workflow returned a finite fit on every replication, and the 32 RLARS-MM
refits whose final MM step did not converge remain in
Table~\ref{tab:modern}, with run-level records in \provref{}. 
A prespecified grid sensitivity doubled the sparse-LTS fraction grid on 100
paired replications per scenario; all 600 grid--scenario fits
completed cleanly and retained all three signals. Refinement reduced the
endpoint-selection rate from $0.46$, $0.78$, and $0.43$ to $0.07$, $0.25$,
and $0.10$ in the homoscedastic, heteroscedastic, and contaminated
scenarios, while the paired MSE differences remained within Monte Carlo
error ($\Delta\le0.38$ in the scaled units of Table~\ref{tab:modern}, every
interval covering zero). The full grid-sensitivity table, in original
units, is retained in the numerical records described in the data and
code availability statement.
{}
Every study in this section ran under the common computational audit of
\ORtab{tab:sim-audit}, and every check passed.

The flat fidelity curve noted in Section~\ref{sec:algo} is a property of
the construction rather than of the SCAD penalty, and a nonparametric
replication reproduces it exactly. We distilled a quantile smoothing
spline into a least-squares smoothing spline across 126 configurations.
Generalized cross-validation on the initial fitted surface selected the
most flexible point of the smoothing-parameter grid in every one of the
25{,}200 replications, so the distilled fit reproduced its initial fit to
within $5.4\times10^{-5}$ relative integrated squared error. The
criterion of Algorithm~\ref{alg:distill} instead chose between about six
and thirteen effective degrees of freedom and lowered the integrated
squared error in $84.4\%$ of replications, by a median factor of $0.63$
(\splineref{}).

\subsection{Finite-sample effect of the GIC multiplier}\label{sec:sim-tuning}

Conditional on a qualifying computed path, Proposition~\ref{prop:gic}
applies to any fixed $\kappa>0$, because
$a_n=\kappa\log n\to\infty$ and $a_n=o(n)$. We therefore choose $\kappa$
on finite-sample grounds. The choice is a genuine trade-off. It goes in opposite directions in the two
signal regimes, so no single value is uniformly best.

\begin{table}[!htbp]\centering\small
\caption{Effect of the GIC multiplier $\kappa$ in two signal regimes.
Both have $p=12$ and AR(1) correlation $0.5$. Strong design: the base
coefficients, $n=100$, raw-scale $t_3$ errors (variance 3), clean.
Weak design: $\bbeta_0=(0.5,0.3,0,0,0.4,0,\dots)$, $n=200$, normal errors,
at $0\%$ and a uniformly sampled fixed-size 10\% response subset shifted
by 8; its displayed TPR and FPR are for the contaminated setting. Each
setting uses 300 replications and a 100-point SCAD path with $a=3.7$; the
parenthesized quantity in every body cell is its Monte Carlo standard
error, and MSE entries with their standard errors are multiplied by
$10^{2}$; rates are on the original scale.
Exact-recovery MCSE is binomial; MSE, TPR, and FPR MCSE use the replication-level
sample standard deviation. Larger $\kappa$ means a heavier GIC penalty on
model size. Boldface marks the smallest MSE within each MSE column and, in
Panel A, the largest exact-recovery rate.}
\label{tab:kappa}
\begin{tabular}{@{}lccc@{}}
\toprule
\multicolumn{4}{@{}l}{\emph{Panel A: strong design, clean}}\\
$\kappa$ & MSE (MCSE) & exact (MCSE) & FPR (MCSE)\\
\midrule
$0.5$ & 14.84 (0.70) & 0.333 (0.027) & 0.250 (0.014)\\
$1.0$ & 8.45 (0.50) & 0.747 (0.025) & 0.059 (0.007)\\
$2.0$ & 5.76 (0.28) & 0.963 (0.011) & 0.005 (0.002)\\
$4.0$ & \textbf{5.59} (0.28) & \textbf{1.000} (0.000) & 0.000 (0.000)\\
\bottomrule
\end{tabular}

\medskip
\begin{tabular}{@{}lcccc@{}}
\toprule
\multicolumn{5}{@{}l}{\emph{Panel B: weak design}}\\
& \multicolumn{2}{c}{MSE (MCSE)} & & \\
\cmidrule(lr){2-3}
$\kappa$ & $0\%$ & $10\%$ & TPR$_{10\%}$ (MCSE) & FPR$_{10\%}$ (MCSE)\\
\midrule
$1.0$ & \textbf{3.74} (0.18) & \textbf{4.61} (0.22) & 0.948 (0.007) & 0.070 (0.006)\\
$2.0$ & 4.71 (0.25) & 5.63 (0.29) & 0.924 (0.009) & 0.036 (0.004)\\
$4.0$ & 6.33 (0.43) & 10.07 (0.85) & 0.856 (0.014) & 0.016 (0.003)\\
\bottomrule
\end{tabular}
\end{table}
\FloatBarrier

In Panel A of Table~\ref{tab:kappa}, increasing $\kappa$ over the
displayed range removes false positives in the strong design, so that MSE
falls from $14.8$ to $5.6$ while exact recovery rises from $0.33$ to $1.00$.
In the displayed weak design the direction reverses. A heavier penalty
discards some small true coefficients, and MSE rises from $4.6$ to $10.1$.

The mechanism is the same in both halves: a larger $\kappa$ always buys
parsimony. Whether parsimony is worth buying depends on the signal regime.
That is why we set the default at $\kappa=2$, which sits between the two,
giving near-oracle behaviour in the strong design and, in the weak design,
a contaminated-sample true-positive rate of $0.924$ at a false-positive
rate of $0.036$. In settings resembling the displayed weak-signal
scenarios, $\kappa=1$ favours signal retention.

\begin{table}[!htbp]
\centering\tiny
\caption{Consolidated computational audit of the proposed safeguarded SCAD
paths in the simulation studies. Every proposed-path state in the archived
runs was accepted, and no proposed path recorded a solver error, nonfinite
output, candidate-objective increase, or path failure. Apart from the
endpoint selections listed here, every state selected from a proposed path
was interior. The PENSE fit and endpoint counts in the modern-comparator and
leverage rows are reported separately and are not covered by these solver
statements. The
gap column gives the largest selected-state and accepted-path fixed-point
gaps for the proposed paths.}
\label{tab:sim-audit}
\setlength{\tabcolsep}{2pt}
\begin{tabular}{@{}lrll@{}}
\toprule
Study & States/fits & Endpoint selections & Gap (selected; accepted)\\
\midrule
Base study (Section~\ref{sec:sim-main}) & 2{,}000{,}000 & 0 & $4.57\times10^{-5}$; $8.04\times10^{-5}$\\
Oracle branch (Section~\ref{sec:sim-oracle}) & 150{,}000 & 0 & $2.44\times10^{-5}$; ---\\
LSA comparison (Section~\ref{sec:sim-lsa}) & 6{,}000 & 0 & $6.05\times10^{-5}$; $8.15\times10^{-5}$\\
Modern comparators (Section~\ref{sec:sim-modern}) & 9{,}000; 3{,}000 PENSE & 0; 0 PENSE & $5.09\times10^{-5}$; $8.32\times10^{-5}$\\
Leverage cell (\ORtab{tab:leverage}) & 3{,}000; 1{,}000 PENSE & 10 SCAD-raw, dense end; 0 PENSE & $1.09\times10^{-4}$; $1.23\times10^{-4}$\\
Multiplier study (Section~\ref{sec:sim-tuning}) & 6{,}600 & 12 (null model, $\kappa=4$) & $3.68\times10^{-5}$; $4.14\times10^{-5}$\\
\bottomrule
\end{tabular}
\end{table}
{}

\FloatBarrier

\providecommand{\UCIXPerformanceRows}{%
Clean & OLS       & 2.149 (0.022) & 81.0 & 1.000 & 1.000 & 0.099 & 0.000\\
      & MM        & 2.150 (0.022) & 81.0 & 1.000 & 1.000 & 0.099 & 0.000\\
      & D-MM      & 2.147 (0.020) &  9.0 & 0.875 & 0.027 & 0.700 & 0.000\\
      & SCAD-raw  & 2.147 (0.019) &  9.4 & 0.875 & 0.033 & 0.675 & 0.000\\
      & Lasso-CV  & 2.147 (0.021) & 29.2 & 1.000 & 0.290 & 0.276 & 0.000\\
\addlinespace
$5\%$, $+8$ MAD
      & OLS       & 2.632 (0.025) & 81.0 & 1.000 & 1.000 & 0.099 & 0.000\\
      & MM        & 2.151 (0.022) & 81.0 & 1.000 & 1.000 & 0.099 & 0.000\\
      & D-MM      & 2.147 (0.020) &  9.2 & 0.875 & 0.030 & 0.687 & 0.000\\
      & SCAD-raw  & 2.604 (0.017) & 12.6 & 0.800 & 0.085 & 0.453 & 0.000\\
      & Lasso-CV  & 2.608 (0.010) & 26.2 & 1.000 & 0.249 & 0.309 & 0.000\\
\addlinespace
$10\%$, $+8$ MAD
      & OLS       & 3.679 (0.024) & 81.0 & 1.000 & 1.000 & 0.099 & 0.000\\
      & MM        & 2.151 (0.023) & 81.0 & 1.000 & 1.000 & 0.099 & 0.000\\
      & D-MM      & 2.148 (0.020) &  9.0 & 0.875 & 0.027 & 0.700 & 0.000\\
      & SCAD-raw  & 3.638 (0.026) & 11.0 & 0.775 & 0.066 & 0.488 & 0.000\\
      & Lasso-CV  & 3.637 (0.021) & 30.4 & 1.000 & 0.307 & 0.272 & 0.000
}

\providecommand{\UCIXStabilityRows}{%
Clean & D-MM     & 0.9200 & 0.8000 & 1.0000 & 0.7333 & 0.9997\\
      & SCAD-raw & 0.9000 & 0.8000 & 1.0000 & 0.6667 & 0.9996\\
      & Lasso-CV & 0.6298 & 0.5500 & 1.0000 & 0.5239 & 0.9978\\
\addlinespace
$5\%$, $+8$ MAD
      & D-MM     & 0.9600 & 0.9000 & 1.0000 & 0.8667 & 0.9996\\
      & SCAD-raw & 0.5332 & 0.3889 & 0.9143 & 0.2721 & 0.9343\\
      & Lasso-CV & 0.3887 & 0.2857 & 1.0000 & 0.2238 & 0.9640\\
\addlinespace
$10\%$, $+8$ MAD
      & D-MM     & 0.9200 & 0.8000 & 1.0000 & 0.7333 & 0.9995\\
      & SCAD-raw & 0.5468 & 0.4118 & 0.9429 & 0.2212 & 0.9413\\
      & Lasso-CV & 0.3780 & 0.2683 & 1.0000 & 0.2376 & 0.8345
}

\providecommand{\UCIXBlockRows}{%
Clean & D-MM     & 8.0 & 1.000 & 0.000 & 1.000 & 1.0000 & 0.9981\\
      & SCAD-raw & 8.0 & 1.000 & 0.000 & 1.000 & 1.0000 & 0.9977\\
      & Lasso-CV & 8.0 & 1.000 & 0.000 & 1.000 & 1.0000 & 0.9926\\
\addlinespace
$5\%$, $+8$ MAD
      & D-MM     & 8.0 & 1.000 & 0.000 & 1.000 & 1.0000 & 0.9979\\
      & SCAD-raw & 8.0 & 1.000 & 0.000 & 1.000 & 1.0000 & 0.9612\\
      & Lasso-CV & 8.4 & 1.000 & 0.400 & 0.600 & 0.9333 & 0.9404\\
\addlinespace
$10\%$, $+8$ MAD
      & D-MM     & 8.0 & 1.000 & 0.000 & 1.000 & 1.0000 & 0.9972\\
      & SCAD-raw & 7.4 & 0.925 & 0.000 & 0.400 & 0.9250 & 0.8002\\
      & Lasso-CV & 8.2 & 1.000 & 0.200 & 0.800 & 0.9556 & 0.7910
}

\providecommand{\UCIXFiveFoldInterpretation}{%
In Table~\ref{tab:uci-x-performance}, D-MM matches the dense MM initial
fit in test RMSE while selecting about nine of the 81 slopes, under clean
training and under the $5\%$ and $10\%$ training shifts alike. SCAD-raw and
Lasso-CV, by contrast, lose substantial accuracy under the contaminated
training responses.\par
 Table~\ref{tab:uci-x-stability} shows that D-MM's supports and
training-standardized coefficients were nearly unchanged across folds and
contamination levels. The D-MM-to-MM fidelity RMSE averaged 0.178, 0.182,
and 0.185 in the three scenarios, and the corresponding prediction
correlations were 0.9992, 0.9991, and 0.9991.
Nevertheless, exact variable recovery was zero in every scenario. D-MM
selected seven planted coordinates in all 15 fits, never selected
\texttt{wtd\_mean\_Valence}, and selected the correlated
\texttt{wtd\_gmean\_Valence} in its place in all 15 fits
(Figure~\ref{fig:ucix-frequency}; the full table is retained in the numerical
records described in the data and code availability statement). The two
observed Valence descriptors have Pearson correlation 0.9949, and the only
other coordinates ever selected outside the planted support were
\texttt{range\_ThermalConductivity}, in 13 fits but with conditional mean
standardized coefficient only $2.38\times10^{-4}$, and
\texttt{wtd\_std\_Valence}, in three fits. Every signal block was thus
represented. The miss is a systematic within-block substitution between
two descriptors with correlation 0.9949, and the Valence signal itself is
neither lost nor sign-reversed.\par
At the predefined block granularity, recovery is exact
(Table~\ref{tab:uci-x-stability}, panel b). D-MM selected the eight active
physical-property blocks and excluded the null composition-count block in
every fold and scenario, with block support unchanged within each fold
across contamination levels. A support can therefore be perfectly stable
across folds and contamination levels while remaining systematically wrong
at the coordinate level. This coarsening is the explanation audit we
prespecified.\par
All 30 selected SCAD states were interior and accepted, with largest
selected-state fixed-point gap $2.20\times10^{-5}$. For each of the 15
method--scenario combinations, out-of-fold predictions cover all 21,263
rows, and all 30 stored validation checks passed.%
}


\providecommand{\bscr}{\tilde{\bm{\beta}}^{\mathrm{scr}}}
\providecommand{\Sscr}{\widehat{S}}

\section{Extension to large \texorpdfstring{$p/n$}{p/n}: screening the
initial estimator}
\label{sec:screen}

The preceding sections take the initial estimator as given on the full
coordinate set, and Assumption~\ref{ass:2} asks it to be
$\sqrt n$-consistent there. For a robust initial fit such as MM this is a
computational requirement as much as an asymptotic one: the estimator is
undefined once $p+1\ge n$, and the dimension study in
Section~\ref{sec:screen-sim} shows that it fails well before that
boundary. This section asks what changes when the ratio $p/n$ is no
longer small. We keep the distillation stage exactly as
defined in Section~\ref{sec:method} and move only the point at which
the robust fit is computed. A sparse screening step precedes it, so that
the robust estimator solves a problem of bounded size regardless of $p$.
The dimension and density studies below include settings in which $p$
grows with, or exceeds, $n$, and they assess the screening construction
empirically; the transfer analysis in Section~\ref{sec:screen-theory}, by
contrast, keeps $p$ fixed.

\subsection{Definition and assumptions}\label{sec:screen-def}

Fix an integer cap $K$ with $K+1<n$ and let
\[
  \mathcal S_K=\bigl\{S\subseteq\{1,\dots,p\}:\ |S|\le K,\
   [\bm 1,\X_S]\ \text{has full column rank}\bigr\}.
\]
The empty set always belongs to $\mathcal S_K$, so the collection is
nonempty, and because $p$ is fixed every collection appearing below is a
subcollection of the fixed finite family $2^{\{1,\dots,p\}}$. The cap may
grow with $n$ subject to $K+1<n$, and our implementation's default
$K=\lfloor n/4\rfloor$ is covered by this reading.

A \emph{screening rule} is a measurable map
$(\X,\y)\mapsto\Sscr(\X,\y)\in\mathcal S_K$; we write $\Sscr$ for its
value at the observed sample. Membership in $\mathcal S_K$ holds for
every input by construction. We complete a rule whose raw output is
larger than $K$ or produces a rank-deficient $[\bm1,\X_S]$ by a
deterministic repair that truncates to the $K$ largest screened
coefficients and then removes linearly dependent columns in index order.
When the screen retains nothing, the screened fit degrades to the
intercept-only robust location fit. We understand both the screening
rule and the per-submodel robust fit as fixed measurable selections
from their solution sets, tie-broken deterministically, so that the
probability statements below are well posed.

Given $\Sscr$, let $(\tilde\alpha_{\Sscr},\btil_{\Sscr})$ be the robust
fit of $\y$ on $[\bm1,\X_{\Sscr}]$, and let $\bscr\in\mathbb R^p$ be its
embedding, equal to $\btil_{\Sscr}$ on $\Sscr$ and zero elsewhere. The
\emph{screened robust initial fit} is $(\tilde\alpha_{\Sscr},\bscr)$,
and the distilled estimator is obtained by running
Algorithm~\ref{alg:distill} with this fit in place of the rule in its
step 1, so that
$\tilde\y=\tilde\alpha_{\Sscr}\bm1+\X\bscr$ and $\hat\sigma$ are computed
from the screened fit throughout. The screen reads the same contaminated
response as the robust fit, so the analysis below never conditions on the
observed selection. The measurability, cap, and rank bookkeeping this
requires is collected in \ORsec{supp:screen-thy}.

In practice, the default screen is a SCAD-penalized median regression
(\texttt{rqPen}), tuned by the quantile-regression GIC with the same
multiplier $\kappa$. When it selects more than $K=\lfloor n/4\rfloor$
coordinates, the $K$ largest coefficients in absolute value are kept, and a
deterministic rank repair then drops linearly dependent columns. A
SCAD-penalized least-squares screen with the same tuning and cap serves as
the non-robust comparison in Section~\ref{sec:screen-sim}.

\subsection{Transfer of the fixed-\texorpdfstring{$p$}{p} analysis}
\label{sec:screen-theory}
\screenthy{}

Three properties, each of which we establish in \ORsec{supp:screen-thy},
carry the fixed-$p$ analysis over to the screened fit. First, the
screened initial fit inherits the rate. Under sure screening, the cap
$K\ge|\cA|$, and a $\sqrt n$ rate on correctly specified submodels (the
conditions collected there as (A2$'$)), the embedded fit
$(\tilde\alpha_{\Sscr},\bscr)$ satisfies the rate requirement of
Assumption~\ref{ass:2}, so nothing is lost by screening first. Second,
its conditional response-replacement floor is the minimum over admissible
submodels, $\min_{S\in\mathcal S_K}\varepsilon^*_{n,y}(\tilde\theta_S;\y\mid
\X_S)$. Because the screen reads the same contaminated response as the
robust fit, the bound must dominate every support the rule can return, and
this guaranteed floor can be smaller than the full-model bound. The floor applies to a
specified submodel robust-fit rule; the failure-aware composite rule that
we implement requires the separate assessment of \ORsec{supp:impl-cascade}.
Finally, the residual scale of \eqref{eq:sigma} computed from the screened
fit is bounded above and below in probability whenever the design row
norms grow more slowly than $\sqrt n$ and the residuals at the target
follow a continuous law with a unique, nondegenerate median absolute
deviation. This supplies the scale hypothesis of
Proposition~\ref{prop:gic}. Under response contamination that law is the
contaminated residual law, and the fixed-count scheme of the simulations is
covered conditionally on the contaminated index set.

\begin{assumption}\label{ass:2p-main}
(A2$'$) The target conditions of (A2) remain in force, and its rate clause
is replaced by three conditions on the screen and the robust fit: (i)
\emph{sure screening}, $P(\cA\subseteq\Sscr)\to1$; (ii) the cap
satisfies $K\ge|\cA|$; and (iii) for every admissible support $S$ with
$\cA\subseteq S$ and $|S|\le K$, the robust fit on $[\bm1,\X_S]$
satisfies $\sqrt n(\tilde\theta_S-\theta_{0S})=O_p(1)$. The
residual-scale conditions are that the design row norms satisfy
$\max_i\|\bm x_i\|=o(\sqrt n)$ and that the residuals at the target
are independent draws from a continuous law whose median and median
absolute deviation are unique and finite.
\end{assumption}

\begin{corollary}[Transfer of the downstream analysis]\label{cor:scr-modular}
Assume (A1), (A3), the screened-fit conditions (A2$'$) and the
residual-scale conditions just described, and replace the
initial fit
$(\tilde\alpha,\btil)$ by the screened fit
$(\tilde\alpha_{\Sscr},\bscr)$ throughout, so that $\tilde\y$,
$\tilde\theta$, the target $\theta_0$ of the screened-fit conditions (A2$'$) and
$\hat\sigma$ are computed from the screened fit. Then:
\begin{enumerate}[label=(\alph*),leftmargin=2.2em,itemsep=1pt]
\item Proposition~\ref{prop:norm}, Lemma~\ref{lem:warm} and
      Corollary~\ref{cor:bp} hold as stated, with the screened rule's
      conditional response-replacement bound supplied by the screened-fit
      lemmas of \ORsec{supp:screen-thy}; as in Section~\ref{sec:finite}, these
      statements require $[\bm1,\X]$ to have full column rank, hence
      $p+1\le n$.
\item Theorem~\ref{thm:support} and Proposition~\ref{prop:gic} hold as
      stated, with the rate and the scale bound supplied by the same
      lemmas; Proposition~\ref{prop:gic}'s qualifying-path hypothesis is
      still assumed, not supplied.
\item Theorems~\ref{thm:oracle} and \ref{thm:equiv} transfer as
      conditional statements: their distributional hypotheses ---
      asymptotic normality of the initial slopes, and the
      covariance-weight limit $n\hat\Sigma\to_pc_\rho M^{-1}$ --- are
      additional properties that (A2$'$) does not
      supply and that must be verified for the screened fit. When the
      screen is selection consistent, $P(\Sscr=\cA)\to1$, the
      oracle-branch projection degenerates to the submodel refit and the
      proportional-covariance efficiency clause is unavailable; the
      second stage then contributes size control and the transferred
      guarantees of (a)--(b) rather than additional efficiency.
\end{enumerate}
The influence-function analysis of Section~\ref{sec:if} lies outside
this transfer: screening is a discrete selection, and a population
influence function for the screened rule would require a functional
formulation of the screen, which we leave to future work.
\end{corollary}

\subsection{Dimension study}\label{sec:screen-sim}

The study crosses five design axes. The first four are dimension
$p\in\{12,24,48,120,240\}$, sample size $n\in\{100,200,400\}$,
contamination fraction $\{0,0.05,0.10,0.20\}$ with the fixed $+8$
vertical shift of Section~\ref{sec:sim-design}, and normal or $t_3$
errors. The fifth is a signal axis that repeats the base
twelve-coordinate block one, two, or four times before zero padding,
giving $3$, $6$, or $12$ nonzero slopes. Covariates follow the AR(1) design with correlation $0.5$
throughout. This yields 288 scenarios, each run with 100 common
replications; replication seeds are derived deterministically from the
scenario index, so every method within a cell sees the same data and the
results are independent of the parallel execution order. We compare the
distilled estimator built on the full-coordinate MM initial fit with its
screened counterparts, which use a quantile and a least-squares screen,
and with the screened robust initial fit alone. The remaining comparators
are two integrated robust-sparse benchmarks, median-SCAD via
\texttt{rqPen} and Huber-penalized regression via \texttt{hqreg} and both
tuned by the same GIC family, together with SCAD on the raw response and
the LS oracle. Each estimator that involves a GIC is
evaluated under both the default $\kappa\log n$ rate and the
dimension-aware $\kappa\log\log n\,\log p$ rate of \citet{fantang13},
computed on one shared path, so that the initial-fit factor and the
penalty-rate factor of the selection rule can be separated.

\begin{figure}[!htbp]
\centering
\includegraphics[width=\linewidth]{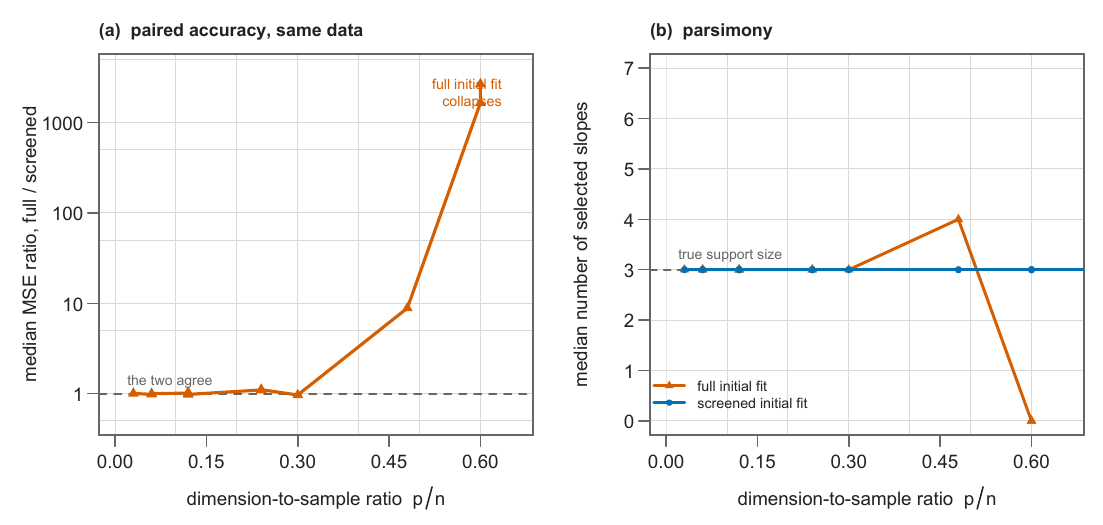}
\caption{Effect of the dimension-to-sample ratio at $10\%$ vertical
contamination with normal errors and three nonzero slopes. Panel (a)
shows the median ratio of the two initial fits' squared errors on the
same replications; panel (b) shows the median number of selected
slopes for each. Both panels use ratios and counts, so that cells with
different $n$ can be read on one axis.}
\label{fig:dim-collapse}
\end{figure}

\dimtable{collapse}

The ratio $p/n$ governs the full-coordinate initial fit.
Figure~\ref{fig:dim-collapse} summarizes the pattern, and
Table~\dimref{tab:dim-collapse} tracks the distilled estimator built on
the full MM fit across the grid, which we order by $p/n$ and summarize
by cell medians. Means are uninformative in the two largest cells, where a
single numerically exploding replication dominates them.
Up to $p/n\approx0.12$ the estimator behaves as in the base study of
Section~\ref{sec:sim}. Performance then degrades steadily through
$p/n=0.24$ and $0.48$, and at $p/n=0.6$ the median selected size is
zero. In most replications the estimator returns the intercept-only
model. Both $p/n=0.6$ cells satisfy $n>p+1$, so the binding constraint is
the ratio $p/n$ rather than the line $p+1\ge n$, and it binds well before
that line is reached.

\begin{figure}[!htbp]
\centering
\includegraphics[width=\linewidth]{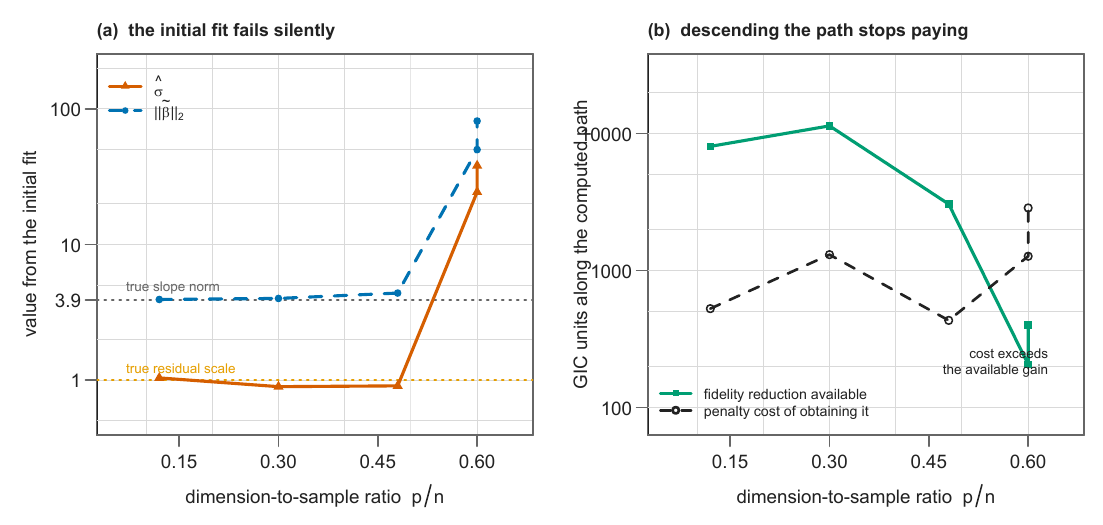}
\caption{Mechanism of the collapse, from the 30-replication path
decomposition. Panel (a): the initial fit's slope norm and residual
scale, against their population values 3.9 and 1. Panel (b): the
fidelity reduction from the null to the densest retained state and the
incremental penalty cost between those states; the two points at $p/n=0.6$ are the
configurations $(p,n)=(240,400)$ and $(120,200)$.}
\label{fig:dim-mechanism}
\end{figure}

\dimtable{mechanism}

Turning to the mechanism, Figure~\ref{fig:dim-mechanism} and
Table~\dimref{tab:dim-mechanism}
give a separate 30-replication path decomposition that diagnoses
finite-sample scale calibration. In every replication of every cell the MM
routine reports successful convergence, so the failure cascade is never
entered. At $p/n=0.6$, however, both the slope norm and the residual scale
$\hat\sigma$ inflate far above their population values. Dividing by an
inflated $\hat\sigma^2$ compresses fidelity differences relative to the
complexity penalty. The reported drop/cost ratio compares only the null
and the densest retained state, and falls below one at $p/n=0.6$.
The densest state therefore does not repay its incremental penalty;
this does not exclude an intermediate state from improving on the null.
Separately, the actual GIC minimizer is null in $87$--$90\%$ of replications:
seven of the 60 replications at $p/n=0.6$ select a non-null state despite
having drop/cost below one. The diagnostic supports scale inflation as a
contributor to null selection, but does not isolate it from deterioration
of the initial coefficients or establish a violation of the fixed-$p$
asymptotic scale condition in Proposition~\ref{prop:gic}.

\begin{table}[!htbp]\centering
\scriptsize
\caption{Dimension study at $n=200$ with $10\%$ vertical contamination, normal errors, three nonzero slopes, AR(1) correlation $0.5$, and the default GIC rate, over 100 common replications. Each cell shows mean MSE (Monte Carlo standard error) over the exact-recovery rate. A dash marks configurations in which the full-coordinate robust initial fit is undefined ($p+1\ge n$). Mean squared errors and their Monte Carlo standard errors are multiplied by $10^{2}$, as in Table~\ref{tab:modern}.}
\label{tab:dim-main}
\setlength{\tabcolsep}{2pt}
\begin{tabular}{@{}lcccc@{}}
\toprule
Method & $p{=}12$ & $p{=}48$ & $p{=}120$ & $p{=}240$\\
\midrule
D-MM, full initial & 1.75 (0.17) & 1.93 (0.19) & 1917.07 (52.74) & -- \\
\quad exact & 0.99 & 0.99 & 0.00 & -- \\
\addlinespace D-MM, screened (quant.) & 1.82 (0.19) & 1.80 (0.18) & 1.70 (0.14) & 1.84 (0.16) \\
\quad exact & 0.99 & 1.00 & 1.00 & 1.00 \\
\addlinespace D-MM, screened (LS) & 1.82 (0.19) & 1.80 (0.18) & 3.42 (1.73) & 5.27 (2.44) \\
\quad exact & 0.99 & 1.00 & 0.99 & 0.98 \\
\addlinespace screened initial fit alone & 1.82 (0.19) & 1.80 (0.18) & 1.70 (0.14) & 1.84 (0.16) \\
\quad exact & 0.99 & 1.00 & 1.00 & 1.00 \\
\addlinespace median-SCAD & 2.80 (0.29) & 2.98 (0.27) & 3.03 (0.30) & 2.88 (0.29) \\
\quad exact & 0.99 & 1.00 & 1.00 & 1.00 \\
\addlinespace Huber-penalized & 12.20 (0.82) & 15.77 (1.09) & 20.87 (1.20) & 23.36 (1.21) \\
\quad exact & 0.83 & 0.87 & 0.88 & 0.86 \\
\addlinespace SCAD-raw & 12.58 (1.26) & 22.01 (2.65) & 34.36 (3.46) & 40.09 (4.03) \\
\quad exact & 0.96 & 0.93 & 0.76 & 0.79 \\
\addlinespace LS oracle & 10.19 (0.85) & 10.28 (0.77) & 10.90 (0.80) & 10.11 (0.77) \\
\quad exact & 1.00 & 1.00 & 1.00 & 1.00 \\
\bottomrule
\end{tabular}
\par\vspace{2pt}
\parbox{0.95\linewidth}{\scriptsize At $p=120$ the full-initial row is dominated by the collapse described in Table~\ref{tab:dim-mechanism}: its mean selected size is 0.37 slopes and its exact-recovery rate is zero, so the MSE entry measures a near-null model.}
\end{table}

By contrast, screening removes the dimension effect. In
Table~\ref{tab:dim-main} the screened distilled estimator is flat in
$p$, with exact recovery at or near one throughout, while the
full-initial version deteriorates by three orders of magnitude at
$p=120$ and is undefined at $p=240$. The integrated benchmarks remain
well defined but drift upward with $p$, and median-SCAD is the strongest
among them. The screened estimator and its initial fit alone post
identical rows. The reason is that in these three-signal configurations
the quantile screen typically returns the true support, so the second
stage degenerates as described in Corollary~\ref{cor:scr-modular}(c). The
contribution of the second stage is therefore visible in selection
rather than in error, and we report it in Table~\dimref{tab:dim-paired}
as a paired comparison over the replications in which both estimators
are defined. The MSE gain is concentrated in the cells where the screen
over-selects; elsewhere the two estimators coincide and the
replication-level win rate stays close to one half.

\dimtable{paired}

\dimtable{rate}

\dimtable{screen}

We also ran three robustness checks, none of which changes the picture.
Neither GIC penalty rate rescues the full-initial version or alters the
screened estimator, and the collapse of Table~\dimref{tab:dim-collapse}
occurs under both (Table~\dimref{tab:dim-rate}). A same-seed nested audit
of the four hardest cells, extending the path's lower endpoint from
$0.05$ to $0.005$ and $0.001$, changes no displayed conclusion
(\ORtab{tab:dim-endpoint}). The third check concerns the screen itself,
which must be robust. Under contamination a least-squares screen feeds
the refit a support already shaped by the outliers, so the quantile
screen's advantage widens with the contamination fraction
(Table~\dimref{tab:dim-screen}). 

Table~\dimref{tab:dim-rate} reports the factorial diagnostic that
attributes the collapse of the full-initial estimator in
Table~\dimref{tab:dim-collapse} to the initial fit rather than to the
GIC rate. For the screened distilled estimator the two GIC rates
select the same fitted model in at least $98\%$ of applicable
replications at every $p$, so the rate choice is immaterial for it. No choice of rate rescues the full-initial version: the collapse of
Table~\dimref{tab:dim-collapse} occurs under both.

The integrated benchmarks are more rate-sensitive ---
agreement for Huber-penalized regression falls to $0.72$ at $p=240$ ---
and the full-initial version falls to $0.87$.

Table~\dimref{tab:dim-screen} pairs the quantile screen against a
least-squares SCAD screen across the full grid. Without contamination the
two are comparable and the least-squares screen recovers the support
marginally more often. The gap opens with the contamination fraction: at
$10\%$ the least-squares screen's MSE is more than twice the quantile
screen's, and at $20\%$ its exact recovery falls to $0.77$ against $0.91$.

\medskip\noindent\emph{Lower-endpoint sensitivity.}
To assess lower-grid truncation under deliberately difficult conditions, we
targeted the four cells crossing $p\in\{120,240\}$ and normal/$t_3$ errors
at $n=100$, $20\%$ contamination, and four signal blocks. We reran them with
the same 100 seeds, retaining every point of the original path ending at
$\lambda/\lambda_{\max}=0.05$ and appending a nested continuation to the
exact ratios $0.005$ and $0.001$. The original selected states and scores
were reproduced in all $1{,}600$ method--selector comparisons. All $800$
screen-specific path computations completed without a solver or path
failure, yielding $4{,}800$ scored method--selector--endpoint rows. The two
deeper ranges selected the same path index in all $1{,}600$ comparisons,
and none of those selections remained at the new lower endpoint.

Table~\ref{tab:dim-endpoint} gives the primary quantile-screened results.
Although the selected path index changes in $28$--$38\%$ of these
deliberately adverse replications, the absolute relative change in mean squared error is
at most $0.72\%$. Pooled over the four cells, the paired MSE change is
$5.91$ (95\% Monte Carlo interval $[2.08,9.75]$), which is $0.52\%$
of the baseline mean. Exact recovery changes by at most $0.01$, and mean
model size changes by at most $0.08$ slopes. The TPR change is at most
$0.00084$ and the FPR change at most $0.00065$. With the least-squares
screen, the selected path index changes in $44$--$65\%$ of replications and
the paired mean MSE increase ranges from $9.5$ to $32.5$. The extension
does not alter its lower exact recovery or larger mean selected size relative
to the quantile screen in these four cells, although the cellwise MSE
ordering is not uniform. These checks support the qualitative comparisons
while making clear that the selected states in these stress cells are
conditional on the prespecified production range; we do not retune that
range after inspecting coefficient truth.

\begin{table}[!htbp]
\centering
\scriptsize
\caption{Nested lower-endpoint audit for the primary quantile-screened
distilled estimator with the default $\kappa\log n$ selector. These are the
four deliberately difficult dimension-study cells crossing $p=120,240$
and normal/$t_3$ errors:
$n=100$, $20\%$ contamination, and four signal blocks. The original
100-point path ending at $\lambda/\lambda_{\max}=0.05$ is preserved exactly
and then extended to $0.005$ and $0.001$. ``State changed'' is the fraction
of replications in which the $0.005$ extension selects a different path index; all
deltas are extended minus original, and parentheses give the Monte Carlo
standard error of the paired MSE delta; both are multiplied by
$10^{2}$, as in Table~\ref{tab:modern}. The $0.001$ extension selected
the same path index as the $0.005$ extension in every replication.}
\label{tab:dim-endpoint}
\setlength{\tabcolsep}{2.2pt}
\begin{tabular}{@{}rrcccccc@{}}
\toprule
$p$ & Error & End at $0.05$ & State changed & $\Delta$MSE (MCSE) & $\Delta$Exact & $\Delta$FPR & $\Delta|S|$\\
\midrule
120 & Normal & 0.27 & 0.30 & $-0.10$ (2.63) & $-0.01$ & 0.00046 & 0.05\\
120 & $t_3$  & 0.35 & 0.38 & $\phantom{-}7.73$ (3.62) & $-0.01$ & 0.00065 & 0.07\\
240 & Normal & 0.33 & 0.35 & $\phantom{-}4.53$ (3.78) & $\phantom{-}0.00$ & 0.00004 & 0.02\\
240 & $t_3$  & 0.29 & 0.28 & $\phantom{-}11.50$ (5.16) & $\phantom{-}0.00$ & 0.00035 & 0.08\\
\bottomrule
\end{tabular}
\end{table}

\begin{figure}[!htbp]
\centering
\includegraphics[width=\linewidth]{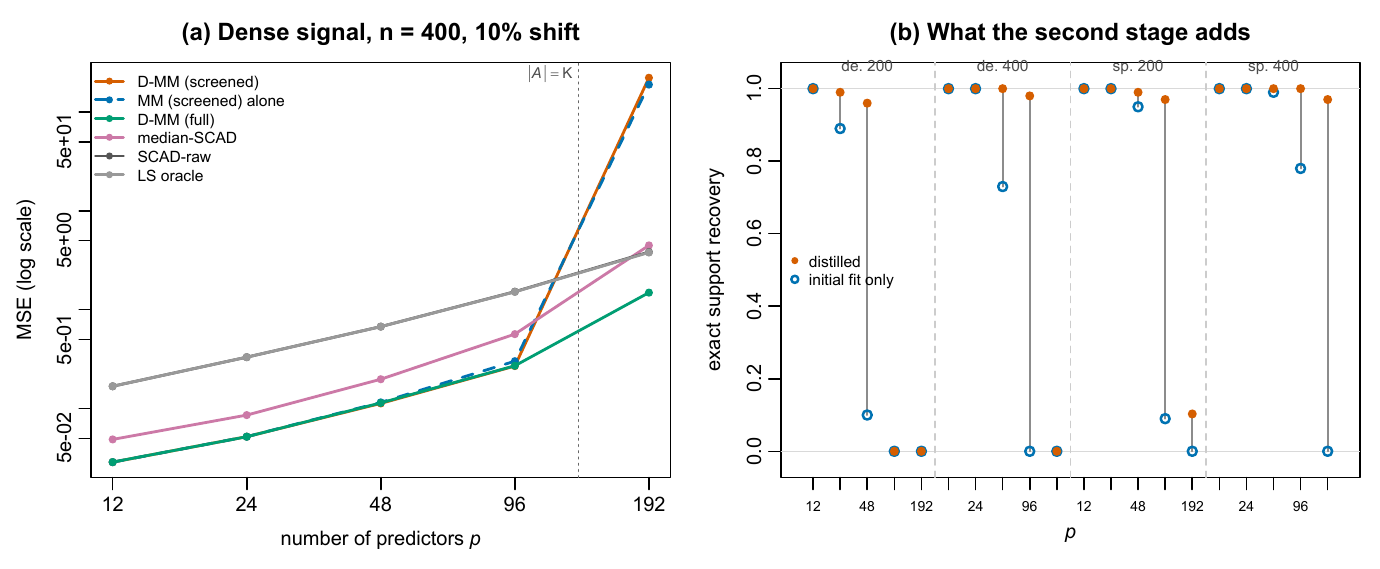}
\caption{Signal-density study under a fixed-size $10\%$ vertical shift.
Panel (a) follows the dense pattern at $n=400$; the dotted line marks where
the number of nonzero slopes reaches the screen cap $K=\lfloor n/4\rfloor$.
Panel (b) pairs exact support recovery of the distilled estimator against
that of its screened initial fit alone in every cell, with a segment joining
the two; the panel is grouped by pattern and sample size.}
\label{fig:density}
\end{figure}

\medskip\noindent\emph{Signal-density design.}
The dimension grid keeps the number of nonzero slopes fixed while $p$ grows,
so the signal becomes sparser as the design widens. That regime is the one
in which the screen typically returns the true support, and the second stage
then has nothing left to remove: the distilled estimator and its screened
initial fit post identical rows, which is the degeneracy of
Corollary~\ref{cor:scr-modular}(c). A second study varies the signal density
instead. Blocks of length twelve are repeated without zero padding, so the
number of nonzero slopes grows with the design: a sparse block with three
nonzero entries and a dense block with nine, repeated $r\in\{1,2,4,8,16\}$
times, giving $p=12r$ and $|\cA|=3r$ or $9r$. Sample sizes are $n\in\{200,
400\}$, errors are normal, and contamination is either absent or the
fixed-size $10\%$ shift used throughout; each of the 40 cells uses 100
replications with the same deterministic seeding as above.
Table~\dimref{tab:density-full} gives the per-cell MSE of every estimator
and Table~\dimref{tab:density-screen} the screen size, cap rate and exact
recovery; the findings are read with Figure~\ref{fig:density} and
Table~\ref{tab:density-gain} of the main text.
{}

\begin{table}[!htbp]\centering\small
\caption{Contribution of the distillation stage under a fixed-size $10\%$
vertical shift, from 100 paired replications per cell. Both estimators use
the same screen and the same screened initial fit on each replication, so
the paired difference isolates the second stage. Negative differences favour
the distilled estimator; the interval is a pointwise 95\% Monte Carlo
interval for the paired mean and the win rate is the proportion of
replications in which the distilled fit has the smaller squared error. Exact
recovery is reported for both. Paired mean squared differences and
their interval endpoints are multiplied by $10^{2}$, as in
Table~\ref{tab:modern}. Cells marked $\dagger$ violate the cap
condition $K\ge|\cA|$ with $K=\lfloor n/4\rfloor$; the cell marked
$\ddagger$ reports a dedicated rerun with 1000 paired replications.}
\label{tab:density-gain}
\setlength{\tabcolsep}{3.5pt}
\begin{tabular}{@{}lccccccc@{}}
\toprule
 & \multicolumn{2}{c}{Paired MSE difference} & & \multicolumn{2}{c}{Exact recovery} & \\
\cmidrule(lr){2-3}\cmidrule(lr){5-6}
Configuration & mean & 95\% interval & win & distilled & initial only & size\\
\midrule
\multicolumn{7}{@{}l}{\emph{Sparse block (3 of 12)}}\\
$p{=}12$, $n{=}200$, $|\cA|{=}3$ & 0.00 & [-0.00, 0.00] & 0.45 & 1.00 & 1.00 & +0.0\\
$p{=}24$, $n{=}200$, $|\cA|{=}6$ & 0.00 & [-0.00, 0.00] & 0.44 & 1.00 & 1.00 & +0.0\\
$p{=}48$, $n{=}200$, $|\cA|{=}12$ & -0.19 & [-0.37, -0.01] & 0.35 & 0.99 & 0.95 & -0.1\\
$p{=}96$, $n{=}200$, $|\cA|{=}24$ & -5.24 & [-6.44, -4.04] & 0.97 & 0.97 & 0.09 & -3.2\\
$p{=}192$, $n{=}200$, $|\cA|{=}48$$^{\ddagger}$ & 41.89 & [30.43, 53.35] & 0.55 & 0.10 & 0.00 & -1.4\\
$p{=}12$, $n{=}400$, $|\cA|{=}3$ & 0.00 & [-0.00, 0.00] & 0.49 & 1.00 & 1.00 & +0.0\\
$p{=}24$, $n{=}400$, $|\cA|{=}6$ & 0.00 & [-0.00, 0.00] & 0.46 & 1.00 & 1.00 & +0.0\\
$p{=}48$, $n{=}400$, $|\cA|{=}12$ & -0.04 & [-0.12, 0.04] & 0.34 & 1.00 & 0.99 & -0.0\\
$p{=}96$, $n{=}400$, $|\cA|{=}24$ & -0.27 & [-0.41, -0.13] & 0.56 & 1.00 & 0.78 & -0.2\\
$p{=}192$, $n{=}400$, $|\cA|{=}48$ & -5.62 & [-6.40, -4.85] & 1.00 & 0.97 & 0.00 & -7.7\\
\addlinespace
\multicolumn{7}{@{}l}{\emph{Dense block (9 of 12)}}\\
$p{=}12$, $n{=}200$, $|\cA|{=}9$ & 0.00 & [-0.00, 0.00] & 0.46 & 1.00 & 1.00 & +0.0\\
$p{=}24$, $n{=}200$, $|\cA|{=}18$ & -0.26 & [-0.47, -0.05] & 0.45 & 0.99 & 0.89 & -0.1\\
$p{=}48$, $n{=}200$, $|\cA|{=}36$ & -2.31 & [-2.86, -1.77] & 0.86 & 0.96 & 0.10 & -2.4\\
$p{=}96$, $n{=}200$, $|\cA|{=}72$$^{\dagger}$ & 1238.76 & [971.65, 1505.88] & 0.11 & 0.00 & 0.00 & -3.0\\
$p{=}192$, $n{=}200$, $|\cA|{=}144$$^{\dagger}$ & 1128.79 & [530.12, 1727.47] & 0.40 & 0.00 & 0.00 & -0.9\\
$p{=}12$, $n{=}400$, $|\cA|{=}9$ & 0.00 & [-0.00, 0.00] & 0.46 & 1.00 & 1.00 & +0.0\\
$p{=}24$, $n{=}400$, $|\cA|{=}18$ & 0.00 & [-0.00, 0.00] & 0.43 & 1.00 & 1.00 & +0.0\\
$p{=}48$, $n{=}400$, $|\cA|{=}36$ & -0.27 & [-0.43, -0.11] & 0.58 & 1.00 & 0.73 & -0.3\\
$p{=}96$, $n{=}400$, $|\cA|{=}72$ & -3.22 & [-3.71, -2.73] & 0.97 & 0.98 & 0.00 & -5.4\\
$p{=}192$, $n{=}400$, $|\cA|{=}144$$^{\dagger}$ & 3250.98 & [2660.88, 3841.09] & 0.06 & 0.00 & 0.00 & -6.2\\
\bottomrule
\end{tabular}
\end{table}

In a second study we vary the signal density instead. Twelve-coordinate
blocks repeat without zero padding over $r\in\{1,2,4,8,16\}$, giving
$p=12r$ with $|\cA|=3r$ or $9r$ at $n\in\{200,400\}$, clean or under the
fixed-size $10\%$ shift, with 100 replications per cell
(Table~\dimref{tab:density-full}). In Table~\ref{tab:density-gain} we
report the paired difference between the distilled estimator and its own
screened initial fit, which isolates the second stage because both share
the screen and the screened fit on each replication. Where the screen
returns the true support the difference is exactly zero, as the corollary
predicts. Where the screen over-selects, the second stage removes the
extra coordinates. Specifically, at $p=192$ with three signals per block the distilled
estimator improves the paired mean squared error by $5.62$, with a Monte
Carlo interval entirely below zero, and wins in every one of the 100
replications. It also raises exact recovery from $0.00$ to $0.97$ while
cutting the mean selected size by $7.7$ slopes. The dense pattern shows
the same reversal one step earlier, at $p=96$. Theorem~\ref{thm:support}
accounts for this, since the distilled fit is the projection of the whole
screened estimate onto the retained columns, which differs from the
screened fit truncated to those columns whenever the dropped coordinates
are correlated with the retained ones.

Figure~\ref{fig:density}(a) shows every method degrading smoothly with
$p$ until the number of nonzero slopes approaches $K$, after which the
screened estimators leave the plot. Beyond that point the binding
constraint is the cap $K$ itself and no longer $p$. With $|\cA|>K$ the
sure-screening condition (A2$'$)(i) cannot hold for any rule that
respects the cap, and the distilled fit built on the full coordinate set
is again the better choice. Contamination tightens the boundary further,
because the quantile screen itself selects fewer coordinates under a
shifted response. At $p=192$, $n=200$ it retains $9.4$ of $144$ signals.
In that near-cap contaminated regime the ordering of
Table~\dimref{tab:dim-screen} reverses. Containment is what matters
there: the least-squares screen over-selects and fills the cap, with
mean screened size $49.6$ of $K=50$ in the sparse $p=192$, $n=200$ cell,
and so keeps exact recovery at $0.85$--$0.91$. The under-selecting
quantile screen retains $26.5$ coordinates and falls to $0.10$
(Table~\dimref{tab:density-screen}). We reran this cell in a dedicated
1000-replication run to settle the paired comparison. Past the boundary
the distillation stage costs $41.9$ in paired mean squared error against
its own screened initial fit, with interval $[30.4,53.4]$ excluding
zero, and raises exact recovery only to $0.10$.
Once contamination pushes the screen below the signal,
the gain of Table~\ref{tab:density-gain} does not merely shrink but changes sign, so a robust screen is
an advantage only in the sparse-signal regime. Read together, the two
studies show that screening can make a robust initial fit usable when $p/n$
is large, while the cap limits this route when the signal is dense or
approaches the cap. Full tables are in Table~\dimref{tab:density-full}, and
the screen sizes and recovery rates behind this paragraph in
Table~\dimref{tab:density-screen}.

Finally, Table~\dimref{tab:dim-timing} reports a separate sequential
single-thread timing run of each complete workflow. The full-initial
version's budget is consumed by the S-estimation step, and it returns
the near-null fit described above.

\dimtable{timing}
\begin{table}[!htbp]\centering\tiny
\caption{Signal-density study, mean squared error over 100 replications per
cell. Blocks of length twelve are repeated without zero padding, so the
number of nonzero slopes grows with $p$: three per block in the sparse
pattern and nine in the dense pattern. Covariates follow the AR(1) design
with correlation $0.5$ and errors are normal; contamination shifts a
fixed-size subset of responses by $+8$. Mean squared errors are
multiplied by $10^{2}$, as in Table~\ref{tab:modern}. Entries are
$--$ where the estimator
is undefined.}
\label{tab:density-full}
\setlength{\tabcolsep}{2.5pt}
\begin{tabular}{@{}lccccccc@{}}
\toprule
Configuration & D-MM (scr.) & MM (scr.) alone & D-MM (full) & median-SCAD & Huber-pen. & SCAD-raw & LS oracle\\
\midrule
\multicolumn{8}{@{}l}{\emph{sparse block, 0\% contamination}}\\
$p{=}12$, $n{=}200$ & 1.61 & 1.64 & 1.62 & 2.19 & 6.53 & 1.55 & 1.55\\
$p{=}24$, $n{=}200$ & 3.87 & 4.06 & 3.75 & 5.04 & 14.30 & 3.45 & 3.43\\
$p{=}48$, $n{=}200$ & 7.39 & 7.39 & 7.49 & 9.86 & 29.35 & 6.54 & 6.53\\
$p{=}96$, $n{=}200$ & 16.03 & 20.21 & 31.25 & 23.05 & 67.27 & 14.08 & 13.98\\
$p{=}192$, $n{=}200$ & 691.35 & 697.79 & 9496.60 & 1047.28 & 6963.94 & 32.38 & 32.38\\
$p{=}12$, $n{=}400$ & 0.84 & 0.92 & 0.85 & 1.25 & 3.22 & 0.81 & 0.81\\
$p{=}24$, $n{=}400$ & 1.54 & 1.58 & 1.51 & 2.19 & 6.59 & 1.45 & 1.45\\
$p{=}48$, $n{=}400$ & 3.32 & 3.39 & 3.29 & 4.47 & 13.72 & 3.12 & 3.08\\
$p{=}96$, $n{=}400$ & 6.81 & 6.91 & 7.03 & 9.15 & 28.13 & 6.46 & 6.46\\
$p{=}192$, $n{=}400$ & 14.96 & 20.49 & 26.04 & 20.59 & 68.40 & 13.18 & 13.18\\
\addlinespace
\multicolumn{8}{@{}l}{\emph{dense block, 0\% contamination}}\\
$p{=}12$, $n{=}200$ & 4.83 & 4.86 & 4.86 & 6.10 & 7.39 & 4.67 & 4.67\\
$p{=}24$, $n{=}200$ & 11.70 & 11.87 & 11.85 & 15.18 & 15.80 & 11.23 & 11.16\\
$p{=}48$, $n{=}200$ & 25.12 & 26.55 & 25.83 & 31.82 & 34.97 & 22.86 & 22.66\\
$p{=}96$, $n{=}200$ & 11030.08 & 9399.39 & 84.67 & 130.35 & 114.46 & 56.42 & 55.48\\
$p{=}192$, $n{=}200$ & 120207.06 & 117276.41 & 51872.68 & 99433.00 & 136263.11 & 52118.66 & 267.55\\
$p{=}12$, $n{=}400$ & 2.26 & 2.28 & 2.24 & 3.17 & 3.91 & 2.14 & 2.14\\
$p{=}24$, $n{=}400$ & 5.02 & 5.04 & 5.06 & 6.97 & 7.78 & 4.76 & 4.76\\
$p{=}48$, $n{=}400$ & 10.87 & 10.99 & 10.93 & 14.06 & 15.54 & 10.28 & 10.28\\
$p{=}96$, $n{=}400$ & 24.91 & 27.18 & 25.32 & 32.44 & 39.76 & 22.80 & 22.65\\
$p{=}192$, $n{=}400$ & 22067.29 & 18892.57 & 90.20 & 160.61 & 156.24 & 57.72 & 56.99\\
\addlinespace
\multicolumn{8}{@{}l}{\emph{sparse block, 10\% contamination}}\\
$p{=}12$, $n{=}200$ & 2.05 & 2.05 & 2.13 & 3.28 & 10.43 & 15.39 & 11.24\\
$p{=}24$, $n{=}200$ & 3.48 & 3.48 & 3.72 & 5.67 & 29.24 & 27.78 & 20.28\\
$p{=}48$, $n{=}200$ & 7.65 & 7.84 & 7.85 & 13.39 & 62.16 & 59.04 & 42.85\\
$p{=}96$, $n{=}200$ & 15.57 & 20.82 & 72.00 & 42.46 & 183.15 & 146.82 & 92.97\\
$p{=}192$, $n{=}200$ & 15342.87 & 15316.57 & 18298.04 & 16405.44 & 34029.15 & 348.97 & 219.27\\
$p{=}12$, $n{=}400$ & 1.01 & 1.01 & 1.01 & 1.54 & 5.18 & 5.24 & 5.24\\
$p{=}24$, $n{=}400$ & 1.74 & 1.74 & 1.76 & 3.16 & 14.86 & 10.60 & 10.27\\
$p{=}48$, $n{=}400$ & 3.54 & 3.58 & 3.63 & 6.07 & 31.67 & 22.56 & 21.51\\
$p{=}96$, $n{=}400$ & 7.39 & 7.66 & 7.57 & 13.24 & 79.91 & 51.23 & 47.46\\
$p{=}192$, $n{=}400$ & 16.15 & 21.77 & 35.28 & 44.04 & 247.99 & 100.73 & 93.65\\
\addlinespace
\multicolumn{8}{@{}l}{\emph{dense block, 10\% contamination}}\\
$p{=}12$, $n{=}200$ & 6.03 & 6.03 & 6.08 & 9.54 & 16.01 & 33.35 & 32.97\\
$p{=}24$, $n{=}200$ & 11.78 & 12.04 & 11.91 & 19.93 & 33.36 & 65.06 & 64.99\\
$p{=}48$, $n{=}200$ & 26.91 & 29.23 & 27.07 & 55.20 & 104.62 & 151.62 & 148.67\\
$p{=}96$, $n{=}200$ & 10826.62 & 9587.86 & 216.05 & 391.81 & 67761.76 & 417.96 & 398.37\\
$p{=}192$, $n{=}200$ & 128651.90 & 127523.10 & 79564.78 & 120826.75 & 136263.11 & 134189.17 & 1854.96\\
$p{=}12$, $n{=}400$ & 2.87 & 2.87 & 2.87 & 4.87 & 8.37 & 16.80 & 16.80\\
$p{=}24$, $n{=}400$ & 5.19 & 5.19 & 5.19 & 8.59 & 17.57 & 33.09 & 33.09\\
$p{=}48$, $n{=}400$ & 11.27 & 11.53 & 11.40 & 19.79 & 42.02 & 67.41 & 67.41\\
$p{=}96$, $n{=}400$ & 26.91 & 30.13 & 27.09 & 56.60 & 157.54 & 152.17 & 151.49\\
$p{=}192$, $n{=}400$ & 22174.38 & 18923.40 & 148.85 & 446.99 & 136217.47 & 383.07 & 379.07\\
\bottomrule
\end{tabular}
\end{table}
\begin{table}[!htbp]\centering\scriptsize
\caption{Screen behaviour in the signal-density study (100 replications
per cell). Screen size is the mean number of coordinates the quantile
screen retains before the second stage, cap rate the proportion of
replications in which the size cap $K=\lfloor n/4\rfloor$ bound, and
exact recovery is reported for the screened distilled estimator and for
its screened initial fit alone; size is the distilled estimator's mean
selected size.}
\label{tab:density-screen}
\setlength{\tabcolsep}{3pt}
\begin{tabular}{@{}lcccccc@{}}
\toprule
Configuration & $|\cA|$ & screen size & cap rate & exact (dist.) & exact (init.) & size\\
\midrule
\multicolumn{7}{@{}l}{\emph{sparse block, 0\% contamination}}\\
$p{=}12$, $n{=}200$ & 3 & 3.0 & 0.00 & 1.00 & 0.99 & 3.0\\
$p{=}24$, $n{=}200$ & 6 & 6.1 & 0.00 & 0.98 & 0.93 & 6.0\\
$p{=}48$, $n{=}200$ & 12 & 12.1 & 0.00 & 0.95 & 0.95 & 12.1\\
$p{=}96$, $n{=}200$ & 24 & 26.3 & 0.00 & 0.88 & 0.18 & 24.2\\
$p{=}192$, $n{=}200$ & 48 & 49.1 & 0.97 & 0.82 & 0.00 & 47.0\\
$p{=}12$, $n{=}400$ & 3 & 3.0 & 0.00 & 1.00 & 0.97 & 3.0\\
$p{=}24$, $n{=}400$ & 6 & 6.0 & 0.00 & 0.99 & 0.97 & 6.0\\
$p{=}48$, $n{=}400$ & 12 & 12.1 & 0.00 & 0.98 & 0.96 & 12.0\\
$p{=}96$, $n{=}400$ & 24 & 24.1 & 0.00 & 0.98 & 0.93 & 24.0\\
$p{=}192$, $n{=}400$ & 48 & 54.1 & 0.00 & 0.91 & 0.00 & 48.2\\
\addlinespace
\multicolumn{7}{@{}l}{\emph{dense block, 0\% contamination}}\\
$p{=}12$, $n{=}200$ & 9 & 9.0 & 0.00 & 1.00 & 0.99 & 9.0\\
$p{=}24$, $n{=}200$ & 18 & 18.1 & 0.00 & 0.99 & 0.93 & 18.0\\
$p{=}48$, $n{=}200$ & 36 & 37.5 & 0.00 & 0.97 & 0.25 & 36.0\\
$p{=}96$, $n{=}200$ & 72 & 50.0 & 1.00 & 0.00 & 0.00 & 46.6\\
$p{=}192$, $n{=}200$ & 144 & 16.4 & 0.27 & 0.00 & 0.00 & 14.3\\
$p{=}12$, $n{=}400$ & 9 & 9.0 & 0.00 & 0.99 & 0.98 & 9.0\\
$p{=}24$, $n{=}400$ & 18 & 18.0 & 0.00 & 1.00 & 0.99 & 18.0\\
$p{=}48$, $n{=}400$ & 36 & 36.1 & 0.00 & 0.99 & 0.92 & 36.0\\
$p{=}96$, $n{=}400$ & 72 & 76.0 & 0.00 & 0.95 & 0.02 & 72.0\\
$p{=}192$, $n{=}400$ & 144 & 100.0 & 1.00 & 0.00 & 0.00 & 93.7\\
\addlinespace
\multicolumn{7}{@{}l}{\emph{sparse block, 10\% contamination}}\\
$p{=}12$, $n{=}200$ & 3 & 3.0 & 0.00 & 1.00 & 1.00 & 3.0\\
$p{=}24$, $n{=}200$ & 6 & 6.0 & 0.00 & 1.00 & 1.00 & 6.0\\
$p{=}48$, $n{=}200$ & 12 & 12.1 & 0.00 & 0.99 & 0.95 & 12.0\\
$p{=}96$, $n{=}200$ & 24 & 27.3 & 0.00 & 0.97 & 0.09 & 24.1\\
$p{=}192$, $n{=}200$ & 48 & 26.5 & 0.32 & 0.10 & 0.00 & 25.1\\
$p{=}12$, $n{=}400$ & 3 & 3.0 & 0.00 & 1.00 & 1.00 & 3.0\\
$p{=}24$, $n{=}400$ & 6 & 6.0 & 0.00 & 1.00 & 1.00 & 6.0\\
$p{=}48$, $n{=}400$ & 12 & 12.0 & 0.00 & 1.00 & 0.99 & 12.0\\
$p{=}96$, $n{=}400$ & 24 & 24.2 & 0.00 & 1.00 & 0.78 & 24.0\\
$p{=}192$, $n{=}400$ & 48 & 55.8 & 0.00 & 0.97 & 0.00 & 48.0\\
\addlinespace
\multicolumn{7}{@{}l}{\emph{dense block, 10\% contamination}}\\
$p{=}12$, $n{=}200$ & 9 & 9.0 & 0.00 & 1.00 & 1.00 & 9.0\\
$p{=}24$, $n{=}200$ & 18 & 18.1 & 0.00 & 0.99 & 0.89 & 18.0\\
$p{=}48$, $n{=}200$ & 36 & 38.4 & 0.00 & 0.96 & 0.10 & 36.0\\
$p{=}96$, $n{=}200$ & 72 & 50.0 & 1.00 & 0.00 & 0.00 & 47.0\\
$p{=}192$, $n{=}200$ & 144 & 9.4 & 0.12 & 0.00 & 0.00 & 8.5\\
$p{=}12$, $n{=}400$ & 9 & 9.0 & 0.00 & 1.00 & 1.00 & 9.0\\
$p{=}24$, $n{=}400$ & 18 & 18.0 & 0.00 & 1.00 & 1.00 & 18.0\\
$p{=}48$, $n{=}400$ & 36 & 36.3 & 0.00 & 1.00 & 0.73 & 36.0\\
$p{=}96$, $n{=}400$ & 72 & 77.5 & 0.00 & 0.98 & 0.00 & 72.0\\
$p{=}192$, $n{=}400$ & 144 & 100.0 & 1.00 & 0.00 & 0.00 & 93.8\\
\bottomrule
\end{tabular}
\end{table}
{}

\section{Superconductivity study}\label{sec:real}

\subsection{Data, grouping, and evaluation protocol}
\label{sec:uci-design}

We use the UCI \emph{Superconductivity Data} set
\citep{hamidiehuci}, distributed under CC BY 4.0. It contains
$n=21{,}263$ compounds, $p=81$ composition-derived predictors, and the
observed critical temperature \texttt{critical\_temp}, measured in kelvin.
The source study describes the feature construction and original prediction
task \citep{hamidieh18}.
The predictors comprise the number of constituent elements and summary
statistics from eight physical-property groups: atomic mass, first
ionization energy, atomic radius, density, electron affinity, fusion heat,
thermal conductivity, and valence.
For block-level diagnostics we call the singleton number-of-elements feature
the composition-count block and the eight ten-coordinate property groups the
physical-property blocks. Together they form nine predefined descriptor
blocks.

A random row split would be misleading for these data. There are 2,418
duplicate exact-predictor groups involving 8,511 rows; in 2,374 of those
groups, involving 8,423 rows, the recorded critical temperatures are not
identical. We therefore form connected components linking rows that share
an exact material string, an exact elemental-composition vector, or an
exact 81-dimensional predictor vector. This produces 15,170 components.
Components, rather than rows, are assigned intact to five folds, with a
fixed-seed greedy allocation balancing fold size and observed-temperature
decile counts.

Once this fold manifest is frozen, all predictor centers and scales and,
when applicable, response centers and scales are estimated from the current
outer-training sample only. We use coordinatewise medians and MADs, with the
sample standard deviation as a prespecified fallback for a non-positive MAD.
Observed-temperature deciles were used only in the pre-fit balancing of
whole components described above, and after that allocation outer-test
responses are used only for evaluation.

In addition, all five outer-training predictor matrices had slope rank 81 and
augmented rank 82 after training-only median/MAD transformation, a
finite-sample fact rather than the asymptotic well-conditioning of
Assumption~\ref{ass:1}. They were nonetheless
substantially ill-conditioned, with a two-norm condition number of the
augmented training design ranging from $2.76\times10^3$ to $7.21\times10^4$.
This geometry is
consistent with coordinate substitution among correlated descriptors
and motivates the block-level audit.

For each outer fold, we consider clean training responses and nested
$5\%$ and $10\%$ contamination sets. The contaminated responses receive
a one-sided shift of eight clean-training response MADs, and test responses
are never modified. We compute RMSE, MAE, and $R^2$ on every untouched test
row. The main five-fold tables report RMSE, while all three metrics are archived;
fold-to-fold standard deviations are descriptive
(Section~\ref{sec:uci-limitations}).

Turning to implementation, the D-MM and SCAD-raw fits use $a=3.7$ and 250 decreasing values of
$\lambda$, from the global-coordinate-null value to $10^{-6}$ times that
value. D-MM uses $\kappa=2$ in its distillation-fidelity
GIC. SCAD-raw uses its response-based BIC, and Lasso-CV uses group-aware
five-fold inner cross-validation that also keeps the connected components
intact. Among the baselines, OLS, SCAD-raw, and Lasso-CV fit the response without
robustness protection, MM supplies the robust initial fitted surface, and
D-MM is its distilled counterpart. Integrated robust-sparse methods
are compared under known truth in Section~\ref{sec:sim-modern}.
We use standardized slopes for within-fit
interpretation and for cross-fold coefficient comparison, and
coefficient-stability calculations use the dimensionless
training-standardized slopes.

A targeted post-production check on clean outer fold 2 separated path
density from endpoint depth. Merely densifying a shallow grid did not
resolve truncation, so we read shallow endpoint-selected models as
truncation artifacts, not evidence of compactness (\provref{}).

\subsection{Observed critical temperature: a five-fold analysis}
\label{sec:uci-observed}

The observed response provides no gold-standard support. We therefore report
out-of-fold predictive error, distillation fidelity, model size, and coordinate
stability, but not support recovery. For every method--scenario combination,
the out-of-fold series contains all 21,263 observations exactly once.

\begin{table}[htbp]
\centering
\scriptsize
\caption{Five-fold observed critical-temperature audit. Entries are outer-fold
means with standard deviations in parentheses. Errors are in kelvin and use
untouched test responses; size excludes the intercept. Fidelity RMSE compares
D-MM with its MM initial fit using predictions on each outer-test predictor
design, without using test responses. D-MM uses the primary
$\kappa=2$ GIC, SCAD-raw uses response-based BIC, and Lasso-CV uses the
expanded path.}
\label{tab:uci-observed}
\setlength{\tabcolsep}{3.5pt}
\begin{tabular}{@{}llccc@{}}
\toprule
Training response & Method & RMSE & Size & Fidelity RMSE\\
\midrule
Clean
  & OLS      & $17.693\;(0.392)$ & $81.0\;(0.0)$ & --\\
  & MM       & $17.950\;(0.356)$ & $81.0\;(0.0)$ & --\\
  & D-MM     & $18.009\;(0.400)$ & $66.8\;(4.3)$ & $1.580\;(0.698)$\\
  & SCAD-raw & $17.727\;(0.414)$ & $66.6\;(2.8)$ & --\\
  & Lasso-CV & $17.704\;(0.403)$ & $81.0\;(0.0)$ & --\\
\addlinespace
$5\%$, $+8$ MAD
  & OLS      & $20.642\;(0.377)$ & $81.0\;(0.0)$ & --\\
  & MM       & $17.871\;(0.367)$ & $81.0\;(0.0)$ & --\\
  & D-MM     & $17.889\;(0.370)$ & $68.2\;(1.1)$ & $0.936\;(0.077)$\\
  & SCAD-raw & $21.006\;(0.238)$ & $39.6\;(5.0)$ & --\\
  & Lasso-CV & $20.659\;(0.349)$ & $75.2\;(5.9)$ & --\\
\addlinespace
$10\%$, $+8$ MAD
  & OLS      & $27.330\;(0.294)$ & $81.0\;(0.0)$ & --\\
  & MM       & $17.808\;(0.372)$ & $81.0\;(0.0)$ & --\\
  & D-MM     & $17.824\;(0.375)$ & $68.8\;(1.5)$ & $0.886\;(0.101)$\\
  & SCAD-raw & $27.738\;(0.333)$ & $29.4\;(3.7)$ & --\\
  & Lasso-CV & $27.308\;(0.238)$ & $69.8\;(5.0)$ & --\\
\bottomrule
\end{tabular}
\end{table}

\begin{figure}[!htbp]
\centering
\includegraphics[width=\linewidth]{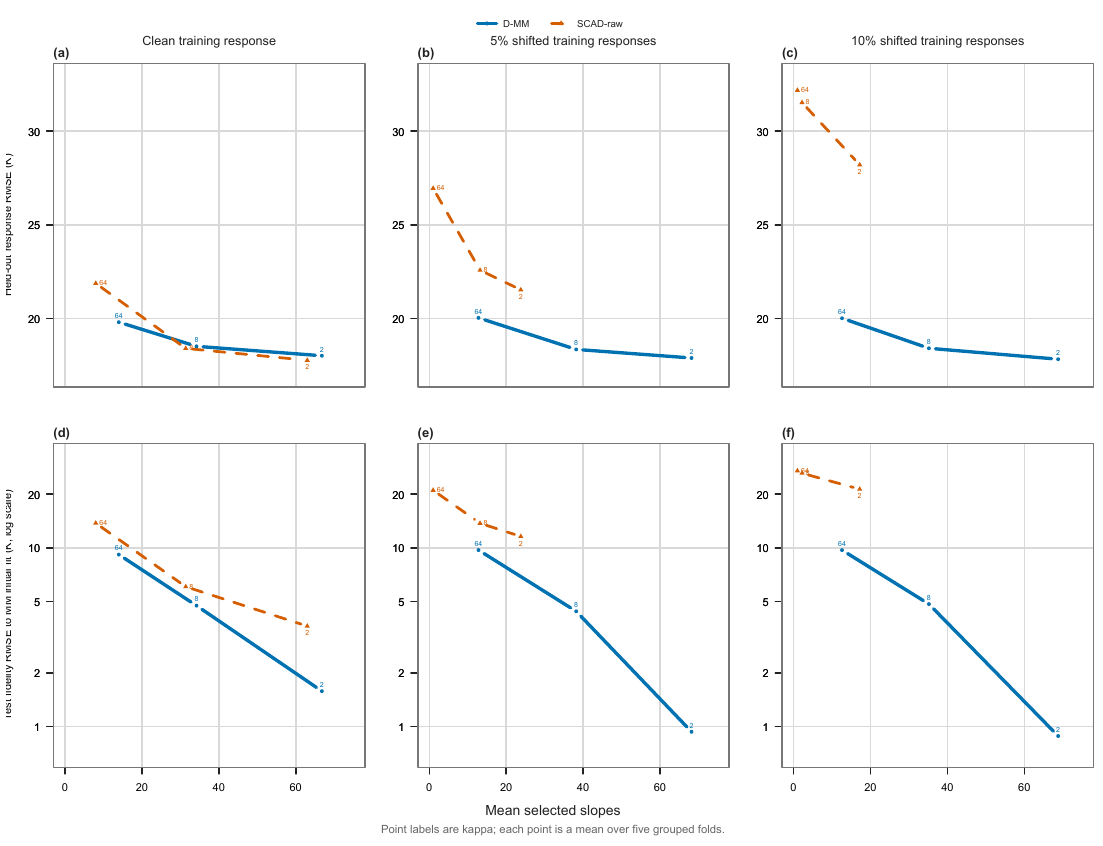}
\caption{Observed-response size--prediction--fidelity frontier. The horizontal
axis is the mean number of selected slopes over five duplicate-grouped outer
folds. The upper panels show held-out response RMSE and the lower panels show
test-design RMSE relative to the fold-specific MM initial fit; lower-panel axes
are logarithmic. Point labels give the archived multiplier $\kappa$.
For D-MM this is the distillation-fidelity GIC multiplier; for SCAD-raw it is the
generalized-BIC complexity multiplier, and the primary multiplier-one BIC
fit is not plotted. Connected points trace the prespecified sensitivity
grid.}
\label{fig:uci-frontier}
\end{figure}

In Table~\ref{tab:uci-observed}, D-MM at the primary $\kappa=2$ tracks MM
closely in all three training scenarios, whereas OLS, SCAD-raw, and Lasso-CV
degrade by roughly 10 K in test RMSE under the $10\%$ shift. The
expanded Lasso path selects its requested lower endpoint in six of the 15
fits and no upper endpoint. Four such selections occur under clean training
and two at $5\%$. The fifth clean fit is interior but also retains all 81
slopes, so the clean result reflects strong endpoint pressure and does not
demonstrate a sparse optimum. Raising the GIC multiplier from
$\kappa=2$ to $\kappa=64$ cuts D-MM's mean size from 66.8--68.8 slopes to
12.6--14.0 at the price of sharply higher response and fidelity RMSE; the
observed response thus supports a robust fitted surface more clearly than a
small coordinate-level explanation.
Figure~\ref{fig:uci-frontier} displays the same multi-criteria trade-off and
also shows how raw-response SCAD moves under contamination.

In the numerical audit of all five folds, every primary selected state is
accepted, interior, and exactly linked to its retained path row. The
deterministic fidelity certificate of Proposition~\ref{prop:norm}(a) was
needed only for the $\kappa=64$ states. \provref{} records the gaps,
validation gates, archive counts and the sublevel-condition check.

Overlap-conditioned stability (\ORtab{tab:uci-observed-stability}) is
mixed. The two shifted fits agree closely with each other, whereas
clean-trained fits agree only partially with them, especially among the
ten largest-magnitude coefficients. The instability sits where the design
admits substitution, namely in the two atomic-radius summaries, which act
as a nearly interchangeable, opposite-signed pair between which
clean-trained fits alternate.

\begin{table}[htbp]
\centering
\scriptsize
\caption{Overlap-conditioned stability of observed-response D-MM.
Between-fold rows summarize the ten pairs of outer folds under a common
training response; within-fold rows summarize the five folds, comparing
training responses inside each fold. Support J is the Jaccard index between
selected supports, Coef.\ corr.\ is the all-coordinate correlation of
training-standardized slopes, and Top-ten J compares the sets of ten
largest-magnitude selected standardized coefficients within each fit; the
latter is distinct from the global across-fit ranking in
Table~\ref{tab:uci-observed-coordinate}. Entries are means, with minima in
parentheses for the clean between-fold row and the clean-versus-$10\%$ row.}
\label{tab:uci-observed-stability}
\setlength{\tabcolsep}{4pt}
\begin{tabular}{@{}llccc@{}}
\toprule
 & Comparison & Support J & Coef.\ corr. & Top-ten J\\
\midrule
Between folds
 & Clean & 0.856 (0.773) & 0.664 (0.414) & 0.621 (0.429)\\
 & $5\%$ & 0.903 & 0.986 & 0.714\\
 & $10\%$ & 0.878 & 0.984 & 0.702\\
\addlinespace
Within fold
 & Clean vs.\ $5\%$ & 0.871 & 0.722 & 0.593\\
 & Clean vs.\ $10\%$ & 0.873 (0.781) & 0.721 (0.516) & 0.510 (0.250)\\
 & $5\%$ vs.\ $10\%$ & 0.941 & 0.993 & 0.861\\
\bottomrule
\end{tabular}
\end{table}
{}

\begin{table}[htbp]
\centering
\scriptsize
\caption{Leading-coordinate audit for observed-response D-MM. Coordinates are
ranked by mean absolute training-standardized coefficient over all 15 fits,
using the prespecified deterministic ranking rule. C, 5, and 10 give selected
folds out of five for clean, $5\%$, and $10\%$ training responses; All is out
of 15. Mean (SD) includes numerical zeros.}
\label{tab:uci-observed-coordinate}
\setlength{\tabcolsep}{3pt}
\begin{tabular}{@{}lrrrrr@{}}
\toprule
Coordinate & C & 5 & 10 & All & Standardized coefficient\\
\midrule
\texttt{wtd\_gmean\_atomic\_radius}       & 2/5 & 5/5 & 5/5 & 12/15 & $-2.219\;(1.277)$\\
\texttt{wtd\_mean\_atomic\_radius}        & 5/5 & 5/5 & 5/5 & 15/15 & $ 1.900\;(0.935)$\\
\texttt{entropy\_Valence}                  & 5/5 & 5/5 & 5/5 & 15/15 & $ 0.818\;(0.150)$\\
\texttt{wtd\_mean\_ThermalConductivity}   & 5/5 & 5/5 & 5/5 & 15/15 & $ 0.697\;(0.088)$\\
\texttt{wtd\_entropy\_Valence}             & 5/5 & 5/5 & 5/5 & 15/15 & $-0.692\;(0.087)$\\
\texttt{wtd\_gmean\_ElectronAffinity}      & 5/5 & 5/5 & 5/5 & 15/15 & $-0.677\;(0.128)$\\
\texttt{range\_atomic\_mass}               & 5/5 & 5/5 & 5/5 & 15/15 & $ 0.591\;(0.101)$\\
\texttt{std\_fie}                           & 5/5 & 5/5 & 5/5 & 15/15 & $-0.491\;(0.099)$\\
\texttt{std\_ElectronAffinity}              & 5/5 & 5/5 & 5/5 & 15/15 & $ 0.474\;(0.075)$\\
\texttt{std\_ThermalConductivity}           & 5/5 & 5/5 & 5/5 & 15/15 & $ 0.453\;(0.082)$\\
\bottomrule
\end{tabular}
\end{table}
{}

Even so, nine leading coordinates are selected with a constant sign in all
15 fits (\ORtab{tab:uci-observed-coordinate}).
The highest-ranked coordinate is absent from three clean fits but negative in
all 12 selections. Conditional on selection, its standardized coefficient is
$-2.773\;(0.630)$. Below this leading set lies a dense, partially
unstable tail of roughly 57--60 further slopes, and strong correlation among
summaries from the same physical-property blocks precludes a physical-effect
reading of the individual coefficients, a difficulty well documented for
correlated features
\citep{tolosilengauer11}. Overall, D-MM preserves the held-out accuracy of the
robust initial fit under all three training scenarios, while a compact and
stable coordinate-level explanation is not supported by these data.

\ucixheading{UCI-X: actual predictors with known support}
\label{sec:uci-x}

UCI-X is a design-based semi-synthetic experiment.  The complete
81-column UCI predictor matrix, its physical-property blocks, and the
duplicate-aware outer-fold membership are observed. Only the response is
generated, so that variable- and block-level recovery can be audited against
known truth.  Observed \texttt{critical\_temp} plays no role in UCI-X beyond
the outer-fold balancing frozen in Section~\ref{sec:uci-design} of the
main text.

Let $m_j$ and $s_j$ denote the full fixed-design median and MAD used only
to define the data-generating law. We generate
\[
  y_i^\star
  =
  \sum_{j\in\mathcal S_0}
  a_j\frac{x_{ij}-m_j}{s_j}
  -
  \overline{\sum_{j\in\mathcal S_0}
  a_j\frac{x_{ij}-m_j}{s_j}}
  +\varepsilon_i,
  \qquad
  \varepsilon_i\stackrel{\mathrm{iid}}{\sim}N(0,\sigma^2),
\]
where the target signal-to-noise ratio is four and the fixed realization has
SNR 4.046. Here SNR means
$\operatorname{Var}(\eta_i)/\sigma^2$, where $\eta_i$ is the centered
linear signal above. The archived law uses $\sigma=2.1578077$, giving the
realized fixed-design variance ratio 4.0458866. The eight-variable support
contains one descriptor from each
physical-property block:
\[
\begin{aligned}
\mathcal S_0=\{&
\texttt{wtd\_mean\_atomic\_mass},
\texttt{range\_fie},\\
&
\texttt{wtd\_std\_atomic\_radius},
\texttt{entropy\_Density},\\
&
\texttt{wtd\_mean\_ElectronAffinity},
\texttt{gmean\_FusionHeat},\\
&
\texttt{wtd\_range\_ThermalConductivity},
\texttt{wtd\_mean\_Valence}\},
\end{aligned}
\]
with standardized coefficients
$(1.20,-1.10,1.00,-0.95,0.90,-0.85,0.80,-0.75)$ in that order.
Because the experiment conditions on the observed predictor matrix, this
full-design normalization is a prespecified constant of the response law,
not estimator preprocessing.  Every fitted estimator is still centered
and scaled from its current outer-training sample only.

\begin{table}[htbp]
\centering
\scriptsize
\caption{Five-fold UCI-X prediction and variable-level support recovery.
RMSE is reported as fold mean (fold SD); the remaining entries are fold
means. Truth-J denotes the Jaccard index between the selected support and
$\mathcal S_0$, and Exact is the proportion of folds with exact support.}
\label{tab:uci-x-performance}
\setlength{\tabcolsep}{2.6pt}
\begin{tabular}{@{}llrrrrrr@{}}
\toprule
Training response & Method & RMSE (SD) & Size & TPR & FPR &
  Truth-J & Exact\\
\midrule
\UCIXPerformanceRows\\
\bottomrule
\end{tabular}
\end{table}

\begin{table}[htbp]
\centering
\scriptsize
\caption{Explanation stability in UCI-X at two granularities. Panel (a) is
coordinate level: Support-J is the Jaccard index between two fitted
supports, and True-J and FP-J apply the same comparison to selected true and
selected noise variables; True-J measures stability among selected true
coordinates, not recall, with completeness reported by TPR in
Table~\ref{tab:uci-x-performance}. Coefficient correlation uses all 81
training-standardized slopes. Panel (b) is block level: a block is selected
when at least one of its coordinates is, the truth contains the eight
physical-property blocks, and B-FPR is the fraction of folds selecting the
single null composition-count block. Mass corr.\ is the mean Pearson
correlation between the nine-dimensional vectors of within-block $L_2$
norms, which measures unsigned magnitude allocation rather than signed or
within-block coordinate stability. Jaccard and correlation entries summarize
the ten pairs of outer-training folds; the remaining block entries are fold
means. Only the three sparse explanation methods are shown.}
\label{tab:uci-x-stability}
\setlength{\tabcolsep}{3.0pt}
\begin{tabular}{@{}llrrrrr@{}}
\multicolumn{7}{@{}l}{\emph{(a) Coordinate level}}\\
\toprule
Training response & Method & Mean J & Min J &
  Mean true-J & Mean FP-J & Coef.\ corr.\\
\midrule
\UCIXStabilityRows\\
\bottomrule
\end{tabular}

\vspace{4pt}

\begin{tabular}{@{}llrrrrrr@{}}
\multicolumn{8}{@{}l}{\emph{(b) Block level}}\\
\toprule
Training response & Method & B-size & B-TPR & B-FPR &
  B-exact & B-J & Mass corr.\\
\midrule
\UCIXBlockRows\\
\bottomrule
\end{tabular}
\end{table}

\begin{figure}[!htbp]
\centering
\includegraphics[width=0.94\linewidth]{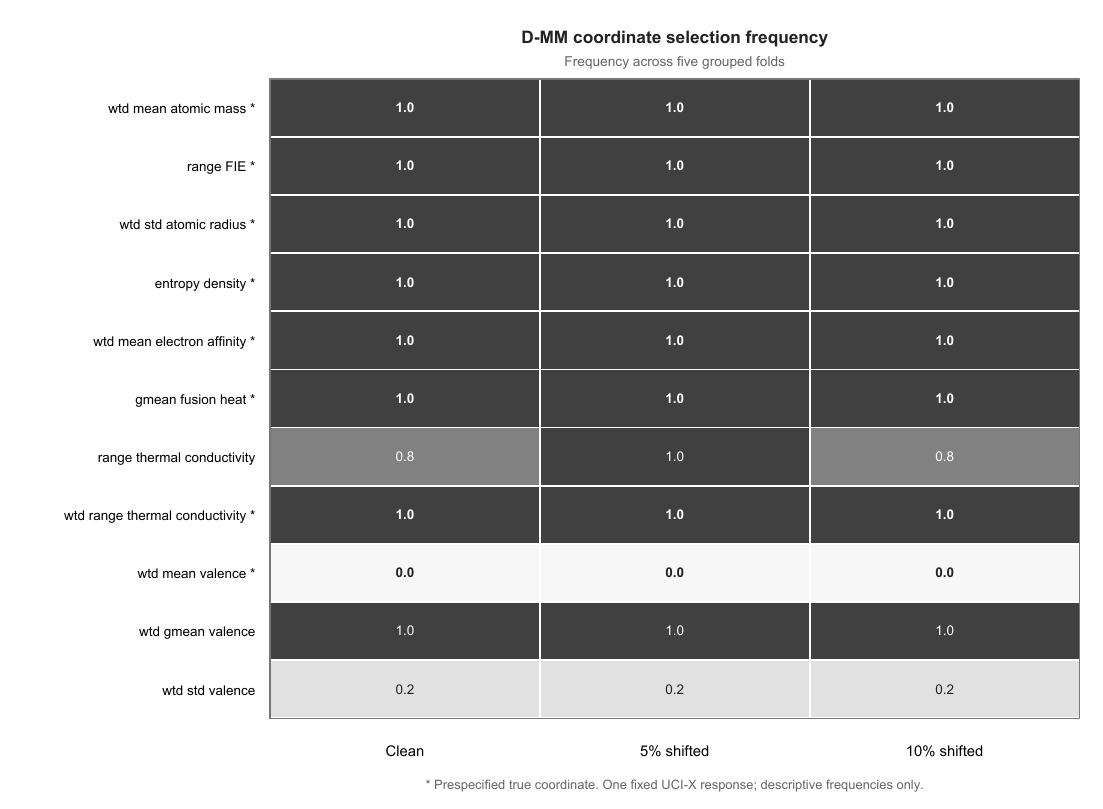}
\caption{D-MM coordinate-selection frequencies in UCI-X. Asterisks mark the
eight prespecified true coordinates; rows also include every non-support
coordinate selected at least once. Each cell is a frequency across five
duplicate-grouped outer folds for one fixed semi-synthetic response
realization.}
\label{fig:ucix-frequency}
\end{figure}

We used outer fold 1 before production only to lock the expanded SCAD path
range. Tables~\ref{tab:uci-x-performance}--\ref{tab:uci-x-stability} and
Figure~\ref{fig:ucix-frequency} report the subsequently validated production
summaries across all five outer folds.

\UCIXFiveFoldInterpretation

To reduce dependence on the production UCI-X noise realization, we
repeated D-MM for four additional independent Gaussian noise realizations while
holding the predictor matrix, true support and coefficient construction,
duplicate-grouped folds, fitting seeds, SCAD path, and contamination plan
fixed. Table~\ref{tab:uci-x-noise-repeatability} summarizes the
production realization together with the four additional realizations.

\begin{table}[!htbp]
\centering
\scriptsize
\caption{D-MM repeatability across five UCI-X Gaussian noise realizations.
Each entry is the mean of five seed-specific five-fold means; parenthesized
values for RMSE, Size, and Truth-J are SDs across those five noise seeds, not
standard errors across independent folds. Exact is the mean exact-support rate.}
\label{tab:uci-x-noise-repeatability}
\setlength{\tabcolsep}{4pt}
\begin{tabular}{@{}lrrrrrr@{}}
\toprule
Training response & RMSE (SD) & Size (SD) & TPR & FPR & Truth-J (SD) & Exact\\
\midrule
Clean & 2.153 (0.015) & 10.40 (1.54) & 0.875 & 0.047 & 0.624 (0.074) & 0.000\\
$10\%$, $+8$ MAD & 2.153 (0.015) & 10.88 (1.76) & 0.875 & 0.053 & 0.609 (0.081) & 0.000\\
\bottomrule
\end{tabular}
\end{table}

Across same-fold pairs of distinct noise seeds (50 pairs per scenario), mean
support Jaccard was 0.777 (minimum 0.500) under clean training and 0.748
(minimum 0.563) under $10\%$ shifts. The corresponding mean original-unit
all-slope coefficient correlations were 0.9987 and 0.9983. Across all 50 fits,
the same seven planted coordinates and the correlated
\texttt{wtd\_gmean\_Valence} substitute were selected, whereas
\texttt{wtd\_mean\_Valence} was never selected. Complete supports still varied
through extra selections. All 50 selected states were interior and accepted,
every 250-point path reached its $10^{-6}$ endpoint, and the largest selected-
state fixed-point gap was $1.13\times 10^{-5}$. In addition, all 15
production-run primary UCI-X D-MM states satisfied the initial-fit sublevel
condition in Proposition~\ref{prop:norm}(a), so its deterministic empirical-
design fidelity certificate, relative to the initial fit on the fixed design,
applies to these retained states.

To check whether the coordinate-level substitution also occurs outside
the distillation pipeline, we ran the
integrated RLARS-MM comparator of Section~\ref{sec:sim-modern} on the
identical archived design. The run used the same folds, the same clean and
shifted training responses, and the same planted support, under robust BIC
with model-size budgets of 20 and 40. The comparator operates at a different
parsimony altogether: at both budgets the selected size fills most of the
allowance (means $19.0$--$20.0$ and $38.6$--$39.6$ of 81), exact recovery
is zero in all 30 fits, and mean test RMSE ranges over $2.38$--$3.07$
against $2.147$--$2.148$ for D-MM. The valence geometry expresses itself
the same way even so: every clean fit at both budgets admits the
correlated \texttt{wtd\_gmean\_Valence}, as do nine of the ten
$5\%$-shift fits, the planted \texttt{wtd\_mean\_Valence} enters in at
most four of five folds in any scenario, and under the $10\%$ shift the
comparator loses the valence block entirely. Thus the substitution is
also observed for the investigated alternative method, consistent with
the near-collinearity of the design; it is not established for every
possible selector. Four of the 30 final MM refits did not converge and are
recorded, with the complete per-fit supports, in the archived run
artifacts.

\subsection{Scope of the superconductivity evidence}\label{sec:uci-limitations}

UCI-X supplies exact truth while preserving a
modern, correlated, 81-predictor real design, but its response is
semi-synthetic and cannot establish physical prediction accuracy for
superconducting temperature.
The experiment is also conditional on one predictor matrix, the production
Gaussian noise realization plus four additional realizations, a single
signal-to-noise ratio and coefficient pattern, and one-sided vertical
shifts. Only D-MM was repeated across the five responses; the
comparator audit remains conditional on the production response. In addition,
the study does not evaluate contaminated predictor rows or leverage, which
remain outside the fixed-design robustness result. Grouping by connected components keeps
exact material, composition, and predictor duplicates within a fold, but
it cannot guarantee independence among scientifically related compounds.
The block audit coarsens the 81 coordinates to eight active blocks and a
single null block, so block-level false-positive evidence is necessarily
coarse; it is a lenient, DGP-specific granularity diagnostic
rather than an independent physical validation.
Any two outer-training sets also share three fifths of the full data.
Accordingly, their support and coefficient similarities are overlap-conditioned
descriptive diagnostics and may overstate stability under genuinely
independent resampling. Finally, fold 1 was used for computational
path-range diagnostics. For these reasons we treat all five-fold summaries
as descriptive fixed-design evidence; the displayed patterns are in any
case robust to excluding the diagnostic fold (\provref{}).

\FloatBarrier

\section{Conclusions and discussion}\label{sec:disc}

In this paper we have proposed penalized distillation, a modular way of
turning a robust linear initial fit into a sparse one. The accompanying
assessment documents whether a particular explanation claim is
substantiated for the stated design, tuning rule, perturbations, and
granularity, where a negative verdict binds only the evaluated candidates at
that granularity.

The guarantees of Section~\ref{sec:theory} are all conditional on the
retained path states and a fixed uncontaminated design. Under this
conditioning, the distilled estimator inherits the initial fit's
response-replacement breakdown bound, and on the oracle-support branch it is
the empirical-Gram projection of the initial fit, with the corresponding
efficiency gain and influence identity. The GIC, in turn, selects that
support consistently along a qualifying computed path. In the evaluated
regimes, D-MM matches or improves on the risk of the integrated robust-sparse
procedures in Table~\ref{tab:modern}. Its reported complete-workflow times
are smaller than those of RLARS-MM, adaptive PENSE and sparse LTS, but
these are not equal-thread speedup estimates. In the
superconductivity study of Section~\ref{sec:real} we found that a
support can be stable across folds and contamination levels yet wrong at
the coordinate level, while block-level recovery is
exact.

The simulations locate two boundaries in the evaluated designs. The
full-coordinate robust initial fit deteriorates at moderate $p/n$, while
screening removes that failure in the sparse-signal dimension study.
A screened fit fails in turn once the number of nonzero slopes
approaches the cap $K$, and past that point the full-coordinate route is
again preferable; there the distillation stage even costs accuracy against
its own initial fit. Between the two boundaries, where the screen retains
the signal but over-selects around it, the sparse stage recovers the
support that the screened fit alone does not, with exact recovery in
$97\%$ of replications against $0\%$ for the initial fit alone
(Section~\ref{sec:screen-sim}). Accordingly, which route to take depends on
the design and the signal, and the cap is the one quantity the user has to set.

At the same time, the present work has three main limitations. The theory is fixed-dimensional, all robustness
transfer statements condition on an uncontaminated design, and the UCI
experiments condition on one predictor matrix and one-sided vertical shifts.
The observed-response study's remaining design-stage
dependencies are stated in Section~\ref{sec:uci-limitations}.
We see three directions for future work. First, growing-dimensional and
structured-group theory remains to be developed. Second, the fixed cap should
be replaced by a data-driven rule that adapts to the signal density. Third,
selectors that recognize interchangeable correlated descriptors are needed.
Extensions that protect the second-stage geometry against contaminated
predictor rows, and surrogate classes for nonlinear initial estimators
whose fitted surfaces a sparse linear surrogate cannot represent, are
further open problems.

\section*{Supporting material}

This integrated preprint includes the proofs, extended simulation studies,
full dimension-study tables, UCI-X semi-synthetic study, nonparametric
selection-rule check, computational provenance, and implementation details
in its main text and appendices. No separate supplementary PDF is needed
to read these materials.

\section*{Data and code availability}

The UCI \emph{Superconductivity Data} set is publicly available under CC BY
4.0 at \url{https://doi.org/10.24432/C53P47}.
The analysis code and supporting numerical records, including row-level
predictions, are maintained by the authors and can be provided for
editorial and peer-review assessment upon request. No public repository
has yet been established. These archives include environment information,
validation and provenance records, the duplicate-aware fold manifest,
signed training-standardized coefficient tables, file-to-result maps,
checksums, and reproduction commands. They are distinct from, and are not
included in, this preprint or its arXiv manuscript source package.

\section*{Statements and declarations}

\subsection*{Funding}
\noindent\textbf{Seunghwan Park.} This work was supported by the National
Research Foundation of Korea (NRF) grant funded by the Korean government
(MSIT) (No. RS-2026-25489716).

\smallskip
\noindent\textbf{Wooyoung Shin.} This work was supported by the National
Research Foundation of Korea (NRF) grant funded by the Korean government
(MSIT) (No. RS-2026-25587171).

\subsection*{Competing interests}
The authors have no relevant financial or non-financial interests to
disclose.

\subsection*{Author contributions}
Both authors contributed equally to this work overall. Seunghwan Park
contributed to the theoretical development and took primary responsibility
for writing the manuscript. Wooyoung Shin contributed to the simulation
studies and the theoretical development.

\subsection*{Author identifiers and contact information}
\noindent Wooyoung Shin:
\href{mailto:sinwy93@kangwon.ac.kr}{\texttt{sinwy93@kangwon.ac.kr}};
ORCID \href{https://orcid.org/0009-0003-3349-3905}{0009-0003-3349-3905}.

\smallskip
\noindent Seunghwan Park (corresponding author):
\href{mailto:stat.shpark@kangwon.ac.kr}{\texttt{stat.shpark@kangwon.ac.kr}};
ORCID \href{https://orcid.org/0009-0007-5238-8368}{0009-0007-5238-8368}.

\appendix

\section{Proofs}\label{app:proofs3}

\providecommand{\bscr}{\tilde{\bm{\beta}}^{\mathrm{scr}}}
\providecommand{\Sscr}{\widehat{S}}

\subsection{Why retained path states inherit a conditional response-replacement bound}\label{sec:mech}

Write the second-stage objective as
$Q_\lambda(\bbeta)=\frac{1}{2n}\|\X\btil-\X\bbeta\|^2+\sum_j
p_\lambda(|\beta_j|)$. The transfer argument is a chain of three links.

The first link is the one already stated: $\y$ appears in $Q_\lambda$
nowhere except inside $\X\btil$. An adversary who replaces $m$ of the
responses by arbitrary values can therefore influence the second stage
only by moving that single vector. The GIC additionally uses the operational
scalar $\hat\sigma$, but the deterministic path bound below does not require
that scalar to be consistent.

The second link converts a bound on the initial estimator into a bound on
the objective's geometry. Suppose the initial intercept--slope vector
$\tilde\theta=(\tilde\alpha,\btil^\top)^\top$ is bounded uniformly
under those $m$ response replacements with $\X$ held fixed, which is exactly the
conditional response-only property formalized in Section~\ref{sec:finite}.
Then $\|\X\btil\|$ is bounded, so $Q_\lambda(\bm0)=\frac1{2n}\|\X\btil\|^2$
is bounded, and the sublevel set $\{\bbeta:Q_\lambda(\bbeta)\le
Q_\lambda(\bm0)\}$ is a bounded set. Because $\X$ has full column rank the
quadratic term is coercive, so any $\bbeta$ far from the origin makes it
large, and the penalty, being nonnegative, cannot compensate. Proposition~\ref{prop:norm}(b) makes this quantitative, with
the explicit constant $2\sqrt{s_{\max}/s_{\min}}$.

The third link is computational. A bound on a sublevel set constrains only
points that actually lie in it, whereas the nonconvex path computation
returns local solutions. Our path audits every coordinate-descent candidate at
the current $\lambda$ and accepts it only when its objective is no larger
than the warm start's. Lemma~\ref{lem:warm} then certifies the null-model
sublevel condition by construction. Corollary~\ref{cor:bp} applies to every
retained path state and hence to any rule, including the stated GIC, that
selects among them.

One direction of contamination escapes the chain entirely. Contaminated
\emph{design rows} $\bm x_i$ survive into the second-stage design, since the
first link removes $\y$ from the objective but leaves $\X$ where it is, so a
bad leverage point distorts both stages. Section~\ref{sec:if}
shows the same gap analytically, as a term that vanishes at the model but
need not vanish and can grow quadratically in $\|\bm x_0\|$ at a general
off-model baseline. A weighted empirical variant, defined in
Section~\ref{sec:init}, is a possible mitigation.

\subsection{Proof of Proposition~\ref{prop:norm}, Lemma~\ref{lem:warm} and
Corollary~\ref{cor:bp}}\label{app:norm}

\emph{Part (a).} Suppose
$Q_\lambda^+(\hat\theta)\le Q_\lambda^+(\tilde\theta)$. The quadratic
part of $Q_\lambda^+$ vanishes at $\tilde\theta$, so
$Q_\lambda^+(\tilde\theta)=\sum_jp_\lambda(|\tilde\beta_j|)$, and
dropping the nonnegative penalty at $\hat\theta$ leaves
\[
\tfrac1{2n}\|\mathbf Z(\tilde\theta-\hat\theta)\|^2
\;\le\;\sum_jp_\lambda(|\tilde\beta_j|).
\]
The SCAD penalty is bounded above by its plateau value
$(a+1)\lambda^2/2$ for every argument, so the right side is at most
$p(a+1)\lambda^2/2$ regardless of how large $\tilde\theta$ is. On the
left, $\tfrac1n\|\mathbf Z\bm v\|^2\ge s_{\min}\|\bm v\|^2$ for every
$\bm v$ by the definition of $s_{\min}$. Combining the two gives
$\|\hat\theta-\tilde\theta\|_2^2
\le p(a+1)\lambda^2/s_{\min}$, which is the claim.
Note where the boundedness of the penalty was used: for the lasso the
right side would grow with $\|\btil\|_1$ and no such uniform bound would
follow.

\emph{Part (b).} Suppose instead
$Q_\lambda^+(\hat\theta)\le Q_\lambda^+(\bm0)$. Since
$p_\lambda(0)=0$, the right side is
$\tfrac1{2n}\|\mathbf Z\tilde\theta\|^2$, and discarding the penalty at
$\hat\theta$ gives
$\|\mathbf Z(\tilde\theta-\hat\theta)\|
\le\|\mathbf Z\tilde\theta\|$. The triangle inequality then yields
$\|\mathbf Z\hat\theta\|
\le\|\mathbf Z\tilde\theta\|
+\|\mathbf Z(\tilde\theta-\hat\theta)\|
\le2\|\mathbf Z\tilde\theta\|$. Converting both sides with the extreme
eigenvalues of $\mathbf Z^\top\mathbf Z/n$ gives
$\|\hat\theta\|_2\le2\sqrt{s_{\max}/s_{\min}}\,
\|\tilde\theta\|_2$. This bound transfers boundedness of the full initial-fit
coefficient, including the intercept, to every qualifying candidate state.

\emph{The warm-start lemma.} It remains to check that the hypothesis of
part (b) is met along the safeguarded computed path. Differentiating the
SCAD penalty with respect to $\lambda$ on its three branches gives $t$ on
$[0,\lambda]$, $(at-\lambda)/(a-1)$ on $[\lambda,a\lambda]$, and
$(a+1)\lambda$ beyond; all three are nonnegative, and they agree at the
two knots, so $\lambda\mapsto p_\lambda(t)$ is nondecreasing and continuous
for each fixed $t$. Let $\lambda_1>\dots>\lambda_K$ be the grid, start at
the intercept-only vector $\theta_0$, and let $\hat\theta_k$ denote the
\emph{retained} full vector at $\lambda_k$. By construction, a finite converged
candidate is accepted only if
$Q_{\lambda_k}^+(\hat\theta_k)
\le Q_{\lambda_k}^+(\hat\theta_{k-1})$; if that audit fails, the algorithm
sets $\hat\theta_k=\hat\theta_{k-1}$, so the same inequality holds with
equality. At $k=1$,
$Q_{\lambda_1}^+(\hat\theta_1)
\le Q_{\lambda_1}^+(\theta_0)\le Q_{\lambda_1}^+(\bm0)$ because
$\theta_0$ is the intercept-only least-squares fit. For $k>1$, the inductive
step is
\[
Q_{\lambda_k}^+(\hat\theta_k)
\;\le\;Q_{\lambda_k}^+(\hat\theta_{k-1})
\;\le\;Q_{\lambda_{k-1}}^+(\hat\theta_{k-1})
\;\le\;Q_{\lambda_{k-1}}^+(\theta_0)
\;=\;Q_{\lambda_k}^+(\theta_0)
\;\le\;Q_{\lambda_k}^+(\bm0),
\]
where the first inequality is the explicit acceptance safeguard (including
the retain-the-start rule on failure), the second follows because
$\lambda_k<\lambda_{k-1}$ and the penalty is nondecreasing in $\lambda$,
the third is the inductive hypothesis, the equality holds because
$\theta_0$ has zero slopes, and the final inequality is its least-squares
property. Hence every retained path
state satisfies the sublevel-set condition, with no descent or
global-optimum property of the solver assumed. \qed

\emph{The breakdown corollary.} Fix an integer $m$ such that
$m/n<\varepsilon^*_{n,y}(\tilde\theta;\y\mid\X)$. By the definition of $\varepsilon^*_{n,y}$ in Section~\ref{sec:finite},
the initial estimator's full intercept--slope vector is uniformly bounded over all
$\y'\in\mathcal Y_m(\y)$ while $\X$ remains unchanged. Consequently the
augmented design and the eigenvalue multiplier in
Proposition~\ref{prop:norm}(b) are fixed, so the retained full coefficient
vector at every position of the finite grid is uniformly bounded over the
same response-replacement neighborhood. Any selector among those states is
bounded there as well. Hence, for every $k$ and for the selected rule,
\[
 \varepsilon^*_{n,y}(\hat\theta_k;\y\mid\X)
 \ge \varepsilon^*_{n,y}(\tilde\theta;\y\mid\X),\qquad
 \varepsilon^*_{n,y}(\hat\theta_{\mathrm{sel}};\y\mid\X)
 \ge \varepsilon^*_{n,y}(\tilde\theta;\y\mid\X).
\]
The argument never varies $\X$. \qed

\subsection{Proof of Theorem~\ref{thm:support}}\label{app:support}
Let $\bar{\bm x}=n^{-1}\sum_i\bm x_i$ and
$\bar{\tilde y}=n^{-1}\sum_i\tilde y_i$, where these are the original
pre-centering variables. Profiling the intercept gives the
exact decomposition
\[
Q^+_{\lambda_n}(\alpha,\bbeta;\tilde\y)
=Q_{\lambda_n}(\bbeta;\btil)
+\tfrac12\{\alpha-\bar{\tilde y}+\bar{\bm x}^{\top}\bbeta\}^2,
\]
where $Q_{\lambda_n}$ uses the centered response and centered design. Thus a
strict local minimum of the profiled slope criterion lifts to a strict joint
local minimum at
$\alpha=\bar{\tilde y}-\bar{\bm x}^{\top}\bbeta$.
Write $H_\cA=\X_\cA(\X_\cA^\top\X_\cA)^{-1}\X_\cA^\top$ and
$m=\min_{j\in\cA}|\beta_{0j}|>0$, and consider the candidate point
$\bora=(\bora_\cA,\bm0)$ with
$\bora_\cA=(\X_\cA^\top\X_\cA)^{-1}\X_\cA^\top\X\btil$. We verify in turn
that the active block is stationary, that the inactive block satisfies the
subgradient condition, and that the point is a strict local minimum; each
holds on an event of probability tending to one, and we work on the
intersection of the three.

\emph{(i) The active block.} By construction $\bora_\cA$ minimizes the
quadratic $\|\X\btil-\X_\cA\bm b\|^2$ over $\bm b$, so the normal
equations $\X_\cA^\top(\X\btil-\X_\cA\bora_\cA)=\bm0$ hold exactly and the
gradient of the quadratic part of $Q_{\lambda_n}$ vanishes in every
coordinate $j\in\cA$. Stationarity therefore requires only that the
penalty contribute no gradient, that is
$p_{\lambda_n}'(|\hat\beta^{\mathrm{ora}}_j|)=0$,
which for SCAD holds as soon as
$|\hat\beta^{\mathrm{ora}}_j|>a\lambda_n$. To see that
this eventually holds, expand the candidate. Splitting
$\X\btil=\X_\cA\btil_\cA+\X_\cZ\btil_\cZ$ in its definition gives the
purely algebraic identity
\begin{equation}\label{eq:Tn}
\bora_\cA=\btil_\cA+(\X_\cA^\top\X_\cA)^{-1}\X_\cA^\top\X_\cZ\btil_\cZ
=T_n\btil,
\qquad
T_n=\big[\mathbf I,\ (\X_\cA^\top\X_\cA)^{-1}\X_\cA^\top\X_\cZ\big],
\end{equation}
and since $\bbeta_{0\cZ}=\bm0$ we have $T_n\bbeta_0=\bbeta_{0\cA}$, so
$\bora_\cA-\bbeta_{0\cA}=T_n(\btil-\bbeta_0)$. By (A1) $T_n$ converges to
a bounded limit and by (A2) $\btil-\bbeta_0=O_p(n^{-1/2})$, whence
$\bora_\cA=\bbeta_{0\cA}+O_p(n^{-1/2})$. Consequently, with probability
tending to one, $\min_{j\in\cA}|\hat\beta^{\mathrm{ora}}_j|\ge3m/4$ and
$a\lambda_n<m/4$. Hence $|\hat\beta^{\mathrm{ora}}_j|>a\lambda_n$ eventually
for every $j\in\cA$. This is where the signal-strength requirement enters:
the active coefficients must clear the flat region of the penalty.

\emph{(ii) The inactive block.} For $j\in\cZ$ the condition for
$\hat\beta^{\mathrm{ora}}_j=0$ to satisfy the subgradient inclusion is
$|\tfrac1n\X_j^\top(\X\btil-\X_\cA\bora_\cA)|\le p_{\lambda_n}'(0+)
=\lambda_n$. The residual simplifies: since
$\X_\cA\bora_\cA=H_\cA\X\btil$ and $(\mathbf I-H_\cA)\X_\cA=\bm0$,
\[
\X\btil-\X_\cA\bora_\cA=(\mathbf I-H_\cA)\X\btil
=(\mathbf I-H_\cA)\big(\X_\cA\btil_\cA+\X_\cZ\btil_\cZ\big)
=(\mathbf I-H_\cA)\X_\cZ\btil_\cZ .
\]
The initial estimate of the \emph{inactive} coefficients is therefore the
only thing that can violate the condition. Since $\bbeta_{0\cZ}=\bm0$,
(A2) gives $\btil_\cZ=O_p(n^{-1/2})$, and
$\tfrac1n\X_j^\top(\mathbf I-H_\cA)\X_\cZ=O(1)$ by (A1), so the left-hand
side is $O_p(n^{-1/2})$. By (A3), $\sqrt n\lambda_n\to\infty$, that is
$n^{-1/2}=o(\lambda_n)$, and the condition holds with the margin
\begin{equation}\label{eq:margin}
\Big|\tfrac1n\X_j^\top(\mathbf I-H_\cA)\X_\cZ\btil_\cZ\Big|
\;\le\;\tfrac12\lambda_n
\qquad\text{for all }j\in\cZ
\end{equation}
with probability tending to one. The factor $\tfrac12$ is not needed for
stationarity but is used in step~(iii).

\emph{(iii) Strict local minimality.} Work on the event where (i) and
\eqref{eq:margin} hold, and perturb by $\bm v$ with
$\|\bm v\|_\infty<\min(\lambda_n,m/4)$. Split
$\bm v=(\bm v_\cA,\bm v_\cZ)$. The active penalty is constant because
$|\hat\beta^{\mathrm{ora}}_j+v_j|>m/2>a\lambda_n$, while the inactive SCAD penalty is
linear because $|v_j|<\lambda_n$. Put
$\bm r=\X\btil-\X_\cA\bora_\cA$ and
$\bm g_\cZ=-\X_\cZ^\top\bm r/n$. The active normal equations give
$\X_\cA^\top\bm r=\bm0$, and \eqref{eq:margin} gives
$\|\bm g_\cZ\|_\infty\le\lambda_n/2$. A single expansion, which retains
the cross-block quadratic term, yields
\[
\begin{aligned}
&Q_{\lambda_n}(\bora+\bm v)-Q_{\lambda_n}(\bora)\\
&\quad=\bm g_\cZ^\top\bm v_\cZ
+\frac1{2n}\|\X_\cA\bm v_\cA+\X_\cZ\bm v_\cZ\|^2
+\lambda_n\|\bm v_\cZ\|_1\\
&\quad\ge \frac{\lambda_n}{2}\|\bm v_\cZ\|_1
+\frac1{2n}\|\X_\cA\bm v_\cA+\X_\cZ\bm v_\cZ\|^2.
\end{aligned}
\]
If $\bm v_\cZ\ne\bm0$, the first term is strictly positive. If
$\bm v_\cZ=\bm0$ but $\bm v_\cA\ne\bm0$, the second term is strictly
positive by (A1). Thus every nonzero sufficiently small perturbation raises
the objective, proving strict local minimality. \qed

\subsection{Proof of Theorem~\ref{thm:oracle}}\label{app:oracle}

\emph{(i) The distilled block is a linear map of the initial estimate.}
This is \eqref{eq:Tn}, established in the previous proof:
$\bora_\cA=T_n\btil$ and, because $\bbeta_{0\cZ}=\bm0$ makes
$T_n\bbeta_0=\bbeta_{0\cA}$,
\[
\bora_\cA-\bbeta_{0\cA}=T_n(\btil-\bbeta_0),
\qquad
T_n=\big[\mathbf I,\ (\X_\cA^\top\X_\cA/n)^{-1}(\X_\cA^\top\X_\cZ/n)\big].
\]
This is the folding-back interpretation discussed after
Theorem~\ref{thm:support}. Note that
the identity is exact in finite samples and requires none of the
assumptions; only the limit in step (ii) does.

\emph{(ii) The limit.} By (A1), $T_n\to
[\mathbf I,\ M_{\cA\cA}^{-1}M_{\cA\cZ}]
=M_{\cA\cA}^{-1}[M_{\cA\cA},M_{\cA\cZ}]=M_{\cA\cA}^{-1}M_{\cA\cdot}=T$, a
constant matrix. Given $\sqrt n(\btil-\bbeta_0)\xrightarrow{d}
N(0,\tilde\Sigma)$, Slutsky's theorem applied to
$\sqrt n(\bora_\cA-\bbeta_{0\cA})=T_n\sqrt n(\btil-\bbeta_0)$ gives the
limit $N(0,T\tilde\Sigma T^\top)$.

\emph{(iii) The oracle variance.} Suppose
$\tilde\Sigma=c_\rho M^{-1}$.
Then $T\tilde\Sigma T^\top=c_\rho
M_{\cA\cA}^{-1}(M_{\cA\cdot}M^{-1}M_{\cdot\cA})M_{\cA\cA}^{-1}$, and the
middle factor collapses because $M_{\cA\cdot}M^{-1}$ is the $\cA$-block of
rows of $MM^{-1}=\mathbf I$, namely $[\mathbf I_\cA,\bm0]$; multiplying by
$M_{\cdot\cA}$ picks out $M_{\cA\cA}$. Hence
$T\tilde\Sigma T^\top=c_\rho M_{\cA\cA}^{-1}M_{\cA\cA}M_{\cA\cA}^{-1}
=c_\rho M_{\cA\cA}^{-1}$. For initial-estimator classes whose restricted refit obeys
the same proportional-covariance formula, this is also the asymptotic
variance of the true-submodel refit.

\emph{(iv) The comparison with truncation.} Write
$S=M_{\cA\cZ}M_{\cZ\cZ}^{-1}M_{\cZ\cA}$ for the Schur correction. The
block-inverse identity states that
$(M^{-1})_{\cA\cA}=(M_{\cA\cA}-S)^{-1}$. Since $M\succ0$ implies
$M_{\cZ\cZ}\succ0$, the matrix $S$ is positive semidefinite, so
$M_{\cA\cA}-S\preceq M_{\cA\cA}$; inverting reverses the order and gives
$(M^{-1})_{\cA\cA}\succeq M_{\cA\cA}^{-1}$. Equality requires $S=\bm0$,
and as $M_{\cZ\cZ}^{-1}\succ0$ this forces $M_{\cA\cZ}=\bm0$.

\emph{(v) The least-squares case is exact in finite samples.} If $\btil$
is the ordinary least-squares estimate then, with $\y$ denoting the centered
response after profiling the common intercept, $\X\btil=H\y$ with
$H=\X(\X^\top\X)^{-1}\X^\top$. Because the column space of $\X_\cA$ is
contained in that of $\X$, $H\X_\cA=\X_\cA$ and hence
$\X_\cA^\top H=\X_\cA^\top$, so
$\X_\cA^\top\X\btil=\X_\cA^\top\y$ and
$\bora_\cA=(\X_\cA^\top\X_\cA)^{-1}\X_\cA^\top\y$. The distilled active
block is then \emph{identically} the oracle least-squares fit, not merely
its asymptotic equal. \qed

\subsection{Proof of Theorem~\ref{thm:equiv}}\label{app:equiv}
Both estimators are folded-concave penalizations of a quadratic form
centred at the same $\btil$; they differ only in the matrix defining that
form. Write $\mathbf{W}_1=\X^\top\X/n$ for the distillation weight and
$\mathbf{W}_2=(n\hat\Sigma)^{-1}$ for the LSA weight in the same normalization,
and for a symmetric positive definite $\mathbf{W}$ define the projection
\[
T^{\mathbf{W}}=(\mathbf{W}_{\cA\cA})^{-1}\mathbf{W}_{\cA\cdot} .
\]

\emph{Step 1: both oracle local branches are such a projection.} By
Theorem~\ref{thm:support}, with probability tending to one the distilled
oracle branch has support $\cA$, its active SCAD derivatives vanish, and its
active block is
$\bora_\cA=(\X_\cA^\top\X_\cA)^{-1}\X_\cA^\top\X\btil=T^{\mathbf{W}_1}\btil$. By
hypothesis the SCAD--LSA criterion has a local branch with support $\cA$
and active coefficients in the flat penalty region. Its active
stationarity equations are therefore precisely those for minimizing
$(\bbeta-\btil)^\top\mathbf{W}_2(\bbeta-\btil)$ subject to
$\bbeta_\cZ=\bm0$, whose solution is $T^{\mathbf{W}_2}\btil$. The two
branch supports therefore agree with probability tending to one.

\emph{Step 2: the leading term cancels identically.} For \emph{any}
positive definite $\mathbf{W}$,
\[
T^{\mathbf{W}}\bbeta_0
=(\mathbf{W}_{\cA\cA})^{-1}\big(\mathbf{W}_{\cA\cA}\bbeta_{0\cA}
+\mathbf{W}_{\cA\cZ}\bbeta_{0\cZ}\big)
=\bbeta_{0\cA},
\]
because $\bbeta_{0\cZ}=\bm0$. The value of the projection at $\bbeta_0$ is
thus the same for every weight, so
\[
\bLSA_\cA-\bora_\cA
=(T^{\mathbf{W}_2}-T^{\mathbf{W}_1})\btil
=(T^{\mathbf{W}_2}-T^{\mathbf{W}_1})(\btil-\bbeta_0).
\]
This cancellation is what makes the result hold without any rate on
$\hat\Sigma$: the difference of the two weights multiplies not $\btil$
itself but the initial estimator's error.

\emph{Step 3: the two weights have the same limit up to scale.} By (A1),
$\mathbf{W}_1\to M$. By the consistency hypothesis,
$n\hat\Sigma\to c_\rho M^{-1}$ in probability, so
$\mathbf{W}_2\to c_\rho^{-1}M$. The map $\mathbf{W}\mapsto T^{\mathbf{W}}$ is
invariant to positive scaling, since $T^{c\mathbf{W}}=(c\mathbf{W}_{\cA\cA})^{-1}
(c\mathbf{W}_{\cA\cdot})=T^{\mathbf{W}}$, and it is continuous at any positive definite
$\mathbf{W}$. Hence $T^{\mathbf{W}_1}$ and $T^{\mathbf{W}_2}$ both converge in probability to
$T=M_{\cA\cA}^{-1}M_{\cA\cdot}$ and $T^{\mathbf{W}_2}-T^{\mathbf{W}_1}=o_p(1)$.

Combining Steps 2 and 3 with (A2), which gives
$\btil-\bbeta_0=O_p(n^{-1/2})$,
\[
\bLSA_\cA-\bora_\cA=o_p(1)\cdot O_p(n^{-1/2})
=o_p(n^{-1/2}),
\]
as claimed. The comparison is symmetric in the two quadratic weights once
both local branches have the projection form. Support recovery alone is
not sufficient: the flat-active-penalty condition is what removes the
penalty derivative from each active stationarity equation. \qed

\subsection{Proof of Theorem~\ref{thm:if}}\label{app:if}
Under the support- and flat-status-stability condition, define the active
branch by
\[
b_{\cA}(F)=T(F)\,\btil(F),\qquad b_{\cZ}(F)=0,\qquad
T(F)=M(F)_{\cA\cA}^{-1}M(F)_{\cA\cdot},
\]
where, because the intercept is profiled,
$M(F)=\operatorname{Cov}_F(\bm x)$.
Contaminate along
\[
F_\varepsilon=(1-\varepsilon)F_0+\varepsilon\delta_z,\qquad
z=(\bm x_0,y_0),\qquad
\dot{(\,\cdot\,)}=
\left.\frac{\partial}{\partial\varepsilon}\right|_{\varepsilon=0},
\]
and write $T=T(F_0)=T_0$. Since $\btil$ is Fisher consistent at $F_0$ by
hypothesis, $\btil(F_0)=\bbeta_0$, and the product rule gives
\begin{equation}\label{eq:ifsplit}
\mathrm{IF}(z;b_\cA,F_0)
=\underbrace{T\cdot\mathrm{IF}(z;\btil,F_0)}_{\text{transferred}}
+\underbrace{\dot T\,\bbeta_0}_{\text{design term}} ,
\end{equation}
so the theorem amounts to showing that the second term vanishes at the
model. Write $\bm\mu=E_{F_0}(\bm x)$ and
$\bm u_0=\bm x_0-\bm\mu$. The covariance functional has derivative
$\dot M=\bm u_0\bm u_0^\top-M$. Differentiating
the identity $M_{\cA\cA}T=M_{\cA\cdot}$ gives
$\dot M_{\cA\cA}T+M_{\cA\cA}\dot T=\dot M_{\cA\cdot}$, whence
\[
\dot T=M_{\cA\cA}^{-1}\big(\dot M_{\cA\cdot}-\dot M_{\cA\cA}T\big)
=M_{\cA\cA}^{-1}\bm u_{0\cA}\big(\bm u_0^\top-\bm u_{0\cA}^\top T\big),
\]
the terms in $M$ itself cancelling for the same reason. Applying this to
$\bbeta_0$ and using $T\bbeta_0=\bbeta_{0\cA}$,
\[
\dot T\bbeta_0=M_{\cA\cA}^{-1}\bm u_{0\cA}
\big(\bm u_0^\top\bbeta_0-\bm u_{0\cA}^\top\bbeta_{0\cA}\big)
=M_{\cA\cA}^{-1}\bm u_{0\cA}\,\bm u_{0\cZ}^\top\bbeta_{0\cZ}=\bm0 ,
\]
because $\bbeta_{0\cZ}=\bm0$ at the model. The bracket is exactly the
centered initial-to-active-projection residual at the contaminating point,
which is what
makes the design term disappear, and \eqref{eq:ifsplit} reduces to the
stated identity. On a contamination class for which
$\mathrm{IF}(z;\btil,F_0)$ is bounded, the active-branch influence is
bounded by $\|T\|$ times that bound. The cancellation is exact only at the
model. At a general baseline $F$, with $\bm u=\bm x_0-E_F(\bm x)$ the
correspondingly recentred contamination, the analogous design term is
\[
 M(F)_{\cA\cA}^{-1}\bm u_{\cA}
 \{\bm u_{\cZ}^{\top}-\bm u_{\cA}^{\top}
 M(F)_{\cA\cA}^{-1}M(F)_{\cA\cZ}\}\btil(F)_{\cZ}.
\]
The residualized bracket need not vanish, and the product can grow
quadratically in $\|\bm x_0\|$, which is
the analytic form of the vulnerability discussed in
\ORsec{sec:mech}. \qed

\subsection{Proof of Proposition~\ref{prop:gic}}\label{app:gic}
Write $q=|\cA|$, $\hat s_n^2=\max(\hat\sigma^2,10^{-12})$, and
$\mathrm{GIC}(\lambda)=\RSS_d(\lambda)/\hat s_n^2+a_n\,\mathrm{df}
(\lambda)$ with $a_n=\kappa\log n$. The common intercept adds one to every
candidate's degrees of freedom and cancels in pairwise comparisons, so in
this proof $\mathrm{df}$ denotes the number of selected slopes. Because
$\hat\sigma$ is bounded above and below in probability by hypothesis,
$\hat s_n^2$ is as well; the numerical floor is therefore immaterial to the
order comparisons. Partition the retained states
on the computed path by their slope support and compare them with the
qualifying state at $\lambda^\circ$. The argument uses the stated
$O_p(1)$ residual property of that computed state.

\emph{(i) The qualifying state.} By hypothesis,
$\mathrm{supp}\,\bcomp(\lambda^\circ)=\cA$ and
$\RSS_d(\lambda^\circ)=O_p(1)$. Its GIC is therefore
$O_p(1)+a_nq$ after removal of the common intercept term.

\emph{(ii) Underfitting supports.} Suppose a retained state has slope
support $S$ omitting some $j\in\cA$, and write
$\ell_n=\lambda_{\min}(\X^\top\X/n)$. Because $\X$ is column-centered,
the intercept direction is orthogonal to the slope fit, so the state's
distillation residual is bounded below by the best slope approximation
from its own support: for every $\bm b$ with
$\mathrm{supp}(\bm b)\subseteq S$ the vector $\btil-\bm b$ agrees with
$\btil$ on $S^c\ni j$, whence
\[
\min_{\mathrm{supp}(\bm b)\subseteq S}
\;n^{-1}\|\X(\btil-\bm b)\|^2
\;\ge\;\ell_n
\min_{\mathrm{supp}(\bm b)\subseteq S}\|\btil-\bm b\|^2
\;=\;\ell_n\|\btil_{S^c}\|^2
\;\ge\;\ell_n\tilde\beta_j^2,
\]
and therefore $\RSS_d\ge n\,\ell_n\tilde\beta_j^2$ at every such state.
By (A1), $\ell_n\to\lambda_{\min}(M)>0$, and
$\tilde\beta_j\to\beta_{0j}\neq0$ in probability, so the lower bound is
at least $n\lambda_{\min}(M)\beta_{0j}^2/4$ with probability tending to
one. The criterion therefore exceeds the qualifying state's value by an
amount of order $n$, against a penalty saving of at most $a_np=o(n)$, so
underfitting supports are rejected.

\emph{(iii) Overfitting supports.} Suppose the support strictly contains
$\cA$. For any such retained state, nonnegativity of its residual sum of
squares gives
\[
\mathrm{GIC}(\lambda)-\mathrm{GIC}(\lambda^\circ)
\ge a_n-\RSS_d(\lambda^\circ)/\hat s_n^2
=a_n-O_p(1)\longrightarrow\infty.
\]
Thus no least-squares nesting property is assumed for arbitrary local path
states, and overfitting supports are rejected as well.

Every support other than $\cA$ falls into class (ii) or class (iii), so
the minimizer over the computed candidate path has support $\cA$ with
probability tending to one. The qualifying-path condition supplies the
computed oracle-support benchmark, $a_n\to\infty$ excludes overfitting,
and $a_n=o(n)$ keeps the penalty from overwhelming the signal in (ii).
Any fixed $\kappa>0$ has both rate properties. \qed

\subsection{Statements for the screened initial fit}\label{supp:screen-thy}
\providecommand{\bscr}{\tilde{\bm{\beta}}^{\mathrm{scr}}}
\providecommand{\Sscr}{\widehat{S}}

The screening rule is part of the estimator: its output is a function of
the same response vector
that the robust fit uses, so a response replacement can change the
selected support itself, and the analysis below never conditions on the
observed selection. The cap keeps $K+1<n$, so every admissible submodel design has full column
rank; a high-breakdown submodel rule imposes its own, stricter sample-size
requirement, which the default $K=\lfloor n/4\rfloor$ respects. The rank condition in $\mathcal S_K$ excludes
degenerate submodels, and the response-replacement bound of
Lemma~\ref{lem:scr-bp} uses both properties.

Write $\theta_{0S}=(\alpha_0,\bbeta_{0S})$ for the target restricted to a
support $S$, and
\[
 \mathcal S^*=\bigl\{S\subseteq\{1,\dots,p\}:\ |S|\le K,\
 \cA\subseteq S\bigr\},
\]
a deterministic collection with at most $2^{p}$ elements. Under (A1),
for every $S\in\mathcal S^*$ the matrix $[\bm1,\X_S]$ has full column
rank for all sufficiently large $n$, so the submodel robust fit
$\tilde\theta_S=(\tilde\alpha_S,\btil_S)$ is eventually defined.

\begin{assumption}\label{ass:2p}
(A2$'$) The target conditions of Assumption~\ref{ass:2} remain in force:
$\bbeta_0$ and $\cA$ do not vary with $n$, $0<|\cA|<p$, and
$\min_{j\in\cA}|\beta_{0j}|>0$. The rate clause of
Assumption~\ref{ass:2} is replaced by the following conditions on the
screening rule and the robust fit.
\begin{enumerate}[label=(\roman*),leftmargin=2.2em,itemsep=1pt]
\item \emph{Sure screening.} $P(\cA\subseteq\Sscr)\to1$.
\item \emph{Cap.} The cap satisfies $K\ge|\cA|$.
\item \emph{Submodel rate.} For every $S\in\mathcal S^*$, the robust fit
      on $[\bm1,\X_S]$ satisfies
      $\sqrt n(\tilde\theta_S-\theta_{0S})=O_p(1)$.
\end{enumerate}
\end{assumption}

Part (i) uses the sure-screening terminology of \citet{fanlv08}, but it is
an assumption on the implemented rule at the data-generating law; it does
not follow from the cap or from the use of a robust loss alone.
Selection-consistency results for nonconvex penalized quantile regression
\citep{wangwuli12} provide a relevant sufficient route under their design,
signal and tuning conditions.

\begin{remark}[The repair preserves sure screening]\label{rem:repair}
If the raw screen is selection consistent at the data-generating law,
$P(\Sscr^{\mathrm{raw}}=\cA)\to1$, then on that event
$|\Sscr^{\mathrm{raw}}|=|\cA|\le K$, so the truncation step is inactive,
and under (A1) the matrix $[\bm1,\X_\cA]$ has full column rank for all
sufficiently large $n$, so the rank repair is inactive as well; the repaired
rule then coincides with the raw screen with probability tending to one and
inherits (A2$'$)(i).
\end{remark}

We do not assert that these literature conditions are verified
for every finite-sample design or data-driven path used below. The overall shape ---
reduce the candidate set, then fit a folded-concave penalized model on
the reduced set --- is the one used by \citet{riostong26}, who screen to
$\lfloor 1.5n/\log n\rfloor$ predictors before a SCAD fit. The cap
$K$ plays the same role here. The implementation uses a median-quantile
loss to reduce sensitivity to response contamination; this choice alone
does not prove (i). Under contaminated response laws, (i) and (iii) are
assumed at that law, in the same spirit
as the residual-law convention of Lemma~\ref{lem:scr-scale}: the screen's
population target and the submodel estimand are then those of the
contaminated distribution. Part
(iii) is Assumption~\ref{ass:2} applied to correctly specified
submodels: because $\cA\subseteq S$, each submodel contains the truth,
and the standard conditions that give the full-model joint rate for M-,
S- and MM-estimators \citep{yohai87} give it on $[\bm1,\X_S]$.

\begin{lemma}[The screened initial fit inherits the rate]\label{lem:scr-rate}
Let $p$ be fixed. Under (A1) and Assumption~\ref{ass:2p},
$\sqrt n\bigl((\tilde\alpha_{\Sscr},\bscr)-(\alpha_0,\bbeta_0)\bigr)
=O_p(1)$; in particular the screened robust initial fit satisfies the
rate requirement of Assumption~\ref{ass:2}.
\end{lemma}

Thus the downstream rate is inherited once (A2$'$) is verified. The
breakdown bound below dominates the fit over every support the rule can
return.

\begin{lemma}[Conditional response-replacement bound]\label{lem:scr-bp}
Fix $\X$. Write $\tilde\theta_S$ for the robust fit rule on
$[\bm1,\X_S]$ and $\tilde\theta_{\mathrm{scr}}$ for the screened rule
$(\tilde\alpha_{\Sscr},\bscr)$. Then, with
$\varepsilon^*_{n,y}$ as in Section~\ref{sec:finite} and $\mathcal S_K$
the admissible collection defined in Section~\ref{sec:screen-def},
\[
  \varepsilon^*_{n,y}(\tilde\theta_{\mathrm{scr}};\y\mid\X)
  \;\ge\;\min_{S\in\mathcal S_K}
  \varepsilon^*_{n,y}(\tilde\theta_S;\y\mid\X_S).
\]
\end{lemma}

The bound is a minimum over submodels, so it can be smaller than the
full-model breakdown point: the screened rule inherits the floor of
whichever admissible submodel is weakest. The bound applies to a
specified submodel robust-fit rule; the implemented failure-aware
composite rule requires the separate assessment of
\ORsec{supp:impl-cascade}.

The remaining ingredient is the residual scale: Proposition~\ref{prop:gic}
consumes a two-sided bound on $\hat\sigma$, and the screened fit must be
shown to supply it.

\begin{lemma}[Residual scale]\label{lem:scr-scale}
Assume (A1) and Assumption~\ref{ass:2p}, that the design additionally
satisfies $\max_{i\le n}\|\bm x_i\|_2=o(\sqrt n)$, and that the residuals at the target,
$e_i=y_i-\alpha_0-\bm x_i^\top\bbeta_0$, are independent draws from a
continuous distribution $F_e$ whose median and median absolute deviation
are unique, with $\mathrm{MAD}(F_e)\in(0,\infty)$. Then the residual
scale $\hat\sigma$ of \eqref{eq:sigma} computed from the screened
initial fit satisfies $0<c\le\hat\sigma\le C<\infty$ with probability
tending to one, so the scale hypothesis of Proposition~\ref{prop:gic}
holds.
\end{lemma}

The additional design condition holds, for example, when the row norms are
uniformly bounded; because $p$ is fixed here, uniformly bounded entries are
an equivalent sufficient condition. It is not implied by (A1) alone.

Under response contamination, $F_e$ is the contaminated residual law,
whose continuity and nondegenerate median absolute deviation hold for
an independent $\varepsilon$-mixture of a continuous error law and a
continuous contamination law. The fixed-count contamination used in the
simulations is a conditional variant of that model: conditionally on the
contaminated index set the residuals are independent draws from two
continuous laws, and the proof of the lemma applies to that array.

\begin{remark}[Scope]\label{rem:scr-scope}
The analysis is a fixed-$p$ statement; growing-dimension theory would need
a rate uniform over supports and is left open. The proposal concerns the
unweighted estimator, and cellwise contamination of the design calls for a
different treatment \citep{stefelova21}. The fixed-$p$ theory is also
silent on finite-sample behaviour: both initial fits satisfy the rate
requirement asymptotically, yet the distilled estimator built on the
full-coordinate fit collapses at moderate $p/n$
(Section~\ref{sec:screen-sim}).
\end{remark}

\begin{remark}[The scale hypothesis in practice]\label{rem:scr-sigma}
Proposition~\ref{prop:gic} assumes $0<c\le\hat\sigma\le C<\infty$ with
probability tending to one. The finite-sample dimension study does not
establish a violation of this fixed-$p$ asymptotic condition. It does show
substantial residual-scale inflation for the full-coordinate initial fit,
which compresses the normalized fidelity gain and can favor the null model.
An initial-fit residual scale far above a problem-specific reference is
therefore a computable warning to investigate the initial fit and the
resulting path before distillation. We propose this diagnostic alongside
the algorithmic audits of Section~\ref{sec:algo}; it is not a formal test
of the proposition's assumptions, and the reference scale is
application-specific.
\end{remark}
{}

\subsection{Proof of Lemma~\ref{lem:scr-rate}}\label{app:scr-rate}
Let $E_n=\{\cA\subseteq\Sscr\}$, so $P(E_n)\to1$ by (A2$'$)(i). On
$E_n$ we have $\Sscr\in\mathcal S^*$ and $\bbeta_{0\Sscr^{c}}=\bm0$, and
$\bscr$ agrees with $\btil_{\Sscr}$ on $\Sscr$ and vanishes off it, so
\[
  \bigl\|(\tilde\alpha_{\Sscr},\bscr)-(\alpha_0,\bbeta_0)\bigr\|_2
  =\bigl\|\tilde\theta_{\Sscr}-\theta_{0\Sscr}\bigr\|_2
  \;\le\;\max_{S\in\mathcal S^*}
  \bigl\|\tilde\theta_S-\theta_{0S}\bigr\|_2 ,
\]
where each term is defined for all large $n$ by the rank remark
preceding Assumption~\ref{ass:2p}. Because $p$ is fixed, $\mathcal S^*$ is a subcollection of the fixed
finite family $2^{\{1,\dots,p\}}$, of cardinality at most
$\sum_{k\le K}\binom pk\le 2^p$ for every $n$; a maximum of
finitely many $O_p(n^{-1/2})$ terms is $O_p(n^{-1/2})$, so by
(A2$'$)(iii) the right-hand side is $O_p(n^{-1/2})$. Since
$P(E_n)\to1$, the conclusion follows. \qed

\subsection{Proof of Lemma~\ref{lem:scr-bp}}\label{app:scr-bp}
Fix $m$ and let $\y'\in\mathcal Y_m(\y)$. By construction
$\Sscr(\X,\y')\in\mathcal S_K$, and embedding by zeros preserves the
Euclidean norm, so
$\|\tilde\theta_{\mathrm{scr}}(\X,\y')\|_2
=\|\tilde\theta_{\Sscr(\X,\y')}(\X_{\Sscr(\X,\y')},\y')\|_2$. Hence
\[
  \sup_{\y'\in\mathcal Y_m(\y)}\|\tilde\theta_{\mathrm{scr}}(\X,\y')\|_2
  \;\le\;\max_{S\in\mathcal S_K}\ \sup_{\y'\in\mathcal Y_m(\y)}
  \|\tilde\theta_S(\X_S,\y')\|_2 .
\]
The collection $\mathcal S_K$ is finite, so the right-hand side is
finite whenever each of its terms is, which holds for every $m$ with
$m/n<\min_{S\in\mathcal S_K}
\varepsilon^*_{n,y}(\tilde\theta_S;\y\mid\X_S)$. Hence no such $m$
belongs to the defining set of
$\varepsilon^*_{n,y}(\tilde\theta_{\mathrm{scr}};\y\mid\X)$, and the
stated bound follows from the definition of that breakdown point as a
minimum. \qed

\subsection{Proof of Lemma~\ref{lem:scr-scale}}\label{app:scr-scale}
The fitted residuals are
$r_i=e_i-(\tilde\alpha_{\Sscr}-\alpha_0)
-\bm x_i^\top(\bscr-\bbeta_0)$, so by Lemma~\ref{lem:scr-rate} and the
design condition
\[
 \Delta_n:=\max_{i\le n}|r_i-e_i|
 \le|\tilde\alpha_{\Sscr}-\alpha_0|
 +\max_{i\le n}\|\bm x_i\|_2\,\|\bscr-\bbeta_0\|_2=o_p(1),
\]
where the intercept term is covered because Lemma~\ref{lem:scr-rate}
bounds the joint vector. Writing $F^r_n$ and $F^e_n$ for the two
empirical distribution functions,
$F^e_n(t-\Delta_n)\le F^r_n(t)\le F^e_n(t+\Delta_n)$ for every $t$, so
\[
 \sup_t|F^r_n(t)-F^e_n(t)|
 \le\sup_t\{F^e_n(t+\Delta_n)-F^e_n(t-\Delta_n)\}.
\]
By Glivenko--Cantelli and the uniform continuity of the continuous
distribution function $F_e$, the right-hand side is $o_p(1)$; hence
$\sup_t|F^r_n(t)-F_e(t)|=o_p(1)$. The median and the median absolute
deviation are continuous functionals with respect to the supremum norm
at any distribution at which they are uniquely defined
\citep[Ch.~3]{huber81}, so
$\hat\sigma\to_p1.4826\,\mathrm{med}_{F_e}|e-\mathrm{med}(F_e)|
\in(0,\infty)$, the constant being the normal-consistency factor of
\eqref{eq:sigma}. This
gives the stated bounds, and the nonrobust-RMS and unit fallbacks in
\eqref{eq:sigma} are not taken with probability tending to one. \qed

\subsection{Proof of Corollary~\ref{cor:scr-modular}}\label{app:scr-modular}
Proposition~\ref{prop:norm} and Lemma~\ref{lem:warm} are deterministic
statements about the objective $Q^+_\lambda$ and the computed grid for a
given $\tilde\theta$; they do not refer to how $\tilde\theta$ was
obtained. Corollary~\ref{cor:bp} is stated for an arbitrary initial
estimator rule and transfers whatever conditional bound that rule
enjoys, which for the screened rule is Lemma~\ref{lem:scr-bp}; this
gives (a). Theorem~\ref{thm:support} uses the initial estimator only
through the rate in Assumption~\ref{ass:2} together with the target
conditions, all of which Assumption~\ref{ass:2p} and
Lemma~\ref{lem:scr-rate} supply, and Proposition~\ref{prop:gic} uses in
addition only the two-sided bound on $\hat\sigma$, which
Lemma~\ref{lem:scr-scale} supplies; this gives (b). For (c),
Theorem~\ref{thm:oracle} hypothesizes
$\sqrt n(\btil-\bbeta_0)\to_dN(0,\tilde\Sigma)$ and
Theorem~\ref{thm:equiv} hypothesizes a positive definite $\hat\Sigma$
with $n\hat\Sigma\to_pc_\rho M^{-1}$; both proofs consume these
hypotheses directly and are otherwise estimator-agnostic, so the
statements transfer conditionally on them. On the event
$\{\Sscr=\cA\}$ the embedded fit satisfies $\X\bscr=\X_\cA\btil_\cA$,
so the candidate of Theorem~\ref{thm:support}'s proof reduces to
$\bora_\cA=\btil_\cA$, which is the degeneracy noted in (c); off the
selected support the screened fit is exactly zero, so no limit
covariance of $\sqrt n(\bscr-\bbeta_0)$ can equal the positive definite
$c_\rho M^{-1}$. \qed

\section{A nonparametric check on the selection rule}\label{app:spline}

Section~\ref{sec:algo} of the main text argues that cross-validation on
the initial fitted surface answers a fidelity question, with an error curve
that can be nearly degenerate because the unpenalized second-stage fit can
reproduce the initial fit exactly. The argument concerns the construction,
not the SCAD penalty, so it should survive a change of function class.
In this section we test that argument in a nonparametric setting.

\subsection{Design}

The regression functions are
$f_1(x)=\sin(2\pi x)$,
$f_2(x)=\exp\{-8(x-0.3)^2\}+\tfrac12\sin(6\pi x)$, which mixes smooth and
oscillatory behaviour, and
$f_3(x)=\{1+\exp(-20(x-0.5))\}^{-1}$, which has a sharp transition.
Design points are equally spaced on $[0,1]$ with
$n\in\{100,200,400\}$, and errors are normal or $t_3$ rescaled to standard
deviation $0.2$. A fraction $\{0,0.05,0.10,0.20\}$ of the responses is
shifted upward by two or four population standard deviations of the clean
response, and the shift size is fixed per configuration rather than
recomputed per replication. Crossing these axes and dropping the redundant
shift size at zero contamination gives 126 configurations, each run with 200
replications on deterministically derived seeds.

We use a quantile smoothing spline
(\texttt{fields::qsreg} at $\alpha=0.5$) as the robust initial fit, selected
by that routine's own pseudo-cross-validation rule over a 100-point
smoothing-parameter grid. The
distilled fit is a least-squares smoothing spline (\texttt{fields::sreg})
fitted to the initial fitted surface on the same grid. One grid fit supplies
every candidate, so the two selection rules are compared on identical
states:
\[
  \mathrm{GCV}(\lambda)=\frac{\RSS(\lambda)/n}{\{1-\mathrm{tr}A_\lambda/n\}^2},
  \qquad
  \mathrm{GIC}(\lambda)=\frac{\RSS(\lambda)}{\hat\sigma^2}
  +\kappa\log(n)\,\mathrm{tr}A_\lambda ,
\]
with $\kappa=2$ and $\hat\sigma$ the median absolute deviation of the
residuals of the initial fit from the \emph{observed} responses, matching
the convention of \eqref{eq:sigma}. Degrees of freedom are the trace of
the smoother matrix, not a count of coordinates, so
Proposition~\ref{prop:gic} does not apply. As a reference we
also fit a least-squares smoothing spline directly to the observed
responses, selected by GCV. We measure accuracy by integrated squared error against the
true function at the design points.

\subsection{Results}

Across all 25{,}200 replications of the distilled fit, generalized
cross-validation selected the most flexible point of the grid \emph{every}
time, so the distilled fit reproduced its initial fit to within
$5.4\times10^{-5}$ relative integrated squared error. The compression step
did no compressing. The criterion of Algorithm~\ref{alg:distill} selected
between about six and thirteen effective degrees of freedom instead, and lowered
the integrated squared error in most replications.
Table~\ref{tab:spline-selection} reports the comparison by regression
function. The gain is largest for $f_1$ and $f_3$ and smallest for $f_2$,
which needs more flexibility than the other two and gets it: 12.6
effective degrees of freedom against 7.4 and 6.2.

\begin{table}[htbp]
\centering
\scriptsize
\caption{Selection rules for the distilled smoothing spline, by regression
function. Entries are the share of replications in which the GIC criterion
of Algorithm~\ref{alg:distill} attains lower integrated squared error than
generalized cross-validation, and the mean selected effective degrees of
freedom under GIC. Generalized cross-validation selected the most flexible
point of the grid in every replication, with effective degrees of freedom
$99.9$, $197.0$ and $335.1$ at $n=100$, $200$ and $400$. Overall, the
median ISE ratio of GIC to GCV is $0.63$, with a paired mean difference of
$-0.00452$ (95\% Monte Carlo interval $[-0.00461,-0.00442]$).}
\label{tab:spline-selection}
\setlength{\tabcolsep}{4pt}
\begin{tabular}{@{}lcc@{}}
\toprule
Function & GIC lowers ISE (\% of repl.) & Mean effective df under GIC\\
\midrule
$f_1$ (sine)             & $98.2$ & $7.4$\\
$f_2$ (mixed smoothness) & $63.7$ & $12.6$\\
$f_3$ (sharp transition) & $91.5$ & $6.2$\\
\midrule
Overall                  & $84.4$ & ---\\
\bottomrule
\end{tabular}
\end{table}

\begin{figure}[!htbp]
\centering
\includegraphics[width=\linewidth]{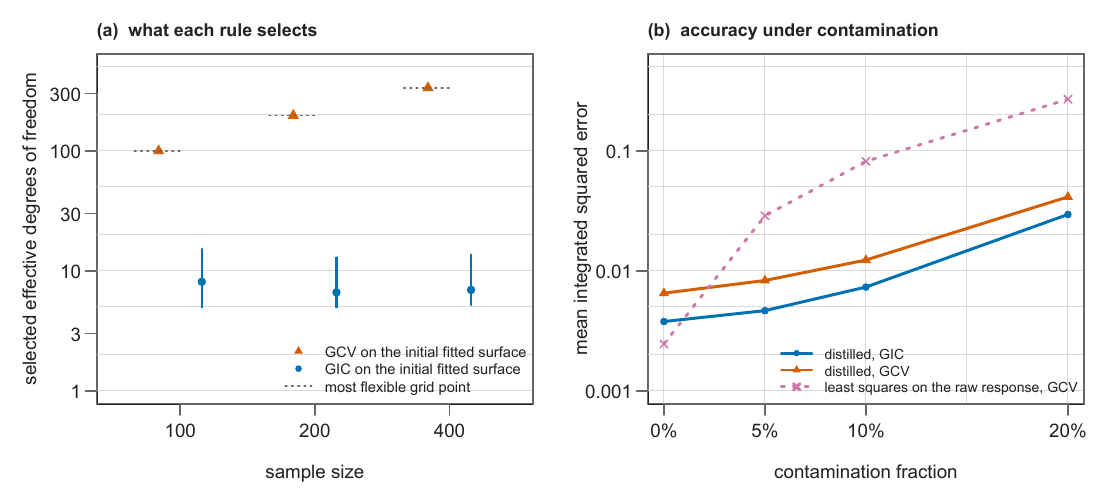}
\caption{Selection rules for a nonparametric distilled fit: a quantile
smoothing spline as the robust initial fit and a least-squares smoothing
spline fitted to its surface. Panel (a) shows the selected effective
degrees of freedom by sample size, with the median and the $5$th--$95$th
percentile range over replications; the dotted marks give the most flexible
point of the smoothing-parameter grid, which generalized cross-validation
selected in every replication. Panel (b) shows mean integrated squared
error against the contamination fraction for normal errors and shifts of
four population standard deviations.}
\label{fig:spline-selector}
\end{figure}

Under contamination the distilled fit inherits the robustness of its initial
fit. With normal errors and shifts of four population standard deviations,
the raw-response least-squares spline is the more accurate of the two in
clean data. The ordering reverses by $5\%$ contamination, and the gap then
widens by an order of magnitude. Figure~\ref{fig:spline-selector}(b) shows
the crossing, and panel (a) shows the selected effective degrees of freedom
behind Table~\ref{tab:spline-selection}.

The initial fit is selected by its own routine's rule rather than by a
criterion of our choosing, so the comparison is between selection rules for
the distilled fit with the initial fit held fixed. The raw-response
reference is a least-squares spline. A robust nonparametric fit selected by
its own criterion would be a different comparison, and the present design
does not speak to it.

\section{Computational provenance}\label{app:prov}

Every number in the dimension study, the mechanism diagnostic, the runtime
table and the nonparametric experiment is produced by a committed script from
a recorded configuration. This section states our conventions. The scripts,
run manifests and checksums are retained in the code and numerical
archives described in the data and code availability statement and can
be provided for editorial and peer-review assessment upon request.

\subsection{Random number generation and reproducibility}

Each experiment is a grid of scenarios, and each scenario is replicated. The
unit of work is one (scenario, replication) cell, and we derive its seed
deterministically from the scenario index and the replication index inside
the cell itself, as
\texttt{set.seed(seed\_base + scenario\_id\,$\times10^{5}$ + rep)}. Per-cell
seeding makes the output independent of worker scheduling. The grid returns
identical results on one core or on sixteen, and partial reruns reproduce
the cells they recompute. We also wrap every fit so that a failure records
a status string and the grid continues. The reported failure rates thus
count every cell of the grid, failed fits included.

\subsection{Parallel execution and what runtimes mean}

For the dimension, density, endpoint-sensitivity and nonparametric
production grids, we used PSOCK workers with BLAS and OpenMP pinned
to one thread per worker, since otherwise each worker starts its own
multithreaded BLAS and the workers contend for the same cores. Tasks were
dispatched with dynamic load balancing at chunk size one, because per-cell
cost varies by three orders of magnitude between the smallest and the
largest configurations. Scenario batches are checkpointed, so an interrupted
grid resumes without recomputation. Because workers share cores and tasks
are balanced dynamically, wall-clock times recorded inside a parallel run
are not comparable across methods; \ORtab{tab:dim-timing} gives the
sequential timings.

\subsection{Experiment inventory}

\begin{itemize}\itemsep2pt
\item \emph{Dimension grid.} 288 scenarios $\times$ 100 replications,
      $403{,}200$ scored fits. Statuses: $388{,}758$ ok, $14{,}400$
      not applicable because the full-coordinate initial fit is undefined at
      $p+1\ge n$, and 42 iteration-limit returns from the robust fitting
      routine. No fatal errors.
\item \emph{Mechanism diagnostic.} Five configurations $\times$ 30
      replications, recording the initial fit's slope norm and residual
      scale, the fidelity reduction available along the computed path, the
      penalty cost of obtaining it, and the selected state.
\item \emph{Sequential timing.} Six configurations $\times$ 20 replications,
      one method at a time, single thread.
\item \emph{Nested lower-endpoint audit.} Four deliberately difficult dimension
      configurations $\times$ 100 common replications, crossing two screens,
      two GIC rates, and endpoint ratios $0.05$, $0.005$, and $0.001$, for
      $800$ screen-specific path computations and $4{,}800$ scored
      method--selector--endpoint rows. The retained 100-point baseline reproduced all
      $1{,}600$ corresponding production selections and scores exactly. All
      path computations completed without a solver or path failure, and the
      $0.005$ and $0.001$ ranges selected the same path index throughout.
\item \emph{Nonparametric experiment.} 126 configurations $\times$ 200
      replications, $126{,}000$ scored fits, no failures.
\end{itemize}

\subsection{Additional simulation provenance}

\emph{Path-audit design.}
A targeted paired path audit used 30 additional clean-Gaussian and 30
additional $10\%$-shifted Gaussian $n=200$, $p=12$ instances, crossing 100
versus 250 grid points, endpoint ratios $10^{-3}$ versus $10^{-6}$, and
safeguarded warm starts versus independent intercept-only starts. Every
controlled comparison selected the same three-variable support with a
largest warm--cold coefficient difference of $1.46\times10^{-5}$. This
bounded low-dimensional diagnostic does not establish initialization
invariance for the 81-variable UCI design.

\emph{Paired base-study intervals.}
Using the replication
pairing, and in the $\times10^{2}$ units of Table~\ref{tab:main}, D-MM minus
SCAD-raw MSE is $-0.154$ with 95\% Monte Carlo interval
$[-0.373,0.065]$ at $n=100$ and $-0.045$
$[-0.129,0.039]$ at $n=200$.
For the displayed $n=100$ clean-$t_3$ row, the paired D-MM minus SCAD-raw difference is
$-9.479$ $[-12.471,-6.488]$, and the D-MM-minus-oracle-LS difference is
$-3.971$ $[-5.404,-2.538]$.

\emph{Adaptive-PENSE configuration.}
We fit adaptive PENSE with
$\alpha=1$, ridge preliminary fit ($\alpha_{\rm pre}=0$), adaptive exponent
one, 100 preliminary and 100 final path points, five-fold robust
information-sharing (RIS) cross-validation, breakdown
setting $0.25$, an
explicit unpenalized intercept, and no additional package standardization.

\emph{Leverage cell.}
The bad-leverage cell reuses both modern-comparator production runners
with one scenario and replication-indexed seeds from a common base, so
the two runs are paired by replication exactly as in the vertical
experiment. All 5{,}000 non-PENSE and 2{,}000 adaptive-PENSE fits
returned finite coefficient vectors. Three RLARS-MM final refits did not
converge and remain included, and the ten SCAD-raw dense-endpoint
selections appear in Table~\ref{tab:sim-audit}.

\emph{Threading and workstation.}
The 1{,}000-replication modern-comparator and bad-leverage experiments used
the same recorded settings: the non-PENSE runner set the numerical thread
environment variables and the \texttt{robustHD} backend budget to 14,
whereas the separate adaptive-PENSE runner fixed its thread settings and
\texttt{ncores} at one. These are configured budgets, not measurements
of actual utilization. The recorded times include the complete fitting
and tuning workflow; the WRMSPE and $\tau$ selections share one
adaptive-PENSE backend fit. These timings are not equal-thread benchmarks.
By contrast, the separate dimension-study timing experiment
(\ORtab{tab:dim-timing}) ran methods sequentially in one R process,
with BLAS and OpenMP configured for one thread. The archived hardware
records and run manifests document these experiment-specific settings.

\emph{Fold-2 grid-density check (observed response).}
At the shallow $10^{-3}$ ratio, both 100- and 250-point
grids selected their last point and the same eight slopes, with training distillation-fidelity RMSE 0.6579 and test-response RMSE
0.9094, both in standardized response units. Holding
250 points but extending the ratio to $10^{-6}$ selected interior index
230/250 and 70 slopes, with corresponding RMSEs of 0.0349 and 0.6916.

\emph{Observed-response audit counts.}
All 30 primary D-MM and SCAD-raw
selected states are accepted, interior, and exactly linked to their retained
path rows. The largest selected-state fixed-point gap is
$2.18\times10^{-5}$, and objective recomputation agrees with the archive.
The merged bundle passes all 108 validation gates and contains 318,945
method--scenario out-of-fold predictions. Unselected lower-$\lambda$ states
that reached the iteration limit remain archived and flagged.

\subsection{Additional superconductivity provenance}

\emph{Data quality.}
The archived source files contain no
missing or non-finite values and no zero-variance predictor.

\emph{Fold construction.}
No component crosses a fold, and fold sizes range from
4,252 to 4,253.

\emph{Path preflight and the diagnostic fold.}
Shallow $10^{-3}$ and $10^{-4}$ paths inspected on outer fold 1 during
computational preflight reached their lower endpoints. The expanded range was
then frozen before fitting the remaining folds. For the observed response,
excluding fold 1 changes D-MM's
pooled out-of-fold RMSE by at most 0.096 K and its mean selected size by at
most 1.45 slopes across the three scenarios. For UCI-X, the analogous
folds-2--5 sensitivity changes D-MM's mean RMSE by at most 0.0058 and its
mean truth-Jaccard index by at most 0.0032.

\emph{Expanded-path Lasso rows.}
In the observed-response analysis, the reported Lasso rows come from
a separately audited expanded-path archive, replace the shallow baseline
Lasso rows, and retain lower-endpoint flags.

\emph{Initial-fit provenance.}
All 30 displayed MM initial fits in the observed-response and UCI-X
analyses completed on the primary branch with status \texttt{ok}, and every
operational residual scale used the MAD branch in \eqref{eq:sigma}. No
fallback in the cascade was triggered (the cascade's breakdown behaviour is
discussed in \ORsec{supp:impl-cascade}).

\emph{Sublevel-condition check.}
We also checked the initial-fit sufficient sublevel condition in
Proposition~\ref{prop:norm}(a). It held in 0 of 15, 0 of 15, and 15 of 15
selected D-MM states at $\kappa=2$, 8, and 64, respectively. The displayed
deterministic fidelity certificate is therefore invoked only for the
$\kappa=64$ states. Because the condition is sufficient rather than
necessary, fidelity at the smaller multipliers is assessed directly through
the observed initial--distilled discrepancies in
Figure~\ref{fig:uci-frontier}.

\subsection{Audit of the computed paths}

Table~\ref{tab:sim-audit} rests on the following run-level detail. In the
base study all 20{,}000 selected SCAD states were accepted, and the 47
archived artifacts, including nine source snapshots, matched an independent
recomputation of the replication summaries. In the modern-comparator study
the 9{,}000 exact-solver selections fell at path indices 23--69. All
3{,}000 selected primary adaptive-PENSE solutions carried package status
code zero, with eight replications containing an isolated nonzero code on an
unselected path solution. The paired designs are regenerated
deterministically from replication-indexed seeds, and the replication
identifiers of the two production runs matched over all 1000 paired
replications per scenario.
In the multiplier study, 3{,}000 of the archived fits underlie the two
displayed panels, no fit selected the lower path endpoint, and a
manifest-verified companion audit reconstructs the strong-design panel's
exact recovery as TPR${}=1$ and FPR${}=0$.

The safeguarded SCAD path records, at every grid point, the fixed-point gap,
the starting, candidate and retained objective values, the acceptance
indicator and any failure reason. Across all $403{,}200$ fits of the
dimension grid, including every $p=240$ configuration, no candidate was
rejected by the safeguard. The screening cap
$K=\lfloor n/4\rfloor$ was binding in 14 of the 1{,}152 screened cells, with
a maximum binding rate of $0.04$ within a cell, so the cap is not shaping
the reported selections.

\section{Further implementation detail}\label{supp:impl}

\subsection{Scalar-update candidate set}\label{supp:impl-candidates}

The coordinate update of Algorithm~\ref{alg:distill} solves each
one-dimensional SCAD problem exactly. In the notation of
\eqref{eq:scalar-scad}, for $\ell_j>0$, write $u_j=|\zeta_j|$ and
$s_j=\operatorname{sign}(\zeta_j)$. The implementation evaluates
\eqref{eq:scalar-scad} at the current coefficient, at the three SCAD
boundaries $0$, $s_j\ell_j$, and $s_ja\ell_j$, and at every feasible
stationary point
\[
 s_j\frac{u_j-\ell_j}{v_j}\in s_j(0,\ell_j),\qquad
 s_j\frac{u_j-a\ell_j/(a-1)}{v_j-1/(a-1)}
       \in s_j(\ell_j,a\ell_j),
 \qquad
 \frac{\zeta_j}{v_j}\in s_j(a\ell_j,\infty).
\]
We include the middle candidate only when $v_j>1/(a-1)$. Otherwise that
piece is concave and its minimum is attained at a boundary already in the
candidate set. The candidate with the smallest scalar objective is used,
with the current value retained under a floating-point tie. For the
unpenalized intercept, $\ell_0=0$ and the update is $\zeta_0/v_0$.
Thus every coordinate update globally minimizes its actual one-dimensional
conditional SCAD problem, without rescaling a non-unit column to a
unit-norm threshold.

\subsection{Implementation defaults}\label{supp:impl-defaults}

The path starts from the intercept-only least-squares state. If $\bm r_0$
is its residual, we set
\[
 \lambda_{\max}=(1+10^{-4})
 \max_{j:f_j>0}
 \frac{|\mathbf Z_j^\top\bm r_0|/n}
      {f_j\min\{1,\sqrt{(a+1)v_j}\}} .
\]
If the maximum score is exactly zero, we set
$\lambda_{\max}=10^{-6}$ as a deterministic positive numerical fallback that
only defines a finite grid. In either case $\lambda_{\max}$ is at least the
null-model KKT value (strictly larger when a column norm is small), so
the intercept-only state is a global minimum of every scalar coordinate
subproblem at the first grid point. We then follow a decreasing geometric grid
with Gram-matrix Gauss--Seidel sweeps and warm starts
\citep[for coordinate descent with nonconvex penalties, see][]{breheny11}.

After a sweep, convergence is accepted only if both the largest
column-norm-scaled coordinate change and a freshly recomputed coordinate
fixed-point gap
\[
  \Delta_\lambda(\bm b)=
  \max_j\sqrt{v_j}\,|T_{\lambda,j}(\bm b)-b_j|
\]
are below the stated tolerance, where $T_{\lambda,j}$ is the global scalar
update above with all other coordinates fixed. At each $\lambda$, the full
SCAD objective is recomputed independently. A candidate is retained only
when it is finite, has passed the fixed-point test, and does not exceed the
same-$\lambda$ objective of the warm start. Otherwise the warm start is
retained and the failure reason is recorded.

\begin{table}[!htbp]
\centering
\scriptsize
\caption{Implementation defaults for the distilled estimator and the
sparse-LTS comparator. Path, selector, and solver constants apply to every
reported unweighted distilled fit; longer paths and their endpoint checks
are stated explicitly for studies that use them.}
\label{tab:impl-defaults}
\setlength{\tabcolsep}{3pt}
\begin{tabular}{@{}llp{0.21\textwidth}p{0.29\textwidth}@{}}
\toprule
Component & Parameter & Default & Notes\\
\midrule
SCAD penalty & $a$ & $3.7$ & \citep{fanli01}\\
Path & grid points & $100$ & decreasing geometric grid\\
Path & min-to-max ratio & $0.001$ / $0.05$ & $n>p+1$ / otherwise\\
GIC & $\kappa$ & $2$ & sensitivity assessed numerically\\
Degrees of freedom & threshold on $|\hat\beta_j|$ & $10^{-10}$ & GIC and BIC; intercept always counted\\
Support reporting & numerical zero & $10^{-8}$ & displayed support metrics\\
Solver & convergence tolerance & $10^{-5}\times$ RMS of $\tilde\y$ & unit fallback\\
Solver & maximum sweeps & $10{,}000$ & per grid point\\
Initial fits & \texttt{MASS::rlm} iterations & $200$ & Huber and MM \citep{venablesripley02}\\
Huber fit & $\psi$ constant & $1.345$ & package default\\
MM fit & bisquare constant & $4.685$ & S-start, final Tukey bisquare\\
Scale & $\hat\sigma$ & MAD--RMS--unit fallback & \eqref{eq:sigma}; computed separately per initial estimator\\
Weighted variant & MCD quantile & $0.975$ & $q_{0.975,p}$, Section~\ref{sec:init}\\
Sparse LTS & primary fraction grid & $0.20$, $0.10$, $0.05$, $0.025$, $0.0125$ & fraction mode; BIC selection\\
Sparse LTS & sensitivity fractions & $0.15$, $0.075$, $0.0375$, $0.01875$, $0.00625$ & added to the primary five\\
Sparse LTS & sampling controls & $500$ / $10$ & initial / subsequent draws\\
Sparse LTS & coefficients & reweighted & default \texttt{coef}; internal normalization enabled\\
\bottomrule
\end{tabular}
\end{table}

Raw-response SCAD
selects
\[
 n\log\{\RSS(\lambda)/n+10^{-12}\}
 +\log(n)\,\mathrm{df}(\lambda),
\]
whereas distilled SCAD uses the distillation-fidelity GIC in
Algorithm~\ref{alg:distill}.

\subsection{The failure-aware initial-fit cascade}\label{supp:impl-cascade}

The reported implementation is failure-aware: an MM error or nonfinite
coefficient return triggers LTS, an analogous failure of LTS triggers Huber,
an analogous Huber failure triggers OLS, and a finite but nonconverged MM
return is retained with status \texttt{maxit}. The cascade is an
implementation safeguard that lies outside the robustness theorem, and it
defines a composite estimator rule. Corollary~\ref{cor:bp} applies only when
the complete initial rule under consideration remains bounded over the
stated response-replacement neighborhood, and the implementation status is
reported only as numerical provenance.

\subsection{The weighted variant}\label{supp:impl-weighted}

Because contaminated rows survive
into the second-stage design, we also define \emph{weighted distillation},
replacing the quadratic term in \eqref{eq:step2} by
\[
 \frac1{2n}\sum_{i=1}^n\hat w_i^*
 \{\tilde\alpha+\bm x_i^\top\btil-\alpha-\bm x_i^\top\bbeta\}^2,
 \qquad \hat w_i^*=\hat u_i/\bar u,
\]
where $\hat u_i=\min\{1,q_{0.975,p}/\max(\mathrm{RD}_i^2,10^{-12})\}$,
$\mathrm{RD}_i$ is the robust Mahalanobis distance from a minimum covariance
determinant (MCD) fit
\citep{rousseeuwvandriessen99}, and
$q_{0.975,p}$ is the $0.975$ chi-square quantile. The normalization makes the
weights average one, so $\hat w_i^*$ need not be at most one. If
the MCD fit fails, the implementation falls back to unit weights.
For this variant the GIC also replaces its unweighted residual sum of
squares by
$\mathrm{RSS}_{d,w}=\sum_i\hat w_i^*
\{\tilde y_i-\hat\alpha_d-\bm x_i^\top\bcomp_d\}^2$, while retaining
the same degrees-of-freedom penalty.
We treat this as an exploratory extension that is not evaluated in the main
numerical study. The conditional fixed-design response-replacement and
oracle-branch results of Sections~\ref{sec:finite} and~\ref{sec:asymp} of
the main text do not automatically extend to data-dependent leverage
weights.


\end{document}